%% file: camera_ready.tex
\documentclass{article} 
\usepackage{iclr2026_conference,times}
\usepackage{bbm}
\usepackage{mathtools}
\newcommand{\defeq}{\vcentcolon=}

\newcommand{\sg}[1]{{\text{sg}{[#1]}}}
\usepackage{newtxtext} 
\usepackage[T1]{fontenc}
\usepackage{amsfonts}
\input{math_commands.tex}
\usepackage{amsthm}
\usepackage{enumitem}
\usepackage{booktabs}
\usepackage{caption}
\usepackage{adjustbox}
\usepackage{hyperref}
\usepackage{url}
\usepackage{cleveref}
\usepackage{wrapfig}
\usepackage{graphicx}
\usepackage{subfig}
\usepackage[noend]{algpseudocode}
\usepackage{algorithm}
\usepackage{algorithmicx}
\usepackage{algpseudocode}
\usepackage[most]{tcolorbox}
\usepackage[dvipsnames]{xcolor}
\usepackage{amsmath}
\usepackage{cleveref}
\usepackage{makecell}
\newtheorem{theorem}{Theorem}
\newtheorem{prop}{Proposition}
\newtheorem{lemma}{Lemma}
\newtheorem{corollary}{Corollary}

\newcommand{\name}{SPG}

\newcommand{\done}{\textsc{D1}}
\newcommand{\wdone}{\textsc{WD1}}

\definecolor{ForestGreen}{RGB}{34,139,34}
\renewcommand{\th}{{\bm{\theta}}}
\newcommand{\lb}[1]{{\small \textcolor{magenta}{\text{#1}}}}

\newcommand{\iconlink}[3]{
  \href{#2}{\raisebox{-0.2ex}{#1}\,\texttt{#3}}\xspace
}
\usepackage{fontawesome5}
\usepackage{xspace}
\newcommand{\projectlinks}[2]{
  \begin{center}
    \iconlink{\faGithub}{https://github.com/facebookresearch/SPG}{#1}
    \quad
    \iconlink{\faGlobe}{https://chenyuwang-monica.github.io/spg/}{#2}
  \end{center}
}

\title{\name: Sandwiched Policy Gradient for Masked Diffusion Language Models}
\author{
\stepcounter{footnote}
\hspace{-5.5pt}\textbf{Chenyu Wang$^{1,2,}$\thanks{Work done at Meta.}$\;\,$, 
Paria Rashidinejad$^{1,3,\dagger}$, DiJia Su$^1$, Song Jiang$^1$,} \\
\textbf{Sid Wang$^1$, Siyan Zhao$^{1,4,\dagger}$, Cai Zhou$^2$, Shannon Zejiang Shen$^{1,2,\dagger}$,} \\
\textbf{Feiyu Chen$^1$, Tommi Jaakkola$^2$, Yuandong Tian$^1$, Bo Liu$^1$} \\
\ $^1$Meta Superintelligence Labs \ $^2$MIT \ $^3$USC \ $^4$UCLA
}

\newcommand{\rebuttal}[1]{\textcolor{black}{{#1}}}

\definecolor{linequote}{RGB}{224,215,188}
\definecolor{backquote}{RGB}{249,245,233}

\newtcolorbox{myquote}{%
    enhanced, breakable, 
    size=fbox,
    frame hidden, boxrule=0pt,
    sharp corners,
    colback=backquote,
    borderline horizontal={.5pt}{0pt}{linequote},
    borderline horizontal={.5pt}{1pt}{linequote}
}
\iclrfinalcopy 
\begin{document}

\maketitle
\vspace{-18pt}
\projectlinks{https://github.com/facebookresearch/SPG}{Project Page}

\begin{abstract}
Diffusion large language models (dLLMs) are emerging as an efficient alternative to autoregressive models due to their ability to decode multiple tokens in parallel. However, aligning dLLMs with human preferences or task-specific rewards via reinforcement learning (RL) is challenging because their intractable log-likelihood precludes the direct application of standard policy gradient methods. While prior work uses surrogates like the evidence lower bound (ELBO), these one-sided approximations can introduce significant policy gradient bias. To address this, we propose the Sandwiched Policy Gradient (\name{}) that leverages both an upper and a lower bound of the true log-likelihood. Experiments show that \name{} significantly outperforms baselines based on ELBO or one-step estimation. Specifically, \name{} improves the accuracy over state-of-the-art RL methods for dLLMs by 3.6\% in GSM8K, 2.6\% in MATH500, 18.4\% in Countdown and 27.0\% in Sudoku.

\begin{figure}[!h]
    \centering
    \vspace{5pt}\includegraphics[width=1.0\linewidth]{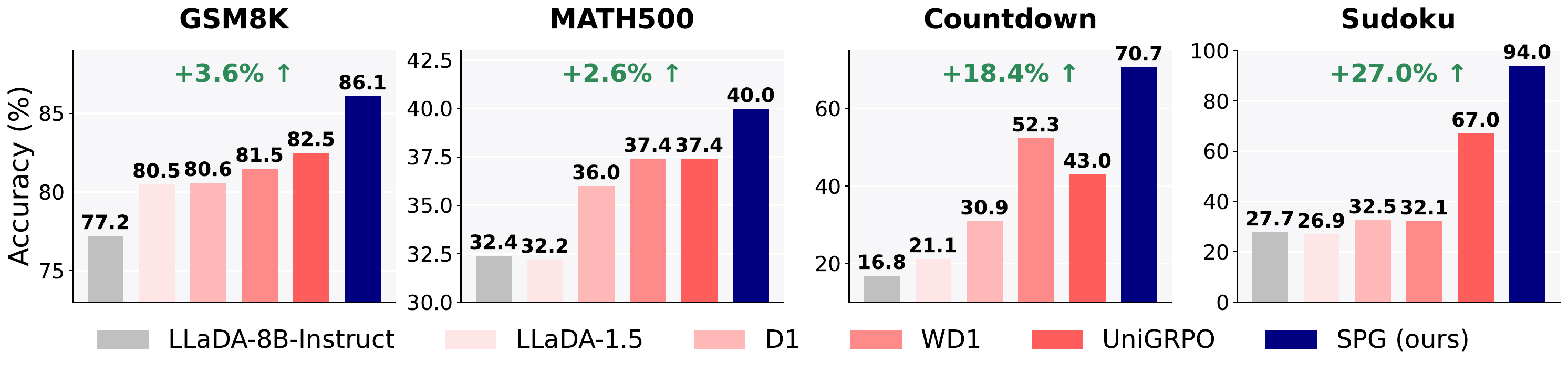}
    \vspace{-15pt}
    \caption{Test accuracy of \name{} and baseline methods on four mathematical and logical reasoning benchmarks. All methods are evaluated with a generation length of 256 in 128 denoising steps. Full results are provided in \Cref{tab:main_results}.}
    \vspace{-5pt}
    \label{fig:barplot}
\end{figure}
\end{abstract}

\section{Introduction}
Diffusion models, originally pioneered for high-fidelity image generation~\citep{song2020score,ho2020denoising}, have recently emerged as a powerful and efficient paradigm for text generation~\citep{austin2021structured,campbell2022continuous,sun2022score,sedd,mdlm,shi2024simplified}. These models operate in a discrete space but share architectural similarities with their continuous counterparts~\citep{dit}. They employ a fixed noising process that progressively corrupts text data, while a neural network is trained to learn the reverse, denoising process. For instance, Masked Diffusion Language Model (MDLM)~\citep{mdlm} uses random masking as its forward noising process and optimizes an Evidence Lower Bound (ELBO) of the log-likelihood. This ELBO-based objective has been widely adopted by subsequent large-scale diffusion language models (dLLMs), including LLaDA~\citep{llada} and DREAM~\citep{dream}.

A key advantage of dLLMs over their autoregressive (AR) counterparts is their ability to decode multiple tokens in parallel. This parallelism can significantly reduce inference latency, making it an attractive alternative for scalable language modeling~\citep{d2f,labs2025mercury}.

Aligning large language models with human preferences~\citep{rlhf} or task-specific rewards (e.g., inducing reasoning behavior)~\citep{shao2024deepseekmath,deepseekr1} typically requires a post-training stage of reinforcement learning (RL). However, applying RL to dLLMs remains underexplored. A principal challenge is the computationally intractable log-likelihood of dLLMs, which is essential for accurate policy gradient estimation. To circumvent this, recent works~\citep{d1,mmada,llada1.5,tang2025wd1} adapt standard RL and preference optimization algorithms, such as GRPO~\citep{shao2024deepseekmath} and DPO~\citep{dpo}, by using the ELBO or a one-step estimation as a surrogate for the true likelihood. While straightforward, this approximation leads to misaligned policy gradients, and potential suboptimal performance. \rebuttal{\citet{lladou} employs a trajectory-level policy update algorithm together with a decoding order prediction module, jointly optimizing both model predictions and the order of unmasking. However, accumulating gradients across the entire diffusion trajectory incurs significant computational costs.}

\begin{figure}
    \centering
    \includegraphics[width=\linewidth]{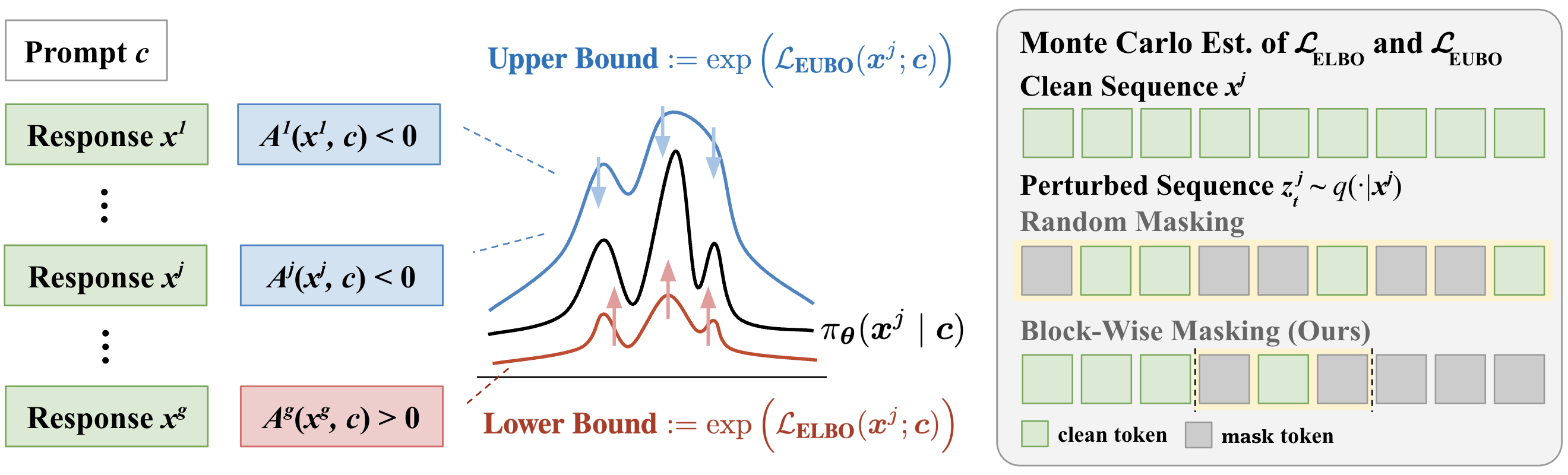}
    \vspace{-15pt}
    \caption{The training process of \name{} for MDLM. \textit{Left:} From a prompt $\bm{c}$, we generate responses $\{\bm{x}^{j}\}_{j=1}^g$. We then maximize a lower bound on the likelihood $\pi_{\bm{\theta}}(\bm{x}^j\mid\bm{c})$ for high-reward responses while minimizing an upper bound for low-reward ones. \textit{Right:} The upper/lower bound of likelihood is estimated via Monte Carlo using a block-wise masking strategy, where a random block is selected for masking, with earlier blocks kept clean and later blocks fully masked. The example shows a sequence of length 9 with a block size of 3, where the current generation block is highlighted in yellow.
    }
    \vspace{-10pt}
    \label{fig:main}
\end{figure}

To address these limitations, we propose Sandwiched Policy Gradient (\name{}), a novel reinforcement learning algorithm for diffusion language models that computes a more robust and less biased policy gradient. As illustrated in \Cref{fig:main}, our core idea is to ``sandwich'' the intractable log-likelihood of a generated sequence: we maximize a tractable lower bound for positive-reward sequences while minimizing an upper bound for negative-reward ones. To ensure a stable estimation of these bounds, we also propose a block-wise masking strategy that better aligns data distributions during policy rollout and optimization. \name{} achieves state-of-the-art performance on four mathematical and logical reasoning benchmarks, improving accuracy by up to 3.6\% on GSM8K, 2.6\% on MATH500, 18.4\% on Countdown, and 27.0\% on Sudoku compared to the state-of-the-art
RL algorithms for diffusion language models. \rebuttal{We additionally extend \name{} to coding tasks, achieving improved accuracy by 1.9\% on HumanEval and 4.7\% on MBPP.}

In summary, our main contributions are:
\begin{itemize}[leftmargin=*]
\item A new policy gradient algorithm, \name{}, which reduces bias by optimizing sandwiched variational bounds based on reward.
\item A block-wise masking technique that improves the stability of the training objective's estimation.
\item State-of-the-art results among RL algorithms for diffusion language models on four reasoning benchmarks \rebuttal{and two coding benchmarks}, demonstrating the effectiveness of our approach.
\end{itemize}


\section{Background}
\label{sec:background}
In this section, we provide a brief overview of the masked diffusion language model (MDLM) and reinforcement learning for text diffusion models.

\paragraph{Notation.}
We denote scalars by lowercase letters ($x$), vectors by bold lowercase ($\bm{x}$), and sequences by $\bm{x}_{1:n}$. $[k]$ represents $\{1, \dots, k\}$. $\text{Cat} (\bm{x}\mid\bm{p})$ is the categorical distribution over $\bm{x}$ with
probabilities $\bm{p}$, and $\mathcal{U}[a, b]$ denotes the uniform distribution in $[a, b]$. Throughout the paper, we use $i \in [n]$ for position of the token, $j \in [g]$ for a sequence in a group of rollouts, and $t$ for the diffusion timestep. For discrete time processes, $t\in[T]$, while for continuous-time Markov chains, $t \in [0,1]$.

\subsection{Masked Diffusion Language Models}
Diffusion models for language learn to generate text by reversing a gradual noising process. Specifically, Masked Diffusion Language Models (MDLMs)~\citep{mdlm} start with clean text $\bm{x}_{1:n}$ and corrupt it into $\bm{z}_t \equiv \bm{z}_{t, 1:n}$ over a continuous timestep $t \in [0, 1]$ by progressively replacing tokens with a special \mask{} token. At $t=0$, the data is original ($\bm{z}_0 = \bm{x}$), while at $t=1$, the sequence is fully masked ($\bm{z}_1$ is all \mask tokens). 
Each token is corrupted independently according to the forward transition kernel:
\begin{equation}
    q_{t \mid 0}(\bm{z}_{t,i} \mid \bm{x}_i) = \text{Cat}\big(\bm{z}_{t,i} \mid \alpha_t \bm{x}_i + (1 - \alpha_t) \bm{m}\big),
\end{equation}
where $\bm{m}$ is the one-hot representation of the \mask token. The noise schedule, $\alpha_t \in [0, 1]$, is a strictly decreasing function, such as the linear schedule $\alpha_t = 1-t$, with $\alpha_0 = 1$ and $\alpha_1 = 0$.

In the reverse process, a neural network, which we denote as the policy $\pi_{\bm{\theta}}$, is then trained to perform the reverse process: predicting the original tokens $\bm{x}$ from a corrupted version $\bm{z}_t$.
The transition from $\bm{z}_{t}$ to $\bm{z}_s$ ($s<t$) is parameterized with $\pi_{\bm{\theta}}$ as follows:
\begin{equation*}
p_\theta(\bm{z}_s \mid \bm{z}_t) = q\left( \bm{z}_s \mid \bm{z}_{t}, \bm{x} = \pi_{\bm{\theta}}(\cdot\mid\bm{z}_{t}) \right)=\begin{cases}
    \text{Cat}(\mathbf{z}_s; \mathbf{z}_{t}), & \mathbf{z}_{t} \neq \mathbf{m}, \\
    \text{Cat}\left(\mathbf{z}_s; \frac{(1-\alpha_s)\mathbf{m} + (\alpha_s - \alpha_{t})\pi_{\bm{\theta}}(\cdot\mid\bm{z}_{t}) }{1-\alpha_{t}}\right), & \mathbf{z}_{t} = \mathbf{m}.
\end{cases}
\end{equation*}

The policy is achieved by maximizing the Evidence Lower Bound (ELBO) of the log-likelihood of each clean sequence $\bm{x}\sim p_{\text{data}}$, which simplifies to the following objective:
\begin{equation}
    \mathcal{L}_\text{ELBO}(\bm{x};\bm{\theta}) = \mathbb{E}_{t, \bm{z}_t} \bigg[\sum_{i=1}^{n} w(t) \cdot \mathbbm{1}(\bm{z}_{t, i} = \bm{m} ) \cdot \log \pi_{\bm{\theta}}(\bm{x}_i \mid \bm{z}_t) \bigg],
\label{eq:elbo}
\end{equation}
where $w(t)=\alpha_t'/(\alpha_t-1)$ is a time-dependent loss weight, and the expectation is over a random timestep $t \sim \mathcal{U}[0,1]$ and the corrupted sequence $\bm{z}_t\sim q_{t\mid0}(\cdot\mid\bm{x})$. In essence, this objective trains the model to ``fill in the blanks'' by predicting the original tokens at masked positions. For a more comprehensive overview of MDLM, please refer to \Cref{app:baiscs-of-dllm} and \citet{mdlm}.

\subsection{Reinforcement Learning for Diffusion Language Models}
Reinforcement Learning (RL) aligns a language model with desired objectives by treating it as a policy $\pi_{\bm{\theta}}$ that generates a response $\bm{x}$ to a prompt $\bm{c}$. \rebuttal{This can be defined as a Markov decision process (MDP)~\citep{black2023training}, where at each timestep $t$, the model observes the prompt $\bm{c}$ and a partially masked sequence $\bm{z}_t$ as the current state, takes an action based on the policy $\pi_{\bm\theta}$, and transitions to a new state $\bm{z}_{t-1}$ based on the action and inference strategy.} A reward function $R(\bm{c}, \bm{x})$ provides a scalar score for the complete response, and the training goal is to update $\bm{\theta}$ to maximize the expected reward: $\mathcal{J}(\bm{\theta}) \defeq \mathbb{E}_{\bm{x} \sim \pi_{\bm{\theta}}(\cdot \mid \bm{c})}[R(\bm{c}, \bm{x})]$. This objective is commonly optimized using policy gradient methods, which rely on the following gradient estimator.
\begin{equation}
\nabla_{\bm{\theta}} \mathcal{J}(\bm{\theta}) = \mathbb{E}_{\bm{x} \sim \pi_{\bm{\theta}}(\cdot \mid \bm{c})} \bigg[ R(\bm{c}, \bm{x}) \nabla_{\bm{\theta}} \log \pi_{\bm{\theta}}(\bm{x} \mid \bm{c}) \bigg].
\label{eq:pg}
\end{equation}

\paragraph{The Intractability Challenge.}
A central challenge in applying RL to diffusion models is that the policy's log-likelihood, $\log \pi_{\bm{\theta}}(\bm{x} \mid \bm{c})$, is \emph{intractable} and cannot be computed directly. To overcome this, prior work~\citep{llada1.5, mmada} approximates this term using its ELBO, effectively replacing $\log \pi_{\bm{\theta}}(\bm{x} \mid \bm{c})$ with a score derived from the pre-training objective in \Cref{eq:elbo}.

However, this popular workaround introduces a critical flaw. The ELBO is only a \textit{lower bound} on the true log-likelihood ($\text{ELBO} \le \log \pi_{\bm{\theta}}$). Consequently, the RL objective is only a valid lower bound on the true expected reward if all rewards $R(\bm{c}, \bm{x})$ are \emph{non-negative}. This constraint prevents the model from effectively learning from negative feedback (i.e., penalizing bad outputs) and is incompatible with advanced RL algorithms that use relative or negative rewards~\citep{shao2024deepseekmath}, biasing the final policy. Our work aims to resolve this limitation.


\begin{algorithm}[t]
\caption{\name{}: Sandwiched Policy Gradient for Masked dLLMs}
\begin{algorithmic}[1]
\Require prompt distribution $\mathcal{D}$, number of completions per prompt $g$, number of inner updates $\mu$, forward process $q$, number of Monte Carlo samples $m$, initial policy $\pi_0$, learning rate $\epsilon$.
\State Initialize $\pi_\theta \gets \pi_0$
\While{not converged}
    \State Sample a prompt $\bm{c}\sim\mathcal{D}$, then $g$ completions $\{\bm{x}^j\sim\pi_{\theta}(\cdot\mid\bm{c})\}_{j=1}^g$
    \State $\forall j \in [g]$, compute reward $R(\bm{c},\bm{x}^j)$ and advantage $A^j(\bm{x}^j,\bm{c})$
    \For{gradient update iterations $\{1,\dots,\mu\}$}
    \State $\forall j \in [g]$, generate $m$ perturbed samples $\{\bm{z}^j_{t_\tau}\}_{\tau=1}^m \sim q(\cdot \mid \bm{x}^j)$ via block-wise masking (\Cref{subsec:method.block}).
    \State Compute the sandwiched policy gradient $\nabla \mathcal{J}_{\text{\name}}(\th)$ where:
    $$\mathcal{J}_{\text{\name}}(\th) =  \mathbb{E} \bigg[ \frac{1}{g} \sum_{j=1}^g \left( \mathbbm{1}_{A^j\geq0} \cdot A^j \mathcal{L}_{\text{ELBO}}(\bm{x}^j\mid \bm{c}; \th) + \mathbbm{1}_{A^j < 0} \cdot A^j \tilde{\mathcal{L}}_{\text{EUBO}}(\bm{x}^j\mid \bm{c}; \th)\right) \bigg],$$
    \State and $\mathcal{L}_{\text{ELBO}}$, $\tilde{\mathcal{L}}_{\text{EUBO}}$ are estimated from $\{\bm{z}^j_{t_\tau}\}_{\tau=1}^m$, using \Eqref{eq:elbo} and \ref{eq:eubo}.
    \State Perform gradient update: $\theta \leftarrow \th + \epsilon \nabla \mathcal{J}_{\text{\name}}(\th)$
    
    \EndFor
\EndWhile
\State \Return $\pi_{\theta}$
\end{algorithmic}
\label{algo:spg}
\end{algorithm}


\section{Sandwiched Policy Gradient with Evidence Bounds}
\label{sec:method}
We introduce \name, a novel policy gradient algorithm designed for masked diffusion language models (\Cref{algo:spg}). Our method aims to address a critical issue in applying reinforcement learning to dLLMs by creating a valid optimization objective based on tractable bounds of the model's evidence.

\subsection{A Lower Bound Objective for Policy Optimization}
\label{subsec:method.objective}
Our approach is based on group relative policy optimization~\citep{shao2024deepseekmath,drgrpo}. For a given prompt $\bm{c}$, we generate a group of $g$ responses $\{\bm{x}^j\}_{j=1}^g$ from the policy $\pi_\th$. We then compute the advantage $A^j(\bm{c}, \bm{x}^j) \defeq R(\bm{c}, \bm{x}^j) - \frac{1}{g}\sum_{\jmath=1}^g R(\bm{c}, \bm{x}^\jmath)$. Moreover, we transform the conventional policy optimization objective as an advantage-weighted log-likelihood objective, for reasons that will be clear later:
\begin{equation}
\label{eq:jgroup}
    \mathcal{J}^{\text{group}}(\th) 
    = \mathbb{E}_{\bm{c}, \{\bm{x}^j\} \sim \pi_{\sg{\th}}} \bigg[ ~\frac{1}{g} \sum_{j=1}^g A^j(\bm{x}^j, \bm{c}) \log \pi_{\th}(\bm{x}^j \mid \bm{c}) ~\bigg],
\end{equation}
where $\sg{\th}$ indicates that gradients are not computed for the policy that generates the samples. This objective encourages generations with positive advantages ($A^j > 0$) and discourages those with negative advantages ($A^j < 0$).

For dLLMs, the log-likelihood $\log \pi_{\th}$ is intractable. A common surrogate is the evidence lower bound (ELBO). While maximizing the ELBO is a valid way to increase the true log-likelihood, \textit{minimizing} the ELBO for negatively-rewarded samples does not guarantee a reduction in the true log-likelihood. To address this, we propose a \emph{sandwiched} objective. For samples with positive advantages, we maximize the ELBO. For samples with negative advantages, we instead minimize a tractable evidence \emph{upper} bound (EUBO), $\mathcal{L}_{\text{EUBO}}$. This creates a true lower bound for the original objective:
\begin{equation}
\label{eq:jspg}
    \mathcal{J}_{\text{\name}}(\th) = \mathbb{E} \bigg[~ \frac{1}{g} \sum_{j=1}^g \left( \mathbbm{1}_{A^j\geq0} \cdot A^j \mathcal{L}_{\text{ELBO}}(\bm{x}^j\mid \bm{c}; \th) + \mathbbm{1}_{A^j < 0} \cdot A^j \mathcal{L}_{\text{EUBO}}(\bm{x}^j\mid \bm{c}; \th) \right) \bigg],
\end{equation}
where the expectation is take with respect to ${\bm{c}, \{\bm{x}^j\} \sim \pi_{\sg{\th}}}$.
Since $\mathcal{L}_{\text{ELBO}} \leq \log \pi_\th \leq \mathcal{L}_{\text{EUBO}}$, it follows that $\mathcal{J}_{\text{\name}}(\th) \leq \mathcal{J}^{\text{group}}(\th)$. Maximizing this tractable bound therefore serves as a valid proxy for optimizing the true objective.

\subsection{A Tractable Evidence Upper Bound}
\label{subsec:method.eubo}
To effectively penalize negatively-rewarded samples by minimizing their log-likelihood, we require a tractable EUBO, which we derive in the following theorem based on the R\'enyi variational bound.

\begin{theorem}[Evidence Upper Bound for Masked Diffusion] Assume the forward denoising process has $T$ steps with a monotonic schedule $\alpha_t$. For any $\beta \geq 1$ and a sequence $\bm{x}_{1:n}$, we have:
\begin{equation}
\label{eq:eubo_formula}
    \mathcal{L}_\text{EUBO}(\bm{x}_{1:n};\bm\th) = \frac{1}{\beta}\sum_{i=1}^n \log \sum_{t=1}^{T-1}\mathbb{E}_{\bm{z}_{t+1}} \Bigg[ \frac{\alpha_t-\alpha_{t+1}}{1-\alpha_{t+1}} \cdot \mathbbm{1}(\bm{z}_{t+1,i} = \bm{m}) \cdot \pi_{\th}^{\beta}(\bm{x}_i \mid \bm{z}_{t+1}) \Bigg] +C(T), 
\end{equation}
where $C(T)\defeq \mathbbm{1}(\beta<n) \cdot \frac{1}{\beta} \log \E_{\bm{z}_{1:T} \sim q(\cdot \mid \bm{x})} \Big[q(\bm{z}_{1:T}\mid\bm{x})^{-n}\Big]$ is a constant independent of $\bm\th$.
\label{thm:eubo}
\end{theorem}
Here, $\beta \geq 1$ is a hyperparameter that controls the tightness of the bound, with values closer to 1 yielding a tighter bound. The expectation is taken over the timestep $t \sim \mathcal{U}[0,1]$ and the noised latent $\bm{z}_t \sim q_{t\mid 0}(\cdot \mid \bm{x})$.
\begin{corollary}
Taking the limit of $T \rightarrow \infty$, we have:
\begin{equation}
\begin{split}
&\nabla_{\bm\th} \mathcal{L}_{\text{EUBO}}(\bm{x}_{1:n};\bm\th) = \nabla_{\bm\th}\Big(\tilde{\mathcal{L}}_{\text{EUBO}}(\bm{x}_{1:n};\bm\th) +C(T)\Big) = \nabla_{\bm\th}\tilde{\mathcal{L}}_{\text{EUBO}}(\bm{x}_{1:n};\bm\th),~~~~\text{where} \\
&\tilde{\mathcal{L}}_\text{EUBO}(\bm{x}_{1:n}; \th) = \frac{1}{\beta} \sum_{i=1}^n \log \mathbb{E}_{t, \bm{z}_t} \left[ 
w(t) \cdot \mathbbm{1}(\bm{z}_{t, i} = \bm{m} ) \cdot \pi^\beta_{\bm{\theta}}(\bm{x}_i \mid \bm{z}_t)
\right].
\end{split}
\label{eq:eubo}
\end{equation}
\label{coro:eubo}
\vspace{-5pt}
\end{corollary}
In practice, we estimate $\tilde{\mathcal{L}}_\text{EUBO}$ using Monte Carlo sampling and plug it in \Eqref{eq:jspg} in place of $\mathcal{L}_\text{EUBO}$. The proof and theoretical analysis are provided in Appendix~\ref{app:theory}.

\paragraph{Remark.}
A key structural difference from $\mathcal{L}_{\text{ELBO}}$ is that the logarithm in $\mathcal{L}_{\text{EUBO}}$ (\Cref{eq:eubo_formula}) appears outside the expectation. 
Therefore, in practice, due to Jensen's inequality, applying the concave logarithm to a Monte Carlo estimate of the expectation's argument yields a biased estimate of the true EUBO. While it is possible to derive a looser but unbiased bound using inequalities like $\log(x) \leq x-1$, we found this approach empirically worse by widening the gap to the true log-likelihood, as shown in \Cref{tab:ablation_loose}. We therefore retain the tighter, albeit slightly biased, formulation.

\subsection{Practical Considerations}

\paragraph{Block-Wise Masking Strategy for Monte Carlo Estimation.}
\label{subsec:method.block}
In practice, we approximate $\mathcal{L}_{\text{ELBO}}$ and $\tilde{\mathcal{L}}_{\text{EUBO}}$ in \Cref{eq:jspg} via Monte Carlo sampling: for each $\bm{x}^j$, we randomly sample $m$ timesteps $\{t_{\tau}\}_{\tau=1}^m$ and generate the corresponding partially masked samples $\{\bm{z}_{t_{\tau}}^j\}_{\tau=1}^{m} \sim q(\cdot\mid\bm{x}^j)$. One straightforward approach as used in \citet{mmada} would be to apply random masking to clean sequences. However, recent dLLMs like LLaDA~\citep{llada} employ a block-wise semi-autoregressive unmasking strategy during generation and achieve state-of-the-art performance over random unmasking. As a result, the policy rollout process actually encounters a much narrower and more structured set of partially masked sequences than with fully random masking.

To better align data distributions during policy rollout and optimization, we adopt a block-wise masking strategy rather than random masking. As depicted in \Cref{fig:main}, the sequence is divided into several blocks, and a random block is selected, with all preceding blocks left clean and all following blocks fully masked. Within the chosen block, tokens are randomly masked. Additionally, following \done~\citep{d1}, we lightly perturb the prompt and clean blocks by randomly masking tokens with a small probability $p_{\text{mask}}=0.15$ to enhance stability and generalization.

Altogether, our block-wise masking strategy improves the stability of the objective's estimation and the efficiency of policy optimization. While similar block-wise masking approaches have been explored in concurrent work for supervised fine-tuning or block diffusion models~\citep{sun2025blockwise,wang2025revolutionizing}, our focus is on RL for full-attention masked dLLMs. As shown in \Cref{fig:ablation_inference}, our models trained with block-wise masking generalize well to various inference strategies.

\paragraph{Mixture of Upper and Lower Bound for Negative Advantage Traces.}
\label{subsec:method.mixture}
Monte Carlo estimation of \Cref{eq:eubo_formula} leads to a biased estimation to $\tilde{\mathcal{L}}_{\text{EUBO}}$ and potentially requires a substantial number of samples to get reliable approximations, resulting in high computational costs and instability during training. To address these challenges, we use a mixture of $\tilde{\mathcal{L}}_{\text{EUBO}}$ and $\mathcal{L}_{\text{ELBO}}$ as a more practical log-likelihood approximation for negative advantage traces:
\begin{equation}
    \label{eq:mixture}
    \tilde{\mathcal{L}}_{\text{Mix}}(\bm{x}\mid \bm{c};\bm\th) := \omega \cdot\tilde{\mathcal{L}}_{\text{EUBO}}(\bm{x}\mid \bm{c};\bm\th) + (1-\omega)\cdot \mathcal{L}_{\text{ELBO}}(\bm{x}\mid \bm{c};\bm\th)
\end{equation}
where $0\leq\omega\leq1$ is a blend coefficient. Intuitively, the upper bound $\tilde{\mathcal{L}}_{\text{EUBO}}$ sharpens the model decisions by applying a $\beta$-power adjustment to the original model output, acting as a strong correction signal for negative advantage traces. In contrast, the lower bound $\mathcal{L}_{\text{ELBO}}$ is easier and more stable to estimate with a small number of Monte Carlo samples, but it tends to introduce larger, systematic bias relative to the true log-likelihood. In particular, as a conservative approximation, $\mathcal{L}_{\text{ELBO}}$ alone is insufficient for effectively penalizing negative advantage traces, thus limiting its efficacy. Therefore, combining them allows us to harness the strengths of each, resulting in a more effective log-likelihood estimation in practice. In the following proposition, we formalize the advantages of using the mixture by deriving the gradient of the mixture loss and analyzing the variance of the gradient.

\begin{prop}[Optimal Mixture Strictly Reduces Variance]
Fix a coordinate $k$ and let $\rho_\beta \coloneqq {w(t, \bm{z}_t) \pi_{\bm{\theta}}^\beta(\bm{x}_i\mid\bm{z}_t,\bm{c}) }/{\E\left[ w(t, \bm{z}_t) \pi_{\bm{\theta}}^\beta(\bm{x}_i\mid\bm{z}_t,\bm{c})  \right]}$, where $w(t, \bm{z}_t) \coloneqq w(t) \mathbbm{1}(z_{t} = \bm{m})$. Then, the gradient of mixture objective (\ref{eq:mixture}) is given by 
\begin{align}\label{eq:gradient}
    g_{\omega,k}= ((1-\omega)w(t, \bm{z}_t)+\omega\rho_\beta)\,\partial_{\bm{\theta}_k} \log \pi_{\bm\theta}(\bm{x} \mid \bm{z}_t, \bm{c}).
\end{align}
If $\operatorname{Var}((\rho_\beta-w(t, \bm{z}_t))\partial_{\bm\theta_k} \log \pi_{\bm\theta}(\bm{x} \mid \bm{z}_t, \bm{c}))>0$, then $\operatorname{Var}[g_{\omega,k}]$ is a strictly convex quadratic in $\omega$ and thus admits a unique minimizer $\omega_k^\star$. Moreover,
\[
\operatorname{Var}[g_{\omega_k^\star,k}] \;<\; \min\!\big\{\operatorname{Var}[g_{0,k}],\ \operatorname{Var}[g_{1,k}]\big\},
\]
\label{prop:mixture}
\vspace{-5pt}
\end{prop}
A proof for the above proposition is provided in \Cref{app:proof}. A few remarks are in order:
\begin{itemize}[leftmargin=*]
    \item \textbf{Confidence-aware weighting:} The mixture gradient in \Cref{eq:gradient} realizes a \textit{confidence-aware weighting}: uncertain tokens with small $\pi_{\bm\theta}^\beta(\bm{x}_i\mid\bm{z}_t,\bm{c})$, indicating a low recovery chance, have a smaller weight, while confident tokens with large $\pi_{\bm\theta}^\beta(\bm{x}_i\mid\bm{z}_t,\bm{c})$ are upweighted. The sharpness is controlled by parameter $\beta$ and the blend by $\omega$. Furthermore, the convex interpolation of the confidence-aware coefficient of the upper bound with the lower bound ensures clipping tiny gradients to a minimum value and thus prevents vanishing gradients. 
    \item \textbf{Lower variance and more stable training:} According to Proposition \ref{prop:mixture}, the gradient of the optimal mixture, i.e., $g_{\omega_k^\star,k}$, has strictly smaller coordinate-wise variance than the gradient of either the lower bound ($g_{0,k}$) or the upper bound ($g_{1,k}$)\footnote{Proposition \ref{prop:mixture} extends directly to a single, coordinate-independent optimizer $\omega^\star$ obtained by minimizing the sum of coordinate-wise variances.}. In our experiments, we fix $\beta$ and $\omega$ as hyperparameters for simplicity. These values can also be adaptively adjusted during training to better match the evolving training dynamics and data distribution.
\end{itemize}
Thus, the mixture approach offers theoretical advantages over using either the upper or lower bound alone, as supported by our experimental results in \Cref{sec:experiment}. Further discussions of the mixture approach and empirical evidence of reduced gradient variance are provided in Appendix~\ref{app:discussion_mixture} and \Cref{fig:grad_norm}, and Appendix~\ref{app:toy} presents a toy example illustrating the distinct behaviors of the lower and upper bounds.
\rebuttal{We provide additional discussions on the RL algorithm, including RL stabilization techniques and a reward shifting alternate, in \Cref{app:sec:rl}.}

\section{Experiments}
\label{sec:experiment}
In this section, we present experimental results highlighting the superior performance of \name \ across various benchmarks. Further, we provide detailed analysis and ablations of \name \ to assess the contribution of each component, examine the influence of key hyperparameters, and evaluate the robustness of our approach under different inference strategies.

\subsection{Experimental Setup and Main Results}
\paragraph{Experimental Setup.}
We conduct RL fine-tuning with \name \ following the experimental settings in \done~\citep{d1} and \wdone~\citep{tang2025wd1}. We employ LLaDA-8B-Instruct~\citep{llada}, a state-of-the-art open-sourced dLLM without post-training, as the base model, and experiment on \rebuttal{six} benchmarks: two for mathematical reasoning (GSM8K~\citep{gsm8k} and MATH500~\citep{math500}), two for logical reasoning (Countdown~\citep{countdown} and Sudoku~\citep{sudoku}), \rebuttal{and two for coding (HumanEval~\citep{humaneval} and MBPP~\citep{mbpp})}. We follow the same train-test splitting, reward functions, and evaluation protocol as \done \ and \wdone, except for Sudoku. For Sudoku, to avoid train-test leakage, we take the training set from \done\ and split the data by Sudoku answers, ensuring that the test set contains entirely new puzzle solutions. This guarantees that the model cannot solve test puzzles merely by memorizing possible answers.
All experiments are conducted in the zero-shot setting, except for Sudoku, where 3-shot generation is used for both training and evaluation\footnote{We use 3-shot generation for Sudoku because zero-shot is too difficult for this task, resulting in very few meaningful RL rollouts. Few-shot examples used in our experiments are provided in \Cref{app:exp.fewshot}.}. 
For all models, we employ Low-Rank Adaptation (LoRA) with a rank of $r=128$ and scaling factor $\alpha=64$. For \name, we report results using both $\tilde{\mathcal{L}}_{\text{EUBO}}$ (i.e., \name \ w/ EUBO) and $\tilde{\mathcal{L}}_{\text{Mix}}$ (i.e., \name \ w/ Mixture) for negative advantage traces. We select the value of $\beta$ in the EUBO from $\{1.0, 1.5, 2.0\}$ based on the best average test accuracy across all generation lengths, and fix the mixture coefficient $\omega$ at 0.5.
Further experimental details are in \Cref{app:exp.dataset} and \Cref{app:exp.hyper}.

\paragraph{Baselines.} We compare our method with several recent RL algorithms for dLLMs, including \done~\citep{d1}, \wdone~\citep{tang2025wd1}, and UniGRPO~\citep{mmada}. For \done \ and \wdone, we reproduce results using the official codebases and instructions, and for fair comparison, we omit the additional SFT stage in \done\ across all models. For UniGRPO, since the code is not publicly available and the original work focuses on vision-language multimodal models, we reimplement the algorithm within our setup. For consistency, we set the number of inner gradient updates $\mu$ to 4 for all models, following GRPO~\citep{shao2024deepseekmath}. We also evaluate LLaDA-1.5~\citep{llada1.5} under our settings, which fine-tune LLaDA-8B-Instruct using VRPO, a preference optimization approach on 350K preference pairs. 
\rebuttal{We provide a detailed overview on these RL algorithms for dLLMs in \Cref{app:sec:baselines_detail} and \Cref{tab:alg_comparison}.}

\begin{table}[t!]
\centering
\caption{Model performance on four reasoning benchmarks. The best results are bolded and the second best are underlined. \name \ consistently outperforms all other methods. We denote the absolute gain of test accuracy to the previous state-of-the-art in green. 
}
\label{tab:main_results}
\adjustbox{width=\textwidth,center}{
\begin{tabular}{lccc ccc ccc ccc}
\toprule
& \multicolumn{3}{c}{\textbf{GSM8K (0-shot)}} & \multicolumn{3}{c}{\textbf{MATH500 (0-shot)}} & \multicolumn{3}{c}{\textbf{Countdown (0-shot)}} & \multicolumn{3}{c}{\textbf{Sudoku (3-shot)}}\\
\cmidrule(lr){2-4} \cmidrule(lr){5-7} \cmidrule(lr){8-10} \cmidrule(lr){11-13}
\textbf{Model / Seq Len} & \textbf{128} & \textbf{256} & \textbf{512} & \textbf{128} & \textbf{256} & \textbf{512} & \textbf{128} & \textbf{256} & \textbf{512} & \textbf{128} & \textbf{256} & \textbf{512} \\
\midrule
LLaDA-8B-Inst.
& 69.5 & 77.2 & 79.8 & 28.2 & 32.4 & 34.6 & 18.8 & 16.8 & 16.8 & 5.7 & 27.7 & 26.2 \\
LLaDA-1.5 
& 70.4 & 80.5 & 81.9 & 26.8 & 32.2 & 35.8 & 21.9 & 21.1 & 21.5 & 7.4 & 26.9 & 29.0 \\
\done
& 72.2 & 80.6 & 81.3 & 31.4 & 36.0 & \underline{39.4} & 30.9 & 30.9 & 34.4& 7.2 & 32.5 & 29.3 \\
\wdone
& 74.6 & 81.5 & 83.0 & 31.0 & 37.4 & 39.0 & 48.8 & 52.3 & 50.8 & 33.1 & 32.1 & 22.5\\
UniGRPO
& 74.9 & 82.5 & 82.7 & 32.4 & 37.4 & \underline{39.4} & 44.5 & 43.0 & 57.0 & 59.0 & 67.0 & 62.9 \\
\midrule
\name\ w/ EUBO & \underline{77.1} & \underline{83.8} & \underline{83.9} & \underline{33.2} & \underline{37.6} & \underline{39.4} & \underline{68.4} & \textbf{71.5} & \underline{68.0} & \underline{81.2} & \underline{87.1} & \underline{89.9} \\
\name \ w/ Mixture & \textbf{78.5}\textcolor{ForestGreen}{\tiny{\textbf{+3.6}}} & \textbf{86.1}\textcolor{ForestGreen}{\tiny{\textbf{+3.6}}} & \textbf{84.5}\textcolor{ForestGreen}{\tiny{\textbf{+1.5}}} & \textbf{33.4}\textcolor{ForestGreen}{\tiny{\textbf{+1.0}}} & \textbf{40.0}\textcolor{ForestGreen}{\tiny{\textbf{+2.6}}} & \textbf{41.8}\textcolor{ForestGreen}{\tiny{\textbf{+2.4}}} & \textbf{68.8}\textcolor{ForestGreen}{\tiny{\textbf{+20}}} & \underline{70.7}\textcolor{ForestGreen}{\tiny{\textbf{+18}}} & \textbf{70.3}\textcolor{ForestGreen}{\tiny{\textbf{+13}}} & \textbf{82.9}\textcolor{ForestGreen}{\tiny{\textbf{+24}}} & \textbf{94.0}\textcolor{ForestGreen}{\tiny{\textbf{+27}}} & \textbf{93.1}\textcolor{ForestGreen}{\tiny{\textbf{+30}}} \\
\bottomrule
\end{tabular}
}
\end{table}

\begin{figure}[t]
    \centering
    \includegraphics[width=\linewidth]{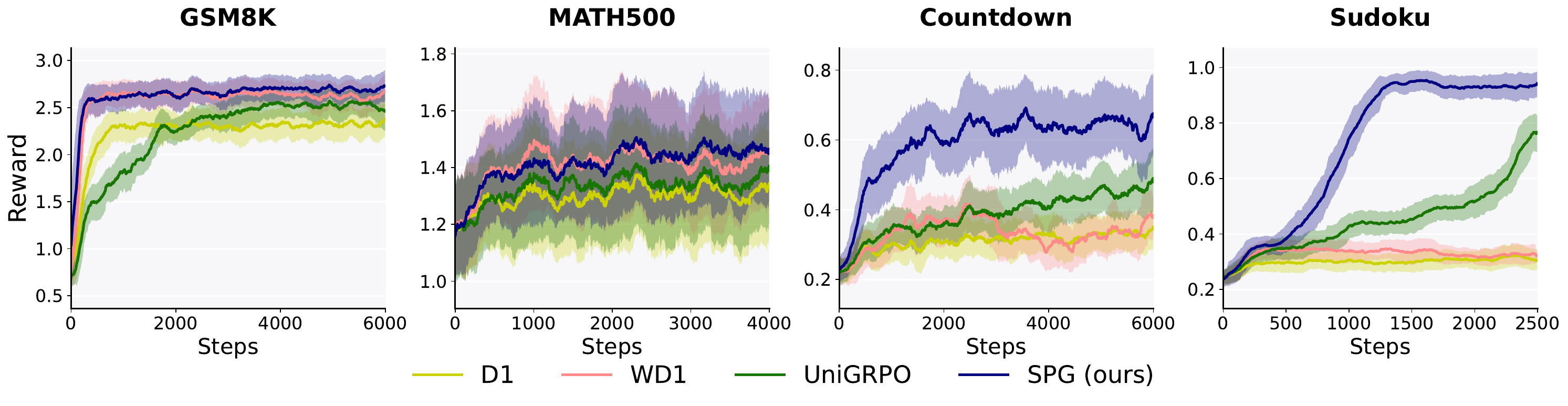}
    \vspace{-15pt}
    \caption{Reward dynamics of \name \ w/ Mixture during RL training, compared with \done, \wdone, and UniGRPO. \name \ consistently leads to faster convergence and higher reward level. We report mean and standard deviation over a rolling window of 50 steps.}
    \label{fig:reward}
\end{figure}

\paragraph{Generation and Evaluation Setup.} For both RL rollouts and evaluation, we use the semi-autoregressive confidence-based decoding strategy, following LLaDA, \done{} and \wdone. We apply the same generation setup as \done, with the denoising timestep set to half the total sequence length. The sequence is divided into blocks of 32 tokens, and in diffusion step, we unmask the 2 tokens with the highest confidence (measured by the probability of the sampled token) within the current incomplete block. During RL rollout, to encourage diverse outputs, we use a generation length of 256 and a sampling temperature of 0.9 across all benchmarks, except for sudoku, where the temperature is set to 0.3 as in \done. During evaluation, the sampling temperature is set to 0.0. We evaluate the models every 100 steps, reporting results from the checkpoint that achieves the highest average test accuracy across generation lengths of 128, 256, and 512.

\paragraph{Results.} We provide the performance of \name \ on each of the four reasoning benchmarks in comparison to the base model and other baselines in \Cref{tab:main_results}. Both \name\ w/ EUBO and \name\ w/ Mixture consistently achieve significant improvements over the baselines across all tasks and generation lengths, with the Mixture approach that combines ELBO and EUBO for negative advantage traces yielding the best performance. In particular, at a generation length of 256, \name \ w/ Mixture improves the test accuracy over the previous state-of-the-art by \textbf{3.6\%} on GSM8K, \textbf{2.6\%} on MATH500, \textbf{18\%} on Countdown, and \textbf{27\%} on Sudoku, showcasing the effectiveness of \name\ to conduct RL for dLLMs.
Reward dynamics throughout training are illustrated in~\Cref{fig:reward}, where \name\ shows a rapid and steady increase in reward over the optimization steps, further demonstrating its efficiency and robustness. 
We provide additional results and comparisons to the baselines in \Cref{tab:main_results_complete} and \Cref{app:results.eval}. 

\begin{wraptable}{r}{0.6\textwidth}
\vspace{-12pt}
\centering
\caption{\rebuttal{Model performance on HumanEval and MBPP. The best results are bolded and the second best are underlined. Absolute test accuracy gains over the previous state-of-the-art are shown in green. \name{} consistently improves accuracies on coding tasks.}}
\label{tab:main_results_coding}
\vspace{-9pt}
\adjustbox{width=0.6\textwidth,center}{
\begin{tabular}{lccc ccc}
\toprule
& \multicolumn{3}{c}{\textbf{HumanEval (0-shot)}} & \multicolumn{3}{c}{\textbf{MBPP (0-shot)}} \\
\cmidrule(lr){2-4} \cmidrule(lr){5-7}
\textbf{Model / Seq Len} & \textbf{128} & \textbf{256} & \textbf{512} & \textbf{128} & \textbf{256} & \textbf{512} \\
\midrule
LLaDA-8B-Inst. & 28.1 & 35.4 & 34.8 & 36.2 & 42.0 & 38.1 \\
LLaDA-1.5 & 22.6 & 34.8 & 32.3 & 33.5 & 44.8 & 40.1 \\
\done~ & \underline{29.3} & 37.8 & 37.2 & 42.0 & 45.5 & 43.2 \\
\wdone~ & 25.6 & 39.0 & \underline{38.4} & 38.9 & 45.9 & 44.8 \\
UniGRPO~ & 28.1 & \underline{39.6} & \underline{38.4} & 39.3 & 45.5 & 45.9 \\
\midrule
\name\ w/ EUBO & \underline{29.3} & \textbf{41.5}\textcolor{ForestGreen}{\scriptsize{+1.9}} & 33.5 & \underline{43.6} & \underline{46.7} & \underline{46.3} \\
\name\ w/ Mixture & \textbf{31.1}\textcolor{ForestGreen}{\scriptsize{+1.8}} & 39.0 & \textbf{39.6}\textcolor{ForestGreen}{\scriptsize{+1.2}} & \textbf{46.3}\textcolor{ForestGreen}{\scriptsize{+4.3}} & \textbf{50.6}\textcolor{ForestGreen}{\scriptsize{+4.7}} & \textbf{49.0}\textcolor{ForestGreen}{\scriptsize{+3.1}} \\
\bottomrule
\end{tabular}
}
\vspace{-30pt}
\end{wraptable}

\paragraph{\rebuttal{Extending \name{} to coding domains.}}
\rebuttal{We further extend \name{} to coding tasks, fine-tuning the LLaDA-8B-Instruct base model on the KodCode-Light-RL-10K dataset~\citep{xu2025kodcode} and evaluating on HumanEval and MBPP benchmarks, following the settings in \done. As shown in \Cref{tab:main_results_coding}, \name{} consistently improves the accuracy on both benchmarks over the baselines across different generation lengths, demonstrating its strong ability in the coding domain.
\vspace{3pt}
}

\subsection{Ablations and Further Analysis}
\label{subsec:ablation}
\begin{figure}[t]
    \centering
    \includegraphics[width=\linewidth]{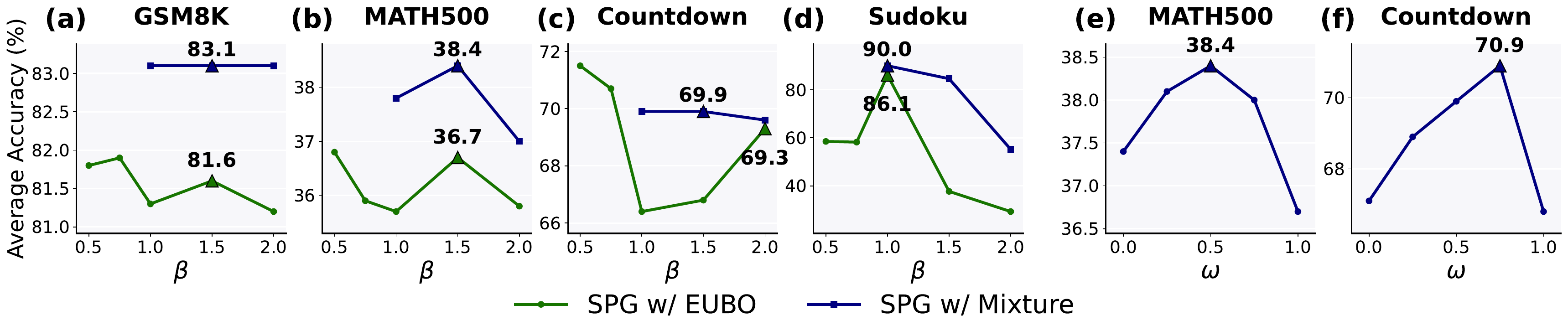}
    \caption{\emph{(a)-(d)}: ablations on the effect of $\beta$ in the upper bound; \emph{(e)-(f)}: ablations on the mixture coefficient $\omega$. The best performed $\beta\geq 1$ and $\omega \in[0,1]$ are marked by triangle in each setting.}
    \label{fig:ablation_beta_omega}
\end{figure}

\begin{figure}
    \centering
    \includegraphics[width=\linewidth]{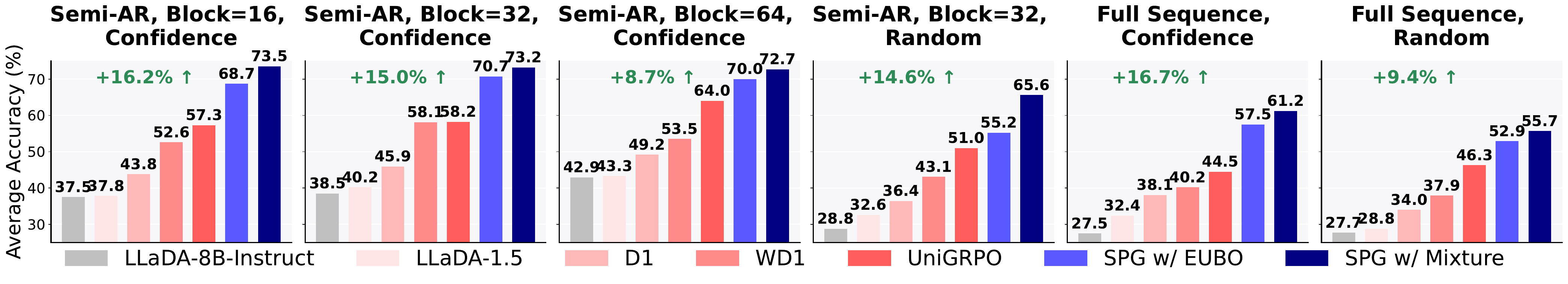}
    \caption{Ablations on inference strategies, including different combinations of decoding orders (i.e., semi-autoregressive (semi-AR) decoding with varying block sizes and full sequence decoding) and unmasking approaches (i.e., confidence-based and random unmasking). We set generation length to 256 and report the average accuracy across four benchmarks. \name \ consistently outperforms all baselines by a large margin across different inference strategies.}
    \label{fig:ablation_inference}
\end{figure}

We conduct a series of ablation studies to gain deeper insights from the following aspects:
\begin{itemize}[leftmargin=*]
    \item The contribution of each individual component, including log-likelihood estimation methods for negative advantage traces (\Cref{tab:ablation_component}) and the masking strategy in Monte Carlo estimation (\Cref{tab:ablation_mask}).
    \item The effect of key hyperparameters, including $\beta$ that controls the tightness of the upper bound and the mixture coefficient $\omega$ (\Cref{fig:ablation_beta_omega}).
    \item The robustness of our approach under various inference strategies (\Cref{fig:ablation_inference}).
\end{itemize}
Due to computational constraints, some ablation experiments are conducted on a representative mathematical reasoning benchmark (MATH500) and a logical reasoning benchmark (Countdown). Unless otherwise noted, we report average test accuracy across generation lengths 128, 256, and 512 for the ablation studies, with detailed results for each generation length provided in \Cref{app:results.ablation}.
In \Cref{app:results.ablation}, we also investigate alternative log-likelihood estimation methods for positive advantage traces in place of ELBO, as detailed in \Cref{tab:ablation_pos}, and study the diversity of model generations by evaluating the pass@K performance of each model in \Cref{tab:ablation_passatk}.
\rebuttal{Additionally, we conduct ablations on full fine-tuning, prompt masking, and number of Monte Carlo samples, and provide the computational cost comparison in \Cref{app:results.ablation}. We also conduct further comparison with the variant of UniGRPO and the validation-based checkpoint selection in \Cref{app:subsec:add_eval}.}

\paragraph{Ablations on Algorithm Components.} We first study the impact of different log-likelihood estimation methods for negative advantage traces in \Cref{tab:ablation_component}. Specifically, we compare our approach using $\tilde{\mathcal{L}}_{\text{EUBO}}$ or $\tilde{\mathcal{L}}_{\text{Mix}}$ with those using $\mathcal{L}_{\text{ELBO}}$ (\name \ w/ ELBO) or omitting the negative advantage loss entirely (\name \ wo/ neg).
Removing the negative advantage loss results in a substantial performance drop, highlighting the importance of negative advantage penalties to RL. Additionally, both Mixture and EUBO methods outperform ELBO (except for EUBO in MATH500), showcasing the benefits of evidence upper bound regularization for negative rewards. We provide complete results for each generation length in \Cref{tab:ablation_component_complete}.

\begin{wrapfigure}{r}{0.35\textwidth}
  \centering
  \includegraphics[width=0.35\textwidth]{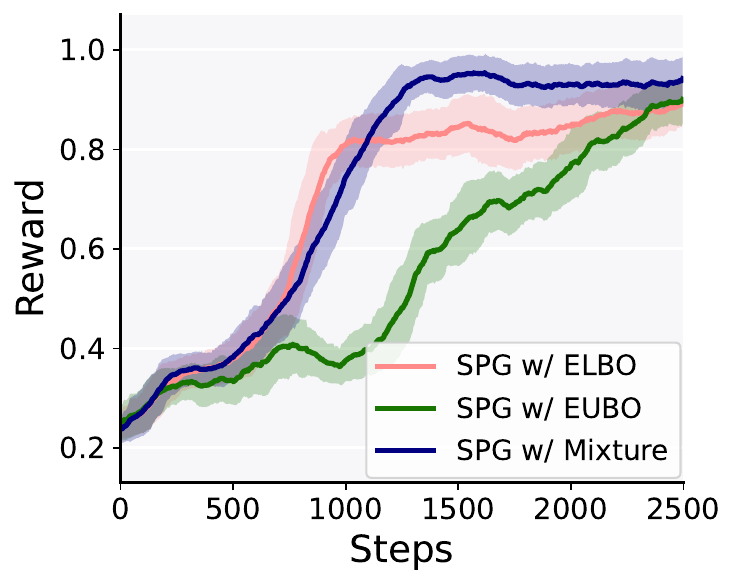}
  \vspace{-10pt}
  \caption{Reward dynamics of different log-likelihood estimation methods for negative advantage traces on Sudoku. \name \ w/ Mixture leads to both fast convergence and high rewards.}
  \vspace{-16pt}
  \label{fig:ablation_sudoku}
\end{wrapfigure}
The effect of log-likelihood estimation methods is further illustrated by the reward dynamics of each model in \Cref{fig:ablation_sudoku}, taking Sudoku as an example. \name \ w/ ELBO converges rapidly during training but plateaus early, as minimizing the lower bound does not necessarily minimize the true log-likelihood for negative advantage traces. In contrast, \name\ w/ EUBO achieves higher final rewards but converges more slowly and less stably. Combining both, \name\ w/ Mixture attains fast, stable convergence and high rewards, leading to an effective balance. This aligns with our discussions in \Cref{subsec:method.mixture}.

We also conduct ablations on the masking strategies in Monte Carlo estimation of $\mathcal{L}_{\text{ELBO}}$, $\tilde{\mathcal{L}}_{\text{EUBO}}$, and $\tilde{\mathcal{L}}_{\text{Mix}}$. As shown in \Cref{tab:ablation_mask}, the block-wise masking strategy outperforms random masking, demonstrating the importance of aligning input distributions between policy rollout and optimization. We provide complete results for each generation length in \Cref{tab:ablation_mask_complete}.

\begin{table*}[t]
\centering
\begin{minipage}[t]{0.49\textwidth}
\centering
\caption{Ablations on log-likelihood estimation methods for negative advantage traces. The best results are bolded and the second best underlined. We denote the absolute gain of test accuracy to \name\ w/ ELBO in green. \name \ w/ Mixture consistently outperforms other likelihood estimation methods.}
\label{tab:ablation_component}
\vspace{-6pt}
\adjustbox{width=\textwidth,center}{
\begin{tabular}{lc c c c}
\toprule
\textbf{Model}& \multicolumn{1}{c}{\textbf{GSM8K}} & \multicolumn{1}{c}{\textbf{MATH500}} & \multicolumn{1}{c}{\textbf{Countdown}} & \multicolumn{1}{c}{\textbf{Sudoku}}\\\midrule
\name \ wo/ neg  & 77.4 & 32.7 & 45.5 & 68.8\\
\name \ w/ ELBO & 80.9 & \underline{37.4} & 67.1 & 82.4 \\
\name \ w/ EUBO & \underline{81.6} & 36.7 & \underline{69.3} & \underline{86.1} \\
\name \ w/ Mixture & \textbf{83.1}\textcolor{ForestGreen}{\tiny{\textbf{+2.2}}} & \textbf{38.4}\textcolor{ForestGreen}{\tiny{\textbf{+1.0}}} & \textbf{69.9}\textcolor{ForestGreen}{\tiny{\textbf{+2.8}}} & \textbf{90.0}\textcolor{ForestGreen}{\tiny{\textbf{+7.6}}} \\
\bottomrule
\end{tabular}}
    \end{minipage}
\hfill
\begin{minipage}[t]{0.48\textwidth}
\centering
\captionof{table}{Ablations on the masking strategies in Monte Carlo estimation. We denote the absolute
gain of test accuracy to random masking for each model in green. Our block-wise masking strategy leads to consistent improvement to random masking on both benchmarks.}
\vspace{-3pt}
\label{tab:ablation_mask}
\adjustbox{width=\textwidth,center}{
\begin{tabular}{llc c}
\toprule
\textbf{Model} & \textbf{Masking} & \multicolumn{1}{c}{\textbf{MATH500}} & \multicolumn{1}{c}{\textbf{Countdown}}  \\
\midrule
 \name \ w/ EUBO & random & 36.7 & 45.4 \\
& block-wise & 36.7\textcolor{ForestGreen}{\tiny{\textbf{+0.0}}} & 69.3\textcolor{ForestGreen}{\tiny{\textbf{+23.9}}} \\\midrule
\name \ w/ Mixture &random &  36.9 & 62.8 \\
& block-wise & 38.4\textcolor{ForestGreen}{\tiny{\textbf{+1.5}}} & 69.9\textcolor{ForestGreen}{\tiny{\textbf{+7.1}}} \\
\bottomrule
\end{tabular}
}
\end{minipage}
\end{table*}

\paragraph{Ablations on Key Hyperparameters $\beta$ and $\omega$.} We first examine the effect of $\beta$, a crucial hyperparameter in evidence upper bound estimation, in panels (a)-(d) of \Cref{fig:ablation_beta_omega}. In general, a relatively small value of $\beta$ (i.e., close to 1.0) leads to a tighter bound and thus better performance. Nevertheless, \name \ consistently performs well across a range of $\beta$ values on most tasks, indicating its robustness. For our main results in \Cref{tab:main_results}, we fix $\omega=0.5$ and select the optimal $\beta\geq 1$, resulting in $\beta=1.0$ for Sudoku and $\beta=1.5$ for the other three benchmarks, except for Countdown with \name \ w/ EUBO where $\beta=2.0$.
Besides, since the ELBO corresponds to the case of $\beta=0$ theoretically and EUBO corresponds to $\beta \geq 1$, we also investigate intermediate values $0<\beta<1$, which may serve as an implicit mixture of lower and upper bounds. However, it is unstable in Sudoku and underperform \name{} w/ Mixture on most benchmarks. 

We also experiment on the effect of the mixture coefficient $\omega$, keeping $\beta$ fixed at its optimal value determined for $\omega=0.5$ as mentioned before. As illustrated in panels (e)-(f) of \Cref{fig:ablation_beta_omega}, combining lower and upper bounds with $\omega \in (0,1)$ leads to better performance than leveraging either bound solely, resulting in an inverted U-shaped curve. This observation is consistent with our analysis in \Cref{prop:mixture} and \Cref{subsec:method.mixture}. We provide complete ablation results of $\beta$ and $\omega$ for each generation length in \Cref{tab:ablation_beta_complete} and \Cref{tab:ablation_omega_complete}.

\paragraph{Ablations on Inference Strategies.} In the above experiments, we adopt a consistent state-of-the-art inference setup during both RL rollout and evaluation, i.e., confidence-based, block-wise semi-autoregressive generation with a block size of 32. The same configuration and block size are also used in our block-wise masking strategy. This raises the question of whether our approach generalizes well to alternative inference strategies. To assess this, we evaluate the base model and all RL fine-tuned models using various inference strategies, as shown in \Cref{fig:ablation_inference}. Despite being trained under confidence-based semi-AR decoding, \name\ consistently outperforms all baselines by a substantial margin across all inference strategies, demonstrating its robustness and strong generalizability. Complete results for each benchmark individually are provided in \Cref{tab:ablation_inference}.

\section{Conclusion}
We propose \name, a novel reinforcement learning algorithm for diffusion large language models. \name\ addresses the intractable log-likelihood in dLLMs by maximizing a tractable lower bound on positive reward sequences and minimizing an upper bound on negative ones, resulting in a more robust and less biased policy gradient. Additionally, we propose a block-wise masking strategy for Monte Carlo estimation to enhance optimization stability and efficiency.
Extensive experiments on four mathematical and logical reasoning benchmarks demonstrate the superior performance of \name, achieving significant improvement over baselines and the state-of-the-art performance.
\clearpage
\newpage

\subsection*{Acknowledgement}
CW and TJ acknowledge support from the NSF Expeditions grant (award 1918839) Understanding the World Through Code. CW is supported in part by the GQS PhD Fellowship.

\bibliography{iclr2026_conference}
\bibliographystyle{iclr2026_conference}

\newpage
\appendix
\setcounter{prop}{0}
\setcounter{theorem}{0}
\setcounter{corollary}{0}

\section{Related Work}
\paragraph{Diffusion Language Models.} Building on the remarkable success of diffusion models for image generation in continuous domains~\citep{song2020score,ho2020denoising}, researchers have explored their extension to discrete data such as text. Initial attempts focused on training continuous diffusion models in the text embedding space~\citep{diffusionlm,diffuseq,ssdlm,duo}, while they face challenges in optimization and generalization due to the discrete nature of text data. Masked diffusion models~\citep{sedd,kong_mdlm,multiflow,mdlm,shi2024simplified} address this by defining the diffusion process directly in the discrete token space, using random masking as the forward process, and have achieved strong empirical results. Block Diffusion~\citep{block_diffusion} further advances this direction by combining the strengths of autoregressive models, such as the capability to generate variable-length outputs and using KV cache to accelerate inference, with the benefits of diffusion language models like parallel decoding and flexible, any-order generation within blocks. Recently, large-scale diffusion language models trained with masked diffusion objectives have demonstrated performance competitive with similarly sized autoregressive models~\citep{llada,dream}. More recent works~\citep{fastdllm,dkvcache,dllmcache,duo,sahoo2025esoteric} have introduced caching and parallel decoding algorithms that greatly enhance the inference efficiency of dLLMs.

\paragraph{Reinforcement Learning for LLMs and Reasoning.} The seminal works apply reinforcement learning to large language models (LLMs) to align them with human preferences via reinforcement learning from human feedback (RLHF)~\citep{christiano2017deep,rlhf}. More recently, reinforcement learning has proven highly effective at enhancing the reasoning abilities of LLMs during the post-training stage, where rewards can be provided by a process reward model~\citep{math500} or verifiable reward signals. Algorithms such as Proximal Policy Optimization (PPO) and Trust Region Policy Optimization (TRPO) constrain policy updates to a trust region, reducing variance and promoting stable learning by preventing excessive shifts from the reference policy~\citep{trpo,ppo}. Group Relative Policy Optimization (GRPO)~\citep{shao2024deepseekmath} introduces group-relative rewards, enabling efficient training without the need for an additional value (critic) model. GRPO and its variants have demonstrated strong empirical performance in state-of-the-art models such as DeepSeek-R1~\citep{deepseekr1}, particularly on mathematical reasoning tasks, where incorporating long reasoning traces with self-reflection and verification steps yields significant improvements. Recent works~\citep{drgrpo,gspo,team2025kimi,spo} further improve RL algorithms for LLMs by reducing the bias introduced by the GRPO objective, enhancing sample efficiency, and introducing additional regularization.

\paragraph{Reinforcement Learning for Diffusion Language Models.} Numerous studies have explored RL-based fine-tuning algorithms for diffusion models with continuous objectives~\citep{fan2023dpok,black2023training,clark2023directly}. While RL algorithms have achieved notable success to LLMs and continuous diffusion models, their applications to diffusion language models in the discrete space remain underexplored.
DRAKES~\citep{drakes} leverages reward backpropagation along the denoising trajectory, but is computationally intensive for large scale models as the gradients are propagated through each denoising step. Alternatively, methods like \done~\citep{d1} and UniGRPO~\cite{mmada} utilize the GRPO framework, approximating the log-likelihood through either a one-step unmasking (as in \done) or Monte Carlo estimation using the ELBO (as in UniGRPO). VRPO~\citep{llada1.5} adapts DPO~\citep{dpo} to fine-tune dLLMs by applying MC estimation of the ELBO. \wdone~\citep{tang2025wd1} starts from the GRPO formulation and the same log-likelihood estimation as in \done, while avoiding direct estimation of the old and reference policy log-likelihoods by integrating them into a weighted policy optimization objective. \rebuttal{LLaDOU~\citep{lladou} decomposes the likelihood along the diffusion steps and utilizes a trajectory-level policy update algorithm for fine-tuning, while such stepwise approach is computationally costly. It also employs a decoding order prediction module and jointly optimizes both model predictions and the order of unmasking during inference.} Despite these advances, a principled analysis of RL algorithms for dLLMs, especially the challenging log-likelihood estimation, is missing. This results in substantial bias in the optimization objective and suboptimal performance.

\section{Basics of dLLMs}
\label{app:baiscs-of-dllm}
In this section, we provide a more self-contained overview of masked dLLMs. Please also refer to \citet{mdlm} for more details. 

\paragraph{Notation.}
We denote scalars by lowercase letters ($x$), vectors by bold lowercase ($\bm{x}$), and sequences by $\bm{x}_{1:n}$. A superscript (e.g., $\bm{x}^j$) denotes an item's index within a group. We define the set of the first $k$ integers as $[k] \defeq \{1, \dots, k\}$ and the $k$-dimensional probability simplex as $\Delta^{k-1}$. Distributions include the categorical $\text{Cat}(\cdot \mid \bm{p})$ and the uniform $\mathcal{U}[a, b]$. Throughout the paper, we use the following primary indices: $i \in [n]$ for position, $j \in [g]$ for a sequence in a group, and $t \in [0, 1]$ for the continuous diffusion timestep.

We start from a discrete time version of the diffusion models with finite $t\in [T]$. Assume a one-hot categorical variable $\bm{x} \in \{e_1, \dots, e_k \} \subset \Delta^{k-1}$.
Further assume we gradually corrupt $\bm{x}$ into an absorbing state $\bm{m}$ (i.e., $e_{\mask}$) with transition matrix $\bm{Q}_t$ at time $t$. Then:
$$
q(\bm{z}_t \mid \bm{x}) = \text{Cat}(\bm{z}_t \mid \overline{\bm{Q}_t} \bm{x}) = \text{Cat}( \bm{z}_t \mid \prod_{\tau=1}^t \bm{Q}_\tau \bm{x}).
$$
Here, $\bm{z}_t$ is also a one-hot categorical random variable in $\Delta^{k-1}$. In practice, one could choose $\bm{Q}_t$ such that:
$$
q(\bm{z}_t \mid \bm{x}) = \text{Cat}(\bm{z}_t \mid \alpha_t \bm{x} + (1 - \alpha_t) \bm{m}).
$$
Here, $\alpha_1 = 1, \alpha_T = 0, \alpha'_t < 0$.

Normally, the goal is to construct the lower bound of the evidence (ELBO) and maximize it. For this particular case, consider the discretized Markov chain with $T$ latent variables $\bm{z}_1, \bm{z}_2, \dots, \bm{z}_T$, where $\bm{z}_T = \bm{m}$ and $\bm{z}_1 = \bm{x}$. We use the shorthand $\bm{z} = \bm{z}_{1:T}$ and write
\begin{equation}
\begin{split}
 \mathcal{L}_\text{ELBO}(\bm{x};\bm\th) &= \E_{\bm{z} \sim q(\cdot \mid \bm{x})} \bigg[ \log \frac{p_{\bm\th}(\bm{x}, \bm{z})}{q(\bm{z} \mid \bm{x})} \bigg] \\
=&\E_{\bm{z} \sim q(\cdot \mid \bm{x})} \bigg[  \underbrace{\log p_{\bm\th}(\bm{x}\mid \bm{z}_1)}_{=0} + \sum_{t=1}^{T-1} \log \frac{p_{\bm\th}(\bm{z}_t \mid \bm{z}_{t+1})}{q(\bm{z}_t \mid \bm{z}_{t+1}, \bm{x})} + \underbrace{\log \frac{p_{\bm\th}(\bm{z}_T)}{q(\bm{z}_T \mid \bm{x})}}_{=0}\bigg] \\
=& \sum_{t=1}^{T-1} \E_{\bm{z}_t, \bm{z}_{t+1} \sim q} \bigg[  \log \frac{p_{\bm\th}(\bm{z}_t \mid \bm{z}_{t+1})}{q(\bm{z}_t \mid \bm{z}_{t+1}, \bm{x})}\bigg] \\
=& \sum_{t=1}^{T-1} \E_{\bm{z}_{t+1} \sim q(\cdot \mid \bm{x})} \E_{\bm{z}_t \sim q(\cdot \mid \bm{z}_{t+1}, \bm{x})}\bigg[ \log \frac{p_{\bm\th}(\bm{z}_t \mid \bm{z}_{t+1})}{q(\bm{z}_t \mid \bm{z}_{t+1}, \bm{x})}\bigg].
\end{split}
\label{eq:apx-elbo-kl}
\end{equation}
Here, $\log p_{\bm\th}(\bm{x}, \bm{z}_1) = 0$ because we assume $\bm{z}_1 = \bm{x}$, and $p_{\bm\th}(\bm{z}_T) = q(\bm{z}_T \mid \bm{x})$ because we assume $\bm{z}_T = \bm{m}$. A common method to parameterize $p_{\bm\th}$ is via predicting $\bm{x}$ with model $\pi_{\bm{\theta}}$ in $q$:
$$
p_{\bm\th}(\bm{z}_t \mid \bm{z}_{t+1}) = q\left( \bm{z}_t \mid \bm{z}_{t+1}, \bm{x} = \pi_{\bm{\theta}}(\cdot\mid\bm{z}_{t+1}) \right).
$$

Now, given that $\bm{z}_{t+1}$ is either $\bm{m}$ or $\bm{x}$ (assuming $\bm{m} \neq \bm{x}$). Then the KL term in \eqref{eq:apx-elbo-kl} decomposes into the following.
\begin{equation}
\begin{split}
 \log \frac{p_{\bm\th}(\bm{z}_t \mid \bm{z}_{t+1})}{q(\bm{z}_t \mid \bm{z}_{t+1}, \bm{x})} = 
 \begin{cases}
    0 & \bm{z}_t = \bm{z}_{t+1} = \bm{x}, \\ 
    0 & \bm{z}_t = \bm{m}, \bm{z}_{t+1} = \bm{x},\qquad\lb{(Impossible)}  \\ 
    \log \pi_{\bm{\theta}}(\bm{x}\mid \bm{z}_{t+1})
    & \bm{z}_t = \bm{x}, \bm{z}_{t+1} = \bm{m}, \\ 
    0 & \bm{z}_t = \bm{z}_{t+1} = \bm{m}. \\
 \end{cases}
\end{split}
\end{equation}

Moreover, $q(\bm{z}_t = \bm{x} \mid \bm{z}_{t+1} = \bm{m}, \bm{x}) = \frac{\alpha_t - \alpha_{t+1}}{1 - \alpha_{t+1}}$, and note that $\pi_{\bm{\theta}}(\bm{x}\mid \bm{z}_{t})=1$
when $\bm{z}_t = \bm{x}$, so we have:
\begin{equation}
\begin{split}
\mathcal{L}_\text{ELBO}(\bm{x};\bm\th) &= \sum_{t=1}^{T-1} \mathbb{E}_{\bm{z}_{t+1} \sim q(\cdot \mid \bm{x})}\bigg[ \frac{\alpha_t - \alpha_{t+1}}{1 - \alpha_{t+1}} \log \pi_{\bm{\theta}}(\bm{x}\mid \bm{z}_{t+1}) \mathbbm{1}(\bm{z}_{t+1} = \bm{m})\bigg] \\
&= \sum_{t=1}^{T-1} \mathbb{E}_{\bm{z}_{t+1} \sim q(\cdot \mid \bm{x})}\bigg[ \frac{\alpha_t - \alpha_{t+1}}{1 - \alpha_{t+1}} \log\pi_{\bm{\theta}}(\bm{x}\mid \bm{z}_{t+1})\bigg].~~~\lb{(If $\bm{z}_{t+1} = \bm{x}$, then $\log \pi_{\bm{\theta}}(\bm{x}\mid \bm{z}_{t+1}) = 0$)}
\end{split}
\end{equation}

Taking the above limit as $T \rightarrow \infty$, we have:
\begin{equation}
\mathcal{L}_\text{ELBO}(\bm{x};\bm\th) = \int_{t=0}^1 \mathbb{E}_{\bm{z}_t \sim q(\cdot \mid \bm{x})}\bigg[ \frac{\alpha'_t}{\alpha_t - 1} \log \pi_{\bm{\theta}}(\bm{x}\mid \bm{z}_{t}) \bigg].
\end{equation}

\paragraph{Generalization to Sequence} The above is for a single categorical variable $\bm{x}$. In practice as in language modeling, it becomes a sequence of categorical variables $\bm{x}_{1:n}$. Then we write 
\begin{equation}
\begin{split}
 \mathcal{L}_\text{ELBO}(\bm{x}_{1:n};\bm\th) 
 &= \E_{\bm{z}_{1:n} \sim q(\cdot \mid \bm{x}_{1:n})} \bigg[ \log \frac{p_{\bm\th}(\bm{x}_{1:n}, \bm{z}_{1:n})}{q(\bm{z}_{1:n} \mid \bm{x}_{1:n})} \bigg] \\
 &= \E_{\{\bm{z}_i \sim q(\cdot \mid \bm{x}_i)\}_{i=1}^n} \bigg[ \sum_{i=1}^n \log \frac{p_{\bm\th}(\bm{x}_{i}, \bm{z}_{1:n})}{q(\bm{z}_i \mid \bm{x}_i)} \bigg]\qquad \lb{(Independence of $q(\cdot \mid \bm{x}_i)$)} \\
 &= \sum_{i=1}^n \E_{\{\bm{z}_{i'} \sim q(\cdot \mid \bm{x}_{i'})\}_{i'=1}^n} \bigg[ \log \frac{p_{\bm\th}(\bm{x}_{i}, \bm{z}_{1:n})}{q(\bm{z}_i \mid \bm{x}_i)} \bigg] \\
 &= \sum_{i=1}^n \mathcal{L}_\text{ELBO}(\bm{x}_i;\bm\th).
\end{split}
\end{equation}
The key distinction from the single-token formulation (mentioned beforehand) is that the reverse process $p_{\bm\th}$ is conditioned on all $\bm{z}_{1:n}$ instead of a single token's $\bm{z}_i$.

\section{\rebuttal{Detailed Overview on RL Algorithms for dLLMs}}
\label{app:sec:baselines_detail}
\rebuttal{In this section, we provide a comprehensive overview of the existing reinforcement learning algorithms for dLLMs that serve as baselines in \Cref{sec:experiment}. The main distinctions among these methods are highlighted in \Cref{tab:alg_comparison}, covering aspects such as the policy update algorithm, log-likelihood estimation technique, masking strategy used in likelihood estimation, and other unique components. Below, we elaborate on each of these aspects.}

\renewcommand{\cellalign}{tl}
\begin{table}[t!]
\centering
\caption{\rebuttal{Comparisons of different RL algorithms for dLLMs.}}
\label{tab:alg_comparison}
\adjustbox{width=\textwidth,center}{
\begin{tabular}{l l l l l}
\toprule
\textbf{Model} &
\textbf{Policy Update Algorithm} &
\textbf{Log-Likelihood Estimation} &
\textbf{Masking Strategy} &
\textbf{Others} \\
\midrule
LLaDA-1.5 & DPO & ELBO & Random masking & \makecell[l]{Variance-reduced sampling\\budget allocation} \\
D1 & GRPO & One-step estimation & -- & Prompt masking \\
WD1 & \makecell[l]{Weighted policy optimization} & One-step estimation & -- & Prompt masking \\
UniGRPO & GRPO & ELBO & Random masking & Structured noising strategy \\
LLaDOU & \makecell[l]{Trajectory-level policy update} & Stepwise probability & -- & \makecell[l]{Decoding order prediction} \\
SPG & Policy gradient & \makecell[l]{ELBO for positive traces;\\EUBO/Mixture for negative traces} & Block-wise masking & Prompt masking \\
\bottomrule
\end{tabular}
}
\end{table}

\rebuttal{LLaDA-1.5~\citep{llada1.5} utilizes the variance-reduced preference optimization (VRPO) algorithm to align the masked diffusion model with preference pairs data. It takes the DPO~\citep{dpo} algorithm and uses ELBO with random masking to estimate the likelihood terms in the DPO objective. Specifically, they analyzes the variance of the ELBO estimators and propose an optimal Monte Carlo budge allocation in the ELBO estimation.}

\rebuttal{\done~\citep{d1} and UniGRPO~\citep{mmada} employs the GRPO~\citep{shao2024deepseekmath} objective. \done uses the one-step prediction to estimate each likelihood term in the GRPO objective, which predicts the the clean response sequence based on the partially masked prompt and the fully masked response. UniGRPO alternatively utilizes ELBO with random masking to estimate each likelihood terms. Moreover, UniGRPO takes a structured noising strategy which uniformly increases the noise level during the gradient update iterations.}

\rebuttal{\wdone~\citep{tang2025wd1} reformulates the GRPO objective while integrates the old and reference policy likelihoods into the policy rollout distribution. Then it leverages a single weighted policy optimization objective with the weights defined with the group relative advantages. It applies the same log-likelihood estimation method as \done, using the one-step prediction from partially masked prompt and the fully masked response.}

\rebuttal{Our \name{} algorithm utilizes a simple the policy gradient objective with group-relative advantage, and estimates the log-likelihood term with ELBO for positive traces while using EUBO or the mixture of ELBO and EUBO for negative ones. We employ a block-wise masking strategy to generate the perturbed samples in likelihood estimation, and apply the same prompt masking trick as in \done.}

\rebuttal{In addition to optimizing the diffusion policy $\pi_{\bm{\theta}}(\bm{x}|\bm{c})$ itself, LLaDOU~\citep{lladou} incorporates a decoding order prediction module that learns the optimal unmasking order during inference. It decomposes the likelihood of the dLLMs along the diffusion steps and utilizes a trajectory-level policy update algorithm which accumulates the gradient from each diffusion step and jointly optimizes both model predictions and the inference-time decoding order. While learning an optimal decoding order is a promising avenue for enhancing inference strategies, which is complementary to advancements in RL algorithms, and can be integrated with \name{}, we leave such an exploration to future work.}

\rebuttal{For the RL algorithm component, we reimplement the trajectory-level RL method from \citet{lladou} in our framework and present the results in \Cref{tab:stepwise} (denoted as “StepWise”). \name{} consistently outperforms StepWise on both the MATH500 and Countdown benchmarks. Additionally, trajectory-level updates as used in \citet{lladou} require backpropagation through every diffusion timestep, leading to high computational costs. For instance, on MATH500 with 8$\times$A100 GPUs, StepWise takes 7.5 minutes per gradient update, while SPG only takes 0.6 minutes.}

\begin{table}[!h]
\centering
\caption{\rebuttal{Comparison with the trajectory-level policy update algorithm~\citep{lladou}, i.e., StepWise, on MATH500 and Countdown.}}
\label{tab:stepwise}
\adjustbox{width=0.65\textwidth,center}{
\begin{tabular}{lcccc cccc}
\toprule
& \multicolumn{4}{c}{\textbf{MATH500 (0-shot)}} & \multicolumn{4}{c}{\textbf{Countdown (0-shot)}}  \\\cmidrule(lr){2-5} \cmidrule(lr){6-9} 
\textbf{Model / Seq Len} & \textbf{128} & \textbf{256} & \textbf{512} & \textit{Avg.} & \textbf{128} & \textbf{256} & \textbf{512} & \textit{Avg.}\\
\midrule
StepWise & 33.2 & 35.2 & 38.0 & 35.5 & 42.8 & 42.4 & 53.2 & 46.1 \\\midrule
\name{} w/ EUBO & 33.2 & 37.6 & 39.4 & 36.7 & 68.4 & 71.5 & 68.0 & 69.3 \\
\name{} w/ Mixture & 33.4 & 40.0 & 41.8 & 38.4 & 68.8 & 70.7 & 70.3 & 69.9\\
\bottomrule
\end{tabular}
}
\end{table}

\section{Evidence Upper Bound for dLLMs}
\label{app:theory}
In this section, we provide the derivation of the evidence upper bound. Following the above section, we start from the discrete time version of the diffusion models.

\begin{lemma}[R\'enyi Variational Bound; \citet{renyi1961measures,van2014renyi}] Fix an observation $x$. Let $q(\cdot \mid x)$ be any distribution on $\mathcal{Z}$ such that $p(\cdot \mid x) \ll q(\cdot \mid x)$, denoting that $p(\cdot \mid x)$ is absolutely continuous with respect to $q(\cdot \mid z)$. Then, the following holds for any $\beta \geq 1$:
\begin{equation}
\E_{z \sim q(\cdot \mid x)} \left[ \log \frac{p(x, z)}{q(z \mid x)} \right] \leq \log p(x) \leq \frac{1}{\beta} \log \E_{z \sim q(\cdot \mid x)} \left[ \bigg(\frac{p(x, z)}{q(z \mid x)}\bigg)^\beta \right].
\end{equation}
\label{lem:renyi}
\end{lemma}

In view of the above lemma, we derive an evidence upper bound for masked diffusion models in the following theorem.

\begin{theorem}[Evidence Upper Bound for Masked Diffusion] Assume the forward denoising process has $T$ steps with a monotonic schedule $\alpha_t$. 
For any $\beta\geq 1$ and a sequence of categorical variables $\bm{x}_{1:n}$, we have:
\begin{equation}
    \log \pi_{\th}(\bm{x}_{1:n}) \leq \mathcal{L}_{\text{EUBO}}(\bm{x}_{1:n};\bm\th), 
\end{equation}
where 
\begin{equation}
    \mathcal{L}_{\text{EUBO}}(\bm{x}_{1:n};\bm\th) \defeq \frac{1}{\beta}\sum_{i=1}^n \log \sum_{t=1}^{T-1}\mathbb{E}_{\bm{z}_{t+1}} \Bigg[ \frac{\alpha_t-\alpha_{t+1}}{1-\alpha_{t+1}} \cdot \mathbbm{1}(\bm{z}_{t+1,i} = \bm{m}) \cdot \pi_{\th}^{\beta}(\bm{x}_i \mid \bm{z}_{t+1}) \Bigg]+C(T),
\end{equation}

and $C(T) = \begin{cases} \frac{1}{\beta}\log \E_{\bm{z}_{1:T} \sim q(\cdot \mid \bm{x})} \Big[q(\bm{z}_{1:T}\mid\bm{x})^{-n}\Big], & 1 \leq \beta < n \\ 0, & \beta \geq n \end{cases}$ is a constant independent of $\bm\th$.

\end{theorem}

\begin{proof}
We first consider the case with a single categorical variable $\bm{x}$. On the account of Lemma~\ref{lem:renyi} and following a similar argument as in \eqref{eq:apx-elbo-kl}, for any $\beta \geq 1$, we can write
\begin{align}\nonumber
    \log \pi_\th(\bm{x}) & \leq  \frac{1}{\beta} \log \E_{\bm{z} \sim q(\cdot \mid \bm{x})} \left[ \bigg(\frac{p_{\bm\th}(\bm{x}, \bm{z})}{q(\bm{z} \mid \bm{x})}\bigg)^\beta \right] \\ \label{eq:chain_rule}
    & \quad =  \frac{1}{\beta} \log \E_{\bm{z}_{1:T} \sim q(\cdot \mid \bm{x})} \bigg[ \prod_{t=1}^{T-1} \bigg(\frac{p_{\bm{\theta}}(\bm{z}_t \mid \bm{z}_{t+1})}{q(\bm{z}_t \mid \bm{z}_{t+1}, \bm{x})}\bigg)^\beta \bigg]
\end{align}
Note that the sequence $\bm{z}_{1:T}$ has a form $\{\bm{x}, \dots, \bm{x}, \bm{m}, \dots, \bm{m}\}$. Define the transition event:
\begin{align}\label{eq:transition_event}
    \mathcal{A}_t \coloneqq \{\bm{z}_t = \bm{x}, \bm{z}_{t+1} = \bm{m} \}
\end{align}
Then, by the law of total expectations, \eqref{eq:chain_rule} can be expressed as:
\begin{align}
    & \frac{1}{\beta} \log \E_{\bm{z}_{1:T} \sim q(\cdot \mid \bm{x})} \bigg[ \prod_{t=1}^{T-1} \bigg(\frac{p_{\bm\th}(\bm{z}_t \mid \bm{z}_{t+1})}{q(\bm{z}_t \mid \bm{z}_{t+1}, \bm{x})}\bigg)^\beta \bigg]\nonumber\\
    = & \frac{1}{\beta} \log \sum_{t=1}^{T-1} \mathbb{P}(\mathcal{A}_t) \mathbb{E}_{\bm{z}\sim q(\cdot\mid \bm{x})} \left[\prod_{s=1}^{T-1} \bigg(\frac{p_{\bm\th}(\bm{z}_s \mid \bm{z}_{s+1})}{q(\bm{z}_s \mid \bm{z}_{s+1}, \bm{x})}\bigg)^\beta \middle| \mathcal{A}_t \right]\nonumber\\
    = & \frac{1}{\beta} \log \sum_{t=1}^{T-1} \E_{\bm{z}_{t+1} \sim q(\cdot \mid \bm{x})} \bigg[\mathbbm{1}(\bm{z}_{t+1} = \bm{m}) q(\bm{z}_t = \bm{x} \mid \bm{z}_{t+1} = \bm{m}, \bm{x}) \bigg(\frac{p_{\bm\th}(\bm{z}_t = \bm{x} \mid \bm{z}_{t+1} = \bm{m})}{q(\bm{z}_t = \bm{x} \mid \bm{z}_{t+1} = \bm{m}, \bm{x})}\bigg)^\beta \bigg]\nonumber \\
    = & \frac{1}{\beta} \log \sum_{t=1}^{T-1} \E_{\bm{z}_{t+1} \sim q(\cdot \mid \bm{x})} \bigg[ \mathbbm{1}(\bm{z}_{t+1} = \bm{m}) \frac{\alpha_t - \alpha_{t+1}}{1 - \alpha_{t+1}} \pi_{\th}^{\beta}(\bm{x} \mid \bm{z}_{t+1})
    \bigg]\label{eq:token_eubo}
\end{align}
The penultimate line is due to the fact that conditioned on the event $\mathcal{A}_t$, the ratio $\frac{p_{\bm\th}(\bm{z}_s \mid \bm{z}_{s+1})}{q(\bm{z}_s \mid \bm{z}_{s+1}, \bm{x})}$ is equal to one for any $s \neq t$. The last line uses the formula for $q$. The indicator $\mathbbm{1}(\bm{z}_t = \bm{m})$ appears in the final expression because the terms in the bound are only non-trivial when the model must make a prediction from a corrupted state.

Now we generalize the above to a sequence of categorical variables $\bm{x}=\bm{x}_{1:n}$. Similar as \Cref{eq:chain_rule}, we have
\begin{align*}
\log \pi_{\bm\th}(\bm{x}_{1:n}) \leq 
    \frac{1}{\beta} \log \E_{\bm{z}_{1:T} \sim q(\cdot \mid \bm{x})} \bigg[ \prod_{t=1}^{T-1} \prod_{i=1}^n \bigg(\frac{p_{\bm\th}(\bm{z}_{t,i} \mid \bm{z}_{t+1})}{q(\bm{z}_{t,i} \mid \bm{z}_{t+1}, \bm{x})}\bigg)^\beta \bigg] 
\end{align*}

Denote $\tilde{C}(T)\defeq \frac{1}{\beta}\log \E_{\bm{z}_{1:T} \sim q(\cdot \mid \bm{x})} \Big[q(\bm{z}_{1:T}\mid\bm{x})^{-n}\Big]$. The upper bound in the RHS can be further derived as
\begin{align}
    &\frac{1}{\beta} \log \E_{\bm{z}_{1:T} \sim q(\cdot \mid \bm{x})} \bigg[ \prod_{t=1}^{T-1} \prod_{i=1}^n \bigg(\frac{p_{\bm\th}(\bm{z}_{t,i} \mid \bm{z}_{t+1})}{q(\bm{z}_{t,i} \mid \bm{z}_{t+1}, \bm{x})}\bigg)^\beta \bigg] \nonumber
    \\ =&  \frac{1}{\beta} \log \E_{\bm{z}_{1:T} \sim q(\cdot \mid \bm{x})} \bigg[q(\bm{z}_{1:T}\mid\bm{x})^{-n}\prod_{i=1}^n \sum_{\bm{y}_{1:T}^i} q(\bm{y}_{1:T}^i\mid\bm{x}) \mathbbm{1}(\bm{y}_{1:T}^i=\bm{z}_{1:T}) \prod_{t=1}^{T-1} \bigg(\frac{p_{\bm\theta}(\bm{y}_{t,i}^i \mid \bm{y}_{t+1}^i}{q(\bm{y}_{t,i}^i \mid \bm{y}_{t+1}^i, \bm{x})}\bigg)^\beta \bigg] \nonumber
    \\ \leq&\frac{1}{\beta} \log \E_{\bm{z}_{1:T} \sim q(\cdot \mid \bm{x})} \bigg[q(\bm{z}_{1:T}\mid\bm{x})^{-n}\prod_{i=1}^n \sum_{\bm{y}_{1:T}^i} q(\bm{y}_{1:T}^i\mid\bm{x}) \prod_{t=1}^{T-1} \bigg(\frac{p_{\bm\theta}(\bm{y}_{t,i}^i \mid \bm{y}_{t+1}^i}{q(\bm{y}_{t,i}^i \mid \bm{y}_{t+1}^i, \bm{x})}\bigg)^\beta \bigg] \nonumber
    \\=& \frac{1}{\beta} \log\Bigg( \E_{\bm{z}_{1:T} \sim q(\cdot \mid \bm{x})} \Big[q(\bm{z}_{1:T}\mid\bm{x})^{-n}\Big]\cdot\bigg(\prod_{i=1}^n \sum_{\bm{y}_{1:T}^i} q(\bm{y}_{1:T}^i\mid\bm{x}) \prod_{t=1}^{T-1} \bigg(\frac{p_{\bm\theta}(\bm{y}_{t,i}^i \mid \bm{y}_{t+1}^i}{q(\bm{y}_{t,i}^i \mid \bm{y}_{t+1}^i, \bm{x})}\bigg)^\beta \bigg)\Bigg) \nonumber
    \\=& \frac{1}{\beta} \log \prod_{i=1}^n \E_{\bm{z}_{1:T} \sim q(\cdot \mid \bm{x})} \bigg[ \prod_{t=1}^{T-1} \bigg(\frac{p_{\bm{\theta}}(\bm{z}_{t,i} \mid \bm{z}_{t+1})}{q(\bm{z}_{t,i} \mid \bm{z}_{t+1}, \bm{x})}\bigg)^\beta \bigg] + \frac{1}{\beta} \log\E_{\bm{z}_{1:T} \sim q(\cdot \mid \bm{x})} \Big[q(\bm{z}_{1:T}\mid\bm{x})^{-n}\Big]\nonumber
    \\=&\frac{1}{\beta} \sum_{i=1}^n\log  \E_{\bm{z}_{1:T} \sim q(\cdot \mid \bm{x})} \bigg[ \prod_{t=1}^{T-1} \bigg(\frac{p_{\bm{\theta}}(\bm{z}_{t,i} \mid \bm{z}_{t+1})}{q(\bm{z}_{t,i} \mid \bm{z}_{t+1}, \bm{x})}\bigg)^\beta \bigg] + \tilde{C}(T)
    \label{eq:seq_eubo}
\end{align}
Here, $\bm{y}_{1:T}^i$ are copies of $\bm{z}_{1:T}$ enforced to agree with $\bm{z}_{1:T}$ using the indicator $\mathbbm{1}(\bm{y}_{1:T}^i=\bm{z}_{1:T})$. $C(T)$ is a constant independent of $\bm\th$, and the first term in \Cref{eq:seq_eubo} can be derived similar to the single variable case in \Cref{eq:token_eubo}:
\begin{align*}
    &\frac{1}{\beta} \sum_{i=1}^n\log  \E_{\bm{z}_{1:T} \sim q(\cdot \mid \bm{x})} \bigg[ \prod_{t=1}^{T-1} \bigg(\frac{p_{\bm{\theta}}(\bm{z}_{t,i} \mid \bm{z}_{t+1})}{q(\bm{z}_{t,i} \mid \bm{z}_{t+1}, \bm{x})}\bigg)^\beta \bigg]
    \\=&\frac{1}{\beta}\sum_{i=1}^n \log \sum_{t=1}^{T-1}\mathbb{E}_{\bm{z}_{t+1}\sim q(\cdot\mid \bm{x})} \Bigg[ \frac{\alpha_t-\alpha_{t+1}}{1-\alpha_{t+1}} \cdot \mathbbm{1}(\bm{z}_{t+1,i} = \bm{m}) \cdot \pi_{\th}^{\beta}(\bm{x}_i \mid \bm{z}_{t+1}) \Bigg]
\end{align*}

Additionally, for $\beta \geq n$, we can obtain a tighter bound using H\"{o}lder's inequality:

\begin{align}
    &\frac{1}{\beta} \log \E_{\bm{z}_{1:T} \sim q(\cdot \mid \bm{x})} \bigg[ \prod_{t=1}^{T-1} \prod_{i=1}^n \bigg(\frac{p_{\bm\th}(\bm{z}_{t,i} \mid \bm{z}_{t+1})}{q(\bm{z}_{t,i} \mid \bm{z}_{t+1}, \bm{x})}\bigg)^\beta \bigg] \nonumber
    \\=& \frac{1}{n\beta} \sum_{i=1}^n \log \E_{\bm{z}_{1:T} \sim q(\cdot \mid \bm{x})} \bigg[ \prod_{t=1}^{T-1} \bigg(\frac{p_{\bm\th}(\bm{z}_{t,i} \mid \bm{z}_{t+1})}{q(\bm{z}_{t,i} \mid \bm{z}_{t+1}, \bm{x})}\bigg)^{n\beta} \bigg] \nonumber
    \\=&\frac{1}{n\beta}\sum_{i=1}^n \log \sum_{t=1}^{T-1}\mathbb{E}_{\bm{z}_{t+1}\sim q(\cdot\mid \bm{x})} \Bigg[ \frac{\alpha_t-\alpha_{t+1}}{1-\alpha_{t+1}} \cdot \mathbbm{1}(\bm{z}_{t+1,i} = \bm{m}) \cdot \pi_{\th}^{n\beta}(\bm{x}_i \mid \bm{z}_{t+1}) \Bigg]
\label{eq:holders}
\end{align}

Denote $\tilde{\beta}\defeq n\beta$. Then the upper bound $\log \pi_{\th}(\bm{x}_{1:n}) \leq \mathcal{L}_{\text{EUBO}}(\bm{x}_{1:n};\bm\th)$ holds with the constant term $C(T)$ being zero and the hyperparameter $\tilde{\beta}\geq n$. Combining these two cases, we have $C(T)=\begin{cases} \tilde{C}(T), & 1 \leq \beta < n \\ 0, & \beta \geq n \end{cases}$.

\end{proof}

Furthermore, we can derive the continuous time version by omitting the constant term that does not affect the gradient with respect to $\bm\th$, and taking the limit of $T \rightarrow \infty$ similar as the derivations for $\mathcal{L}_{\text{ELBO}}$, as shown in \Cref{coro:eubo}:

\begin{corollary}
Taking the limit of $T \rightarrow \infty$, we have:
\begin{equation}
\begin{split}
&\nabla_{\bm\th} \mathcal{L}_{\text{EUBO}}(\bm{x}_{1:n};\bm\th) = \nabla_{\bm\th}\Big(\tilde{\mathcal{L}}_{\text{EUBO}}(\bm{x}_{1:n};\bm\th) +C(T)\Big) = \nabla_{\bm\th}\tilde{\mathcal{L}}_{\text{EUBO}}(\bm{x}_{1:n};\bm\th),~~~~\text{where} \\
&\tilde{\mathcal{L}}_\text{EUBO}(\bm{x}_{1:n}; \th) = \frac{1}{\beta} \sum_{i=1}^n \log \mathbb{E}_{t, \bm{z}_t} \left[ 
w(t) \cdot \mathbbm{1}(\bm{z}_{t, i} = \bm{m} ) \cdot \pi^\beta_{\bm{\theta}}(\bm{x}_i \mid \bm{z}_t)
\right].
\end{split}
\end{equation}
\end{corollary}

One caveat of the above $\tilde{\mathcal{L}}_{\text{EUBO}}$ is that the $\log$ is outside of the expectation, which in general makes Monte Carlo sample estimates biased. One could certainly further loosen the bound using the inequality $\log x \leq x - 1$:
\begin{align}
\label{eq:loose}
\mathcal{L}_{\text{EUBO}}(\bm{x}) \leq \frac{1}{\beta} \sum_{i=1}^n\E_{t \sim \mathcal{U}[0, 1], \bm{z}_t \sim q} \Bigg[ w(t) \cdot \mathbbm{1}(\bm{z}_{t,i} = \bm{m}) \cdot \pi_{\bm\th}^{\beta}(\bm{x}_i\mid \bm{z}_t)  \Bigg] - \frac{n}{\beta}
\end{align}

But in practice we found this results in much worse performance, as demonstrated in \Cref{tab:ablation_loose}, potentially due to the much larger gap between EUBO and likelihood.

\section{Additional Analysis on Upper and Lower Bounds}

\subsection{Proof of Proposition \ref{prop:mixture}}\label{app:proof}
\begin{prop}[Optimal Mixture Strictly Reduces Variance]
Fix a coordinate $k$ and let $\rho_\beta \coloneqq {w(t, \bm{z}_t) \pi_{\bm{\theta}}^\beta(\bm{x}_i\mid\bm{z}_t,\bm{c}) }/{\E\left[ w(t, \bm{z}_t) \pi_{\bm{\theta}}^\beta(\bm{x}_i\mid\bm{z}_t,\bm{c})  \right]}$, where $w(t, \bm{z}_t) \coloneqq w(t) \mathbbm{1}(z_{t} = \bm{m})$. Then, the gradient of mixture objective (\ref{eq:mixture}) is given by 
\begin{align}
    g_{\omega,k}= ((1-\omega)w(t, \bm{z}_t)+\omega\rho_\beta)\,\partial_{\bm{\theta}_k} \log \pi_{\bm\theta}(\bm{x} \mid \bm{z}_t, \bm{c}).
\end{align}
If $\operatorname{Var}((\rho_\beta-w(t, \bm{z}_t))\partial_{\bm\theta_k} \log \pi_{\bm\theta}(\bm{x} \mid \bm{z}_t, \bm{c}))>0$, then $\operatorname{Var}[g_{\omega,k}]$ is a strictly convex quadratic in $\omega$ and thus admits a unique minimizer $\omega_k^\star$. Moreover,
\[
\operatorname{Var}[g_{\omega_k^\star,k}] \;<\; \min\!\big\{\operatorname{Var}[g_{0,k}],\ \operatorname{Var}[g_{1,k}]\big\},
\]
\end{prop}
\begin{proof}
We first derive the formulas for the gradient of each objective. Consider a specific example $x_i$. 
The gradient of the $\mathcal{L}_{\text{ELBO}}$ and $\tilde{\mathcal{L}}_{\text{ELBO}}$ are given by:
\begin{align}
    \nabla_{\bm\theta} \mathcal{L}_{\text{ELBO}} & = \E \left[ w(t, \bm{z}_t) \nabla \log \pi_{\bm\theta}(\bm{x}_i\mid\bm{z}_t,\bm{c}) \right]\label{eq:grad_elbo}\\
    \nabla_{\bm\theta} \tilde{\mathcal{L}}_{\text{EUBO}} & = \frac{\E \left[ w(t, \bm{z}_t) \pi_{\bm\theta}^\beta(\bm{x}_i\mid\bm{z}_t,\bm{c}) \nabla \log \pi_{\bm\theta }(\bm{x}_i\mid\bm{z}_t,\bm{c}) \right]}{\E\left[ w(t, \bm{z}_t) \pi_{\bm\theta}^\beta(\bm{x}_i\mid\bm{z}_t,\bm{c})  \right]}\label{eq:grad_eubo}
\end{align}
Then the gradient of the mixture objective $\tilde{\mathcal{L}}_{\text{Mix}}$ is given by:
\begin{align}
    \nabla_{\bm\theta} \tilde{\mathcal{L}}_{\text{Mix}} = \E \left[\Big((1-\omega)w(t, \bm{z}_t) + \omega \rho_\beta \Big) \nabla_{\bm\theta} \log \pi_{\bm\theta} (\bm{x}_i\mid\bm{z}_t,\bm{c})\right]\label{eq:grad_mix}
\end{align}

We further compute the per-parameter (per-dimension) variance of the gradient of $\tilde{\mathcal{L}}_{\text{Mix}}$ and consider the optimal mixture coefficient $\omega$ to minimize the variance. For simplicity, we use the following short-hand notation:
\begin{align*}
    s_k & \coloneqq \partial_{\bm\theta_k} \log \pi_{\bm\theta}(\bm{x}_i \mid \bm{z}_t, \bm{c})
\end{align*}
We denote the $k$-th coordinate of the gradient $\nabla_{\bm\theta} \tilde{\mathcal{L}}_{\text{Mix}}$ by $g_{\omega, k}$. Then, the coordinate-wise variance of the gradient is given by
\begin{align*}
    \operatorname{Var}\!\big[g_{\omega,k}\big]
& =
\mathbb{E}\!\Big[\big((1-\omega)\,w+\omega\,\rho_\beta\big)^2\,s_k^{2}\Big]
-
\Big(\mathbb{E}\!\big[\big((1-\omega)\,w+\omega\,\rho_\beta\big)\,s_k\big]\Big)^2\\
& = \operatorname{Var}(w s_k) + 2 \omega \operatorname{Cov} (w s_k, (\rho_\beta - w) s_k) + \omega^2 \operatorname{Var}((\rho_\beta - w) s_k)
\end{align*}
where we used the shorthand $w \equiv w(t, z_t)$. The above expression is quadratic in $\omega$ and we find the optimal $\omega$ by setting the derivative of variance to zero:
\begin{align*}
    & \frac{\partial}{\partial \omega}\operatorname{Var}\!\big[g_{\omega,k}\big]
=
2\,\operatorname{Cov}\!\big(w\,s_k,\;(\rho_\beta-w)\,s_k\big)
+
2\omega\,\operatorname{Var}\!\big((\rho_\beta-w)\,s_k\big)
= 0 \\
\Rightarrow & \; \omega_k^\star
=
-\,\frac{\operatorname{Cov}\!\big(w\,s_k,\;(\rho_\beta-w)\,s_k\big)}
{\operatorname{Var}\!\big((\rho_\beta-w)\,s_k\big)} .
\end{align*}
The above yields a per-coordinate optimal $\omega^\star_k$. Equivalently, we can write $\omega^\star_k$ as follows:
\begin{align*}
\omega_k^\star
= \frac{\operatorname{Var}\!\big(w\,s_k\big)-\operatorname{Cov}\!\big(w\,s_k,\rho_\beta\,s_k\big)}
{\operatorname{Var}\!\big(w\,s_k\big)+\operatorname{Var}\!\big(\rho_\beta\,s_k\big)-2\,\operatorname{Cov}\!\big(w\,s_k,\rho_\beta\,s_k\big)}
\end{align*}
Furthermore, $\omega^\star_k$ is a minimizer of coordinate-wise variance in the non-degenerative case with $\operatorname{Var}\!\big((\rho_\beta-w)\,s_k\big) > 0$, as the variance is strongly convex in $\omega$. 

The coordinate-wise variance of gradients in $\mathcal{L}_{\text{ELBO}}$ ($\omega = 0$) and $\tilde{\mathcal{L}}_{\text{ELBO}}$ ($\omega = 1$), and the optimal mixture coefficient $\omega^\star$ are then given by
\begin{align*}
    \mathcal{L}_{\text{ELBO}}: \quad \operatorname{Var}\!\big[g_{0,k}\big]
& = \operatorname{Var}\!\big[w\,s_k\big],\\
\tilde{\mathcal{L}}_{\text{ELBO}}: \quad \operatorname{Var}\!\big[g_{1,k}\big]
& = \operatorname{Var}\!\big[w\,s_k\big]
+ 2\,\operatorname{Cov}\!\big(w\,s_k,\;(\rho_\beta-w)\,s_k\big)
+ \operatorname{Var}\!\big((\rho_\beta-w)\,s_k\big),\\ 
\text{Optimal:} \; \; \operatorname{Var}\!\big[g_{\omega_k^\star,k}\big]
& = \operatorname{Var}\!\big[w\,s_k\big]
- \frac{\Big(\operatorname{Cov}\!\big(w\,s_k,\;(\rho_\beta-w)\,s_k\big)\Big)^{2}}
{\operatorname{Var}\!\big((\rho_\beta-w)\,s_k\big)}\,,
\end{align*}
The difference between the variance of $\mathcal{L}_{\text{ELBO}}$ and $\tilde{\mathcal{L}}_{\text{ELBO}}$ with the optimal mixture coefficient can then be derived as follows:
\begin{align*}
\operatorname{Var}\!\big[w\,s_k\big] - \operatorname{Var}\!\big[g_{\omega_k^\star,k}\big]
&= \frac{\Big(\operatorname{Cov}\!\big(w\,s_k,\;(\rho_\beta-w)\,s_k\big)\Big)^{2}}
         {\operatorname{Var}\!\big((\rho_\beta-w)\,s_k\big)} \geq 0 \\
\operatorname{Var}\!\big[\rho_\beta\,s_k\big] - \operatorname{Var}\!\big[g_{\omega_k^\star,k}\big]
&= \frac{\Big(\operatorname{Cov}\!\big(w\,s_k,\;(\rho_\beta-w)\,s_k\big)
           + \operatorname{Var}\!\big((\rho_\beta-w)\,s_k\big)\Big)^{2}}
         {\operatorname{Var}\!\big((\rho_\beta-w)\,s_k\big)} \geq 0
\end{align*}
\end{proof}

\subsection{Additional Comparison Between the Mixture Loss and the Lower and Upper Bounds}\label{app:discussion_mixture}

\paragraph{Comparing Mixture with the Lower Bound.} Consider the ratio of the coefficient of score function $\nabla_{\bm\theta} \log \pi_{\bm\theta} (\bm{x}_i\mid\bm{z}_t,\bm{c})$ in the gradient in the case of the mixture objective (i.e., $\nabla_{\bm\th}\tilde{\mathcal{L}}_{\text{Mix}}$ in \Cref{eq:grad_mix}) over using only the lower bound (i.e., $\nabla_{\bm\th}\mathcal{L}_{\text{ELBO}}$ in \Cref{eq:grad_elbo}):
\begin{align*}
    \frac{w_{\text{Mix}}}{w_{\text{ELBO}}} = \frac{(1-\omega)w(t, \bm{z}_t) + \omega \rho_\beta }{w(t, \bm{z}_t)} = (1-\omega) + \omega \frac{\pi_{\bm\theta}^\beta(\bm{x}_i\mid\bm{z}_t,\bm{c}) }{\E\left[ w(t, \bm{z}_t) \pi_{\bm\theta}^\beta(\bm{x}_i\mid\bm{z}_t,\bm{c})  \right]}
\end{align*}
Treating the expectation over all samples $\E\left[ w(t, \bm{z}_t) \pi_{\bm\theta}^\beta(\bm{x}_i\mid\bm{z}_t,\bm{c})  \right]$ as a constant (since it is averaged), the second term in the above ratio is strictly increasing in $\pi_{\bm\theta}^\beta(\bm{x}_i\mid\bm{z}_t,\bm{c})$. This realizes a \textit{confidence-aware weighting}: uncertain tokens with small $\pi_{\bm\theta}^\beta(\bm{x}_i\mid\bm{z}_t,\bm{c})$, i.e., those with a low recovery chance, have a smaller weight, while confident tokens with large $\pi_{\bm\theta}^\beta(\bm{x}_i\mid\bm{z}_t,\bm{c})$ are upweighted, with sharpness being controlled by parameter $\beta$ and the blend by $\omega$.

\paragraph{Comparing Mixture with the Upper Bound.} We compute the ratio of coefficient of score function in the gradient of upper bound (i.e., $\nabla_{\bm\th}\tilde{\mathcal{L}}_{\text{EUBO}}$ in \Cref{eq:grad_eubo}) over the mixture gradient:
\begin{align*}
    \frac{w_{\text{EUBO}}}{w_{\text{Mix}}} = \frac{\omega \rho_\beta}{(1-\omega)w(t, \bm{z}_t) + \omega \rho_\beta}
\end{align*}
Considering the above ratio, when $\pi_{\bm\theta}^\beta(\bm{x}_i\mid\bm{z}_t,\bm{c})$ is very small, the coefficient of score function in $\nabla_{\bm\th}\tilde{\mathcal{L}}_{\text{EUBO}}$, $w_{\text{EUBO}}$, becomes very small, preventing updates to the parameters. However, the mixing approach maintains per-sample weights by preventing that from collapsing to (near) zero. In other words, for each sample, the mixture coefficient computes a convex interpolation that simultaneously floors very small EUBO weights to a minimum value and applies an uncertainty-aware capping to large EUBO weights.

\begin{figure}
    \centering
    \includegraphics[width=\linewidth]{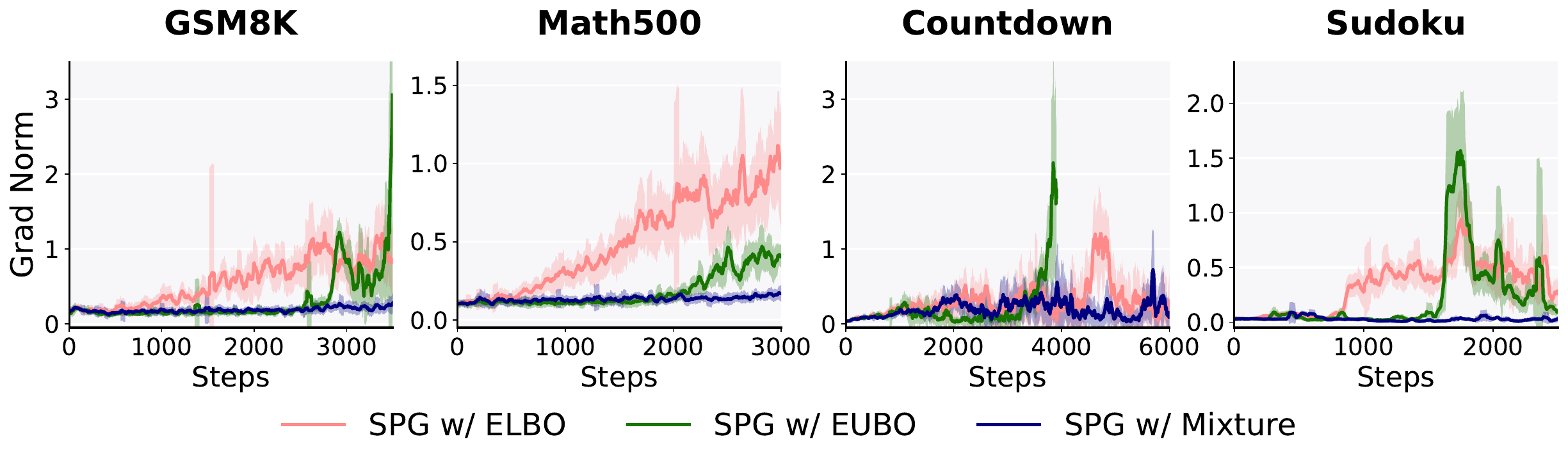}
    \caption{Dynamics of the gradient norm of models trained with different log-likelihood estimation methods. \name \ w/ Mixture achieves lower gradient norm and more stable optimization. We report mean and standard deviation over a rolling window of 50 steps.}
    \label{fig:grad_norm}
\end{figure}

\paragraph{Empirical Evidence of Reduced Gradient Variance.} As a practical indicator of gradient variance, we plot the gradient norm of each model trained with different log-likelihood estimation methods for negative advantage traces in \Cref{fig:grad_norm}. When using the mixture objective, the model has consistently smaller and more stable gradient norm throughout training,  aligning well with our theoretical analysis.

\subsection{Toy Example for Upper and Lower Bounds.} 
\label{app:toy}
In this section, we provide a toy example highlighting the contrasting behaviors and landscapes of the upper and lower bounds, further demonstrating the necessity to select the appropriate bound for optimization based on the optimization direction.

Consider a simple case where the sequence length is 2 and the vocabulary size is 2, i.e., $\bm{x}=[\bm{x}_1,\bm{x}_2]$ and $\mathcal{V}=\{A,B\}$. Then, We can calculate $\mathcal{L}_{\text{ELBO}}$ and $\tilde{\mathcal{L}}_{\text{EUBO}}$ in closed form:
\begin{align}
\mathcal{L}_{\text{ELBO}}(\bm{x}=\text{AA})&= \frac{1}{2} \Big[ \log \pi_{\bm{\theta}}(\bm{x}_1=\text{A}\mid\text{MA}) + \log \pi_{\bm{\theta}}(\bm{x}_1=\text{A}\mid\text{MM}) \\& +\log \pi_{\bm{\theta}}(\bm{x}_2=\text{A}\mid\text{AM}) + \log \pi_{\bm{\theta}}(\bm{x}_2=\text{A}\mid\text{MM})\Big] \\
\tilde{\mathcal{L}}_{\text{EUBO}}(\bm{x}=\text{AA})&= \frac{1}{\beta}\log \Big( \frac{\pi_{\bm{\theta}}^{\beta}(\bm{x}_1=\text{A}\mid\text{MA})+\pi_{\bm{\theta}}^{\beta}(\bm{x}_1=\text{A}\mid\text{MM})}{2}\Big) \\&+ \frac{1}{\beta}\log \Big( \frac{\pi_{\bm{\theta}}^{\beta}(\bm{x}_2=\text{A}\mid\text{AM})+\pi_{\bm{\theta}}^{\beta}(\bm{x}_2=\text{A}\mid\text{MM})}{2}\Big)
\end{align}

For simplicity, denote $a:=\pi_{\bm{\theta}}(\bm{x}_1=\text{A}\mid\text{MA})$ and $b:=\pi_{\bm{\theta}}(\bm{x}_1=\text{A}\mid\text{MM})$, and consider the of the likelihood of the first token $\bm{x}_1$. We have 
\begin{align}
\mathcal{L}_{\text{ELBO}}(\bm{x}_1)&=\frac{1}{2} (\log a + \log b) \\
\tilde{\mathcal{L}}_{\text{EUBO}}(\bm{x}_1)&=\frac{1}{\beta}\log \Big( \frac{a^{\beta}+b^{\beta}}{2}\Big)
\end{align}

Take the partial gradient with respect to $a$ and $b$ respectively,
\begin{align}
    \frac{\partial \mathcal{L}_{\text{ELBO}}(\bm{x}_1)}{\partial a} &= \frac{1}{2a}; \ \frac{\partial \mathcal{L}_{\text{ELBO}}(\bm{x}_1)}{\partial b} = \frac{1}{2b} \\
    \frac{\partial \tilde{\mathcal{L}}_{\text{EUBO}}(\bm{x}_1)}{\partial a} &= \frac{a^{\beta-1}}{a^{\beta}+b^{\beta}}; \ \frac{\partial \tilde{\mathcal{L}}_{\text{EUBO}}(\bm{x}_1)}{\partial b} = \frac{b^{\beta-1}}{a^{\beta}+b^{\beta}}
\end{align}

\begin{figure}[h]
    \centering
    \includegraphics[width=0.48\textwidth]{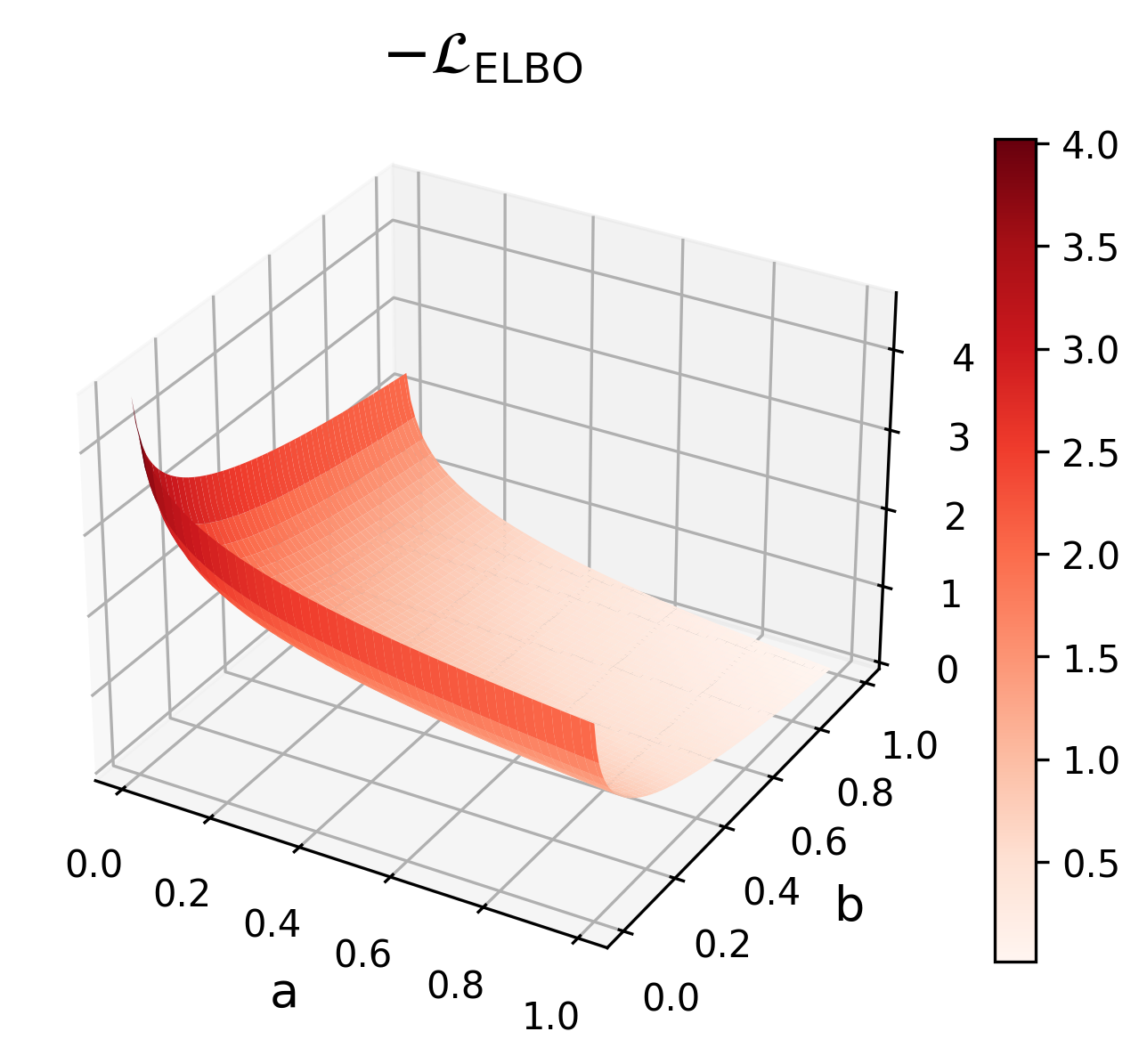}
    \hfill
    \includegraphics[width=0.48\textwidth]{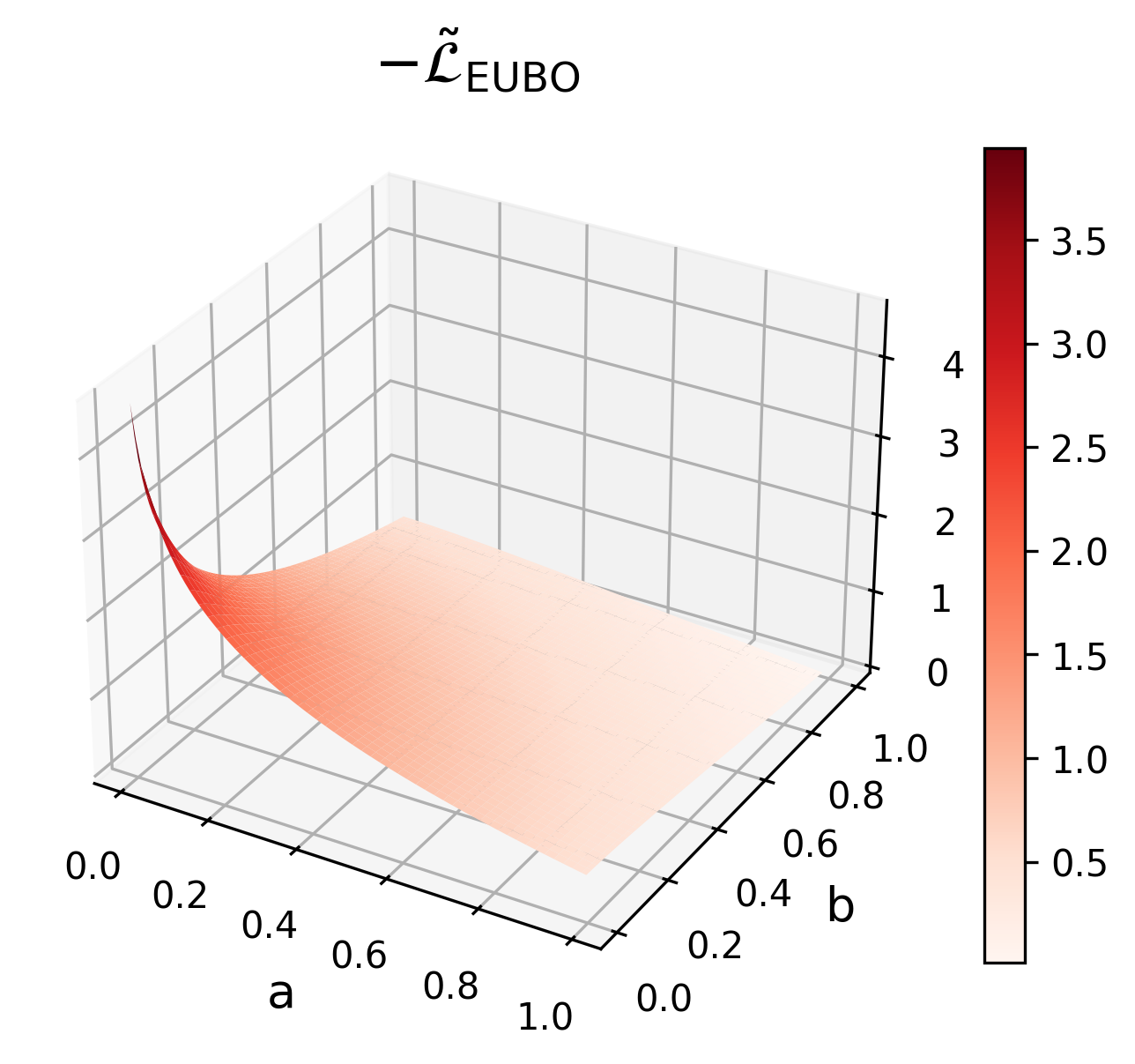}
    \caption{Landscapes of $-\mathcal{L}_{\text{ELBO}}$ and $-\tilde{\mathcal{L}}_{\text{EUBO}}$ for $0<a,b<1$. $-\tilde{\mathcal{L}}_{\text{EUBO}}$ is flatter among low value regions while sharper among high value regions, making it more suitable for log-likelihood minimization; vice versa for $-\mathcal{L}_{\text{ELBO}}$.}
    \label{fig:toy_exmple}
\end{figure}

Therefore, for $\tilde{\mathcal{L}}_{\text{EUBO}}$, the gradient direction is dominated by the larger one between $a$ and $b$, while for $\mathcal{L}_{\text{ELBO}}$, the gradient direction is dominated by the smaller one. Such property is illustrated in the landscapes of  $-\mathcal{L}_{\text{ELBO}}$ and $-\tilde{\mathcal{L}}_{\text{EUBO}}$ for $a,b\in(0,1)$ in \Cref{fig:toy_exmple}.

When $\bm{x}=\text{AA}$ has negative advantage, the corresponding $\mathcal{L}_{\text{ELBO}}$ and $\tilde{\mathcal{L}}_{\text{EUBO}}$ are minimized. For $\mathcal{L}_{\text{ELBO}}$, the model benefits more from further decreasing the smaller one between probabilities $a$ and $b$. In the extreme case, $\mathcal{L}_{\text{ELBO}}=-\infty$ when either $a$ or $b$ equals to zero, leaving the other term not sufficiently decreased. Instead, when using $\tilde{\mathcal{L}}_{\text{EUBO}}$ for negative advantage traces, the larger one between $a$ and $b$ is preferentially minimized, leading to a more balanced optimization that stably decreases the log-likelihood.

Similarly, when $\bm{x}=\text{AA}$ has positive advantage, the corresponding $\mathcal{L}_{\text{ELBO}}$ and $\tilde{\mathcal{L}}_{\text{EUBO}}$ are maximized. Using $\mathcal{L}_{\text{ELBO}}$ enables effectively increasing the smaller likelihood, while $\tilde{\mathcal{L}}_{\text{EUBO}}$ focuses on the larger one, leading to a less efficient optimization.

\section{\rebuttal{Additional Discussions and Future Directions on the RL Algorithm}}
\label{app:sec:rl}
\paragraph{\rebuttal{RL Stabilization.}}
\rebuttal{In \name{}, we employ a simple policy gradient objective. However, as proposed in PPO~\citep{ppo} and GRPO~\citep{shao2024deepseekmath}, various components including the importance sampling, clipping, and KL regularization are applied to resolve issues like off-policyness and stabilize the training. In principle, we should be able to adopt these techniques in \name{}. For example, the same heuristic trick as in \done or UniGRPO can be applied, using a one-step estimation or the ELBO as a proxy for log-likelihood in the importance sampling ratio and KL divergence terms. However, it remains unclear what is a principally correct and elegant way to incorporate these techniques to dLLMs, especially with the intractable log-likelihood challenge.}

\rebuttal{Empirically, we found our current SPG algorithms already performant. Therefore, we chose to keep the current SPG in its simplest form. We leave correctly adapting RL stabilization to dLLMs as an interesting future direction, such as how to use KL and trust-region optimization correctly for dLLMs.}

\rebuttal{Additionally, calculating group-relative advantages can introduce biases. These biases can be addressed by employing the leave-one-out empirical expectation of the reward, known as RLOO~\citep{ahmadian2024back,rloo}. Notably, RLOO remains unbiased even in off-policy settings~\citep{flet2024contrastive}, eliminating the need for importance sampling. Further reducing biases in the policy update objective for dLLMs using RLOO or other methods is an interesting direction for future work.}

\paragraph{\rebuttal{The Reward Shifting Alternate.}}
\rebuttal{We note that in episodic RL with guaranteed termination and end-point reward only, or when KL is applied (the maximal entropy RL framework), optimal policy is invariant to reward shifting. This suggests an alternative approach to the log-likelihood estimation problem in dLLMs: simply shift reward values by a constant so they are always positive, and then apply the ELBO across all traces. However, this method breaks down when partial rewards are present (such as with long CoT traces), or in infinite horizon settings (e.g., in online learning), where reward shifting is no longer applicable. More importantly, even if reward shifting were possible, it could alter the learning dynamics. Prior work in RL for language models shows that RL can reduce forgetting in pretrained models and enhance exploration compared to SFT-only approaches (while SFT could be viewed as assigning 0 reward to negative traces and 1 to positive ones). In fact, decreasing the log-likelihood of negative traces increases entropy, thereby improving exploration and reducing overfitting. Such benefit is missing when we conduct the reward shifting. Therefore, we believe our \name{} method remains advantageous.}

\rebuttal{Empirically, we experiment with a reward-shifted objective by normalizing each trace’s advantage using the minimum value within the batch, applying both REINFORCE and GRPO and estimating the log-likelihood with the ELBO. The results are shown in \Cref{tab:rwd_shifting}. Under both settings, the reward shifting method underperforms \name{} and most other baselines.}

\begin{table}[!h]
\centering
\caption{\rebuttal{Performance of the reward shifting method on MATH500 and Countdown. \name{} outperforms the reward shifting with both REINFORCE and GRPO objectives.}}
\label{tab:rwd_shifting}
\adjustbox{width=0.75\textwidth,center}{
\begin{tabular}{lcccc cccc}
\toprule
& \multicolumn{4}{c}{\textbf{MATH500 (0-shot)}} & \multicolumn{4}{c}{\textbf{Countdown (0-shot)}}  \\\cmidrule(lr){2-5} \cmidrule(lr){6-9} 
\textbf{Model / Seq Len} & \textbf{128} & \textbf{256} & \textbf{512} & \textit{Avg.} & \textbf{128} & \textbf{256} & \textbf{512} & \textit{Avg.}\\
\midrule
REINFORCE + Reward Shift & 28.0 & 34.4 & 39.0 & 33.8 & 13.3 & 25.0 & 37.9 & 25.4\\
GRPO + Reward Shift & 24.4 & 33.4 & 36.2 & 31.3 & 23.1 & 23.8 & 48.1 & 31.6 \\
\midrule
\name{} w/ EUBO & 33.2 & 37.6 & 39.4 & 36.7 & 68.4 & 71.5 & 68.0 & 69.3 \\
\name{} w/ Mixture & 33.4 & 40.0 & 41.8 & 38.4 & 68.8 & 70.7 & 70.3 & 69.9\\
\bottomrule
\end{tabular}
}
\end{table}

\section{Additional Experimental Details}
\label{app:exp}
\subsection{Datasets and Reward Functions}
\label{app:exp.dataset}
We follow the setting in \done~\citep{d1} and \wdone~\citep{tang2025wd1}, using the same reward functions and train-test splitting, except for Sudoku. The rewards are designed to encourage both correctness and proper formatting, with varying levels of granularity tailored for each task. For completeness, we provide details as follows.

\paragraph{GSM8K.} We utilize the train split of the GSM8K dataset\footnote{\url{https://huggingface.co/datasets/openai/gsm8k}} for RL training, and evaluate model performance on the test split. We follow the Unsloth reward setup\footnote{\url{https://unsloth.ai/blog/r1-reasoning}}, utilizing five equally-weighted additive components: 
\begin{itemize}
    \item XML Structure Reward: +0.125 per correct formatting tag; small penalties for extra contents after the closing tag.
    \item Soft Format Reward: +0.5 for outputs matching the pattern:\\
    \texttt{<reasoning>...</reasoning><answer>...</answer>}
    \item Strict Format Reward: +0.5 for exact formatting with correct line breaks.
    \item Integer Answer Reward: +0.5 if the answer is a valid integer.
    \item Correctness Reward: +2.0 of the answer matches the ground truth.
\end{itemize}

\paragraph{MATH500.} We utilize the train split of the MATH dataset\footnote{\url{https://huggingface.co/datasets/ankner/math-500}} for RL training, and evaluate model performance on the test split. We use a format reward and a correctness reward:
\begin{itemize}
    \item Format Reward: We award 1.00 if <answer></answer> tags are present with \textbackslash boxed inside them; 0.75 if answer tags are present without \textbackslash boxed; 0.50 if answer tags are not present but \textbackslash boxed is present; 0.25 if neither the answer tags nor \textbackslash boxed is present.
    \item Correctness Reward: We award 2.00 if the answer in \texttt{\textbackslash boxed\{\}} matches the ground truth.
\end{itemize}

\paragraph{Countdown.} We utilize the train split of the Countdown dataset\footnote{\url{https://huggingface.co/datasets/Jiayi-Pan/Countdown-Tasks-3to4}} for RL training, restricting to instances that use only three numbers. We evaluate on the same set of 256 synthetically generated countdown questions with 3 numbers as in \done~\citep{d1}. The reward covers three cases: +1.0 if the expression reaches the target using the exact numbers; +0.1 if the numbers are correct but does not reach the target; +0.0 otherwise.

\paragraph{Sudoku.} We experiment on the 4$\times$4 Sudoku dataset\footnote{\url{https://github.com/Black-Phoenix/4x4-Sudoku-Dataset}} generated by \cite{sudoku}. The original training split contains 1M unique Sudoku puzzles covering all 288 4$\times$4 Soduku solutions. To avoid train-test leakage and potential cheating by memorizing all the solutions, we randomly select 200 solutions and include all puzzles corresponding to these solutions into the new training set, resulting in 694,006 training puzzles. We then randomly select 2 or 3 puzzles corresponding to the left 88 solutions to construct the test set, which has 256 Soduku puzzles in total. 

We observe that the zero-shot setting is too difficult for the base LLaDA-8B-Instruct model, which has test accuracy below 7\% with a generation length of 256 and struggles to correctly interpret the questions, leading to very few meaningful RL rollouts. Therefore, we instead use 3-shot for all the Sudoku experiments. We ensure that the solutions presented in the 3-shot samples do not appear in test set solutions, and the puzzles do not appear in both train and test set. The detailed few-shot samples are provided in \Cref{app:exp.fewshot}.

\paragraph{\rebuttal{Coding.}}
\rebuttal{We fine-tune the base model on the KodCode-Light-RL-10k\footnote{\url{https://huggingface.co/datasets/KodCode/KodCode-Light-RL-10K}} dataset. We use a structure reward and a correctness reward, together with a safe code guarantee:}
\begin{itemize}
    \item \rebuttal{XML Structure Reward: the same function for GSM8K is used, with an additional 0.5 points if the program is within answer tags. If the code block is not wrapped in} \verb|```python```|\rebuttal{, 0 points are given.}
    \item \rebuttal{Correctness Reward: we use unit tests to verify the correctness of the code, and use the fraction of unit tests passed as the reward.}
    \item \rebuttal{Safe Code: we assign 0 reward if any blocked modules are used, including \texttt{os}, \texttt{sys}, \texttt{shutil}, \texttt{subprocess}, \texttt{socket}, \texttt{psutil}, \texttt{ctypes}, \texttt{pathlib}, \texttt{builtins}, and} \verb|__import__|.
\end{itemize}

\subsection{Hyperparameter Settings and Implementation Details}
\label{app:exp.hyper}
We follow \done~\citep{d1} for most hyperparameter settings. We employ Low-Rank Adaptation (LoRA) with a rank of $r=128$ and scaling factor $\alpha=64$. The training was conducted on 8 NVIDIA A100-80G or NVIDIA H100-80G GPU, with the following hyperparameters: batch size of 6 per GPU, and gradient accumulation steps of 2. We set the number of inner gradient update $\mu$ as 4 for all models. We use the AdamW optimizer~\citep{adamw}, with $\beta_1=0.9$, $\beta_2=0.99$, weight decay of 0.1, learning rate of $3 \times 10^{-6}$, and gradient clipping at 0.2. We utilize Flash Attention 2~\citep{dao2023flashattention}and 4-bit quantization. 

For RL rollout, we use sequence length of 256 tokens, and 128 diffusion steps. We employ confidence-based semi-autoregressive generation with block size 32, and set the temperature as 0.9 (except for Sudoku where temperature is set as 0.3 following \done). We set number of completions per prompt $g$ as 6, and number of Monte Carlo estimation samples $m$ as 2 due to computational constraint. Since the rollout stage dominates the training time, the average time per gradient update step for \name\ is similar to that of the other baselines. 

We train 6000 steps (i.e., number of gradient updates) for GSM8K, Countdown, and coding, 4000 steps for MATH500, and 2500 steps for Sudoku. For all RL models, we run evaluation every 100 steps with generation sequence length 128, 256, 512, and report the result of the checkpoint with the highest average accuracy over the three generation lengths, except for the ablations on inference strategies in \Cref{tab:ablation_inference}, where we only evaluate on generation length 256. 

\subsection{Few-Shot Samples for Sudoku}
\label{app:exp.fewshot}
\textbf{Sample 1:}
\begin{myquote}
    Question:\\Solve the following Sudoku puzzle: 3014002020004130\\Answer:\\ <reasoning>\\Interpret puzzle as 4 rows of 4:\\R1: 3 0 1 4\\R2: 0 0 2 0\\R3: 2 0 0 0\\R4: 4 1 3 0\\ \\Fill easy singles:\\R1 missing 2 → R1C2=2.\\R4 missing 2 → R4C4=2.\\Box D (R3-4,C3-4) then needs {1,4}; column4 can only accept 1 → R3C4=1, R3C3=4.\\R3 now missing 3 → R3C2=3.\\Column1 missing 1 → R2C1=1.\\Column2 missing 4 → R2C2=4.\\Last cell R2C4=3.\\ \\Final grid:\\R1: 3 2 1 4\\R2: 1 4 2 3\\R3: 2 3 4 1\\R4: 4 1 3 2\\</reasoning>\\<answer>\\3214142323414132\\</answer>
\end{myquote}

\textbf{Sample 2:}
\begin{myquote}
    Question:\\Solve the following Sudoku puzzle: 0000100420013142\\Answer:\\<reasoning>\\Interpret puzzle as 4 rows of 4:\\R1: 0 0 0 0\\R2: 1 0 0 4\\R3: 2 0 0 1\\R4: 3 1 4 2\\ \\Fill easy singles:\\Col1 missing 4 → R1C1=4.\\Col4 missing 3 → R1C4=3.\\Box A (R1-2,C1-2) missing {2,3} and R1 now needs {1,2} → R1C2=2, R2C2=3.\\R1C3=1.\\R2 now missing 2 → R2C3=2.\\Col2 missing 4 → R3C2=4, then R3C3=3.\\ \\Final grid:\\R1: 4 2 1 3\\R2: 1 3 2 4\\R3: 2 4 3 1\\R4: 3 1 4 2\\</reasoning>\\<answer>\\4213132424313142\\</answer>
\end{myquote}

\textbf{Sample 3:}
\begin{myquote}
    Question:\\Solve the following Sudoku puzzle: 2001403002001420\\Answer:\\<reasoning>\\Interpret puzzle as 4 rows of 4:\\R1: 2 0 0 1\\R2: 4 0 3 0\\R3: 0 2 0 0\\R4: 1 4 2 0\\ \\Fill easy singles:\\R1 missing {3,4}; Col2 can't be 1 so R1C2=3 → R1C3=4.\\R4 missing 3 → R4C4=3.\\Col4 missing {2,4}; R2 must take 2 → R2C4=2 → R2C2=1.\\Col1 missing 3 → R3C1=3.\\Col3 missing 1 → R3C3=1 → R3C4=4.\\ \\Final grid:\\R1: 2 3 4 1\\R2: 4 1 3 2\\R3: 3 2 1 4\\R4: 1 4 2 3\\</reasoning>\\<answer>\\2341413232141423\\</answer>
\end{myquote}

\section{Additional Results}
\subsection{Additional Evaluations to the Main Results}
\label{app:results.eval}

\begin{table}[h]
\centering
\caption{Complete model performance on four reasoning benchmarks compared with baselines. We provide both the reported and the reproduced results for \done\ and \wdone. The best results are bolded and the second best are underlined. \name \ consistently outperforms all other models.}
\label{tab:main_results_complete}
\adjustbox{width=\textwidth,center}{
\begin{tabular}{lccc ccc ccc ccc}
\toprule
& \multicolumn{3}{c}{\textbf{GSM8K (0-shot)}} & \multicolumn{3}{c}{\textbf{MATH500 (0-shot)}} & \multicolumn{3}{c}{\textbf{Countdown (0-shot)}} & \multicolumn{3}{c}{\textbf{Sudoku (3-shot)}}\\
\cmidrule(lr){2-4} \cmidrule(lr){5-7} \cmidrule(lr){8-10} \cmidrule(lr){11-13}
\textbf{Model / Seq Len} & \textbf{128} & \textbf{256} & \textbf{512} & \textbf{128} & \textbf{256} & \textbf{512} & \textbf{128} & \textbf{256} & \textbf{512} & \textbf{128} & \textbf{256} & \textbf{512} \\
\midrule
LLaDA-8B-Instruct & 69.5 & 77.2 & 79.8 & 28.2 & 32.4 & 34.6 & 18.8 & 16.8 & 16.8 & 5.7 & 27.7 & 26.2 \\
LLaDA-1.5 & 70.4 & 80.5 & 81.9 & 26.8 & 32.2 & 35.8 & 21.9 & 21.1 & 21.5 & 7.4 & 26.9 & 29.0 \\
\done \ \textcolor{gray}{(reported)} & 72.6 & 79.8 & 81.9 & 33.2 & 37.2 & 39.2 & 33.2 & 31.3 & 37.1 & - & - & - \\
\done \ \textcolor{gray}{(reproduced)} & 72.2 & 80.6 & 81.3 & 31.4 & 36.0 & 39.4 & 30.9 & 30.9 & 34.4& 7.2 & 32.5 & 29.3 \\
d1-LLaDA \ \textcolor{gray}{(reported)} & 73.2 & 81.1 & 82.1 & \textbf{33.8} & \underline{38.6} & \underline{40.2} & 34.8 & 32.0 & 42.2 & - & - & - \\
\wdone\ \textcolor{gray}{(reported)} & - & 80.8 & 82.3 & - & 34.4 & 39.0 & - & 51.2 & 46.1 & - & - & - \\
\wdone  \ \textcolor{gray}{(reproduced)}& 74.6 & 81.5 & 83.0 & 31.0 & 37.4 & 39.0 & 48.8 & 52.3 & 50.8 & 33.1 & 32.1 & 22.5 \\
UniGRPO& 74.9 & 82.5 & 82.7 & 32.4 & 37.4 & 39.4 & 44.5 & 43.0 & 57.0 & 59.0 & 67.0 & 62.9 \\
\midrule

\name\ w/ EUBO (ours) & \underline{77.1} & \underline{83.8} & \underline{83.9} & 33.2 & 37.6 & 39.4 & \underline{68.4} & \textbf{71.5} & \underline{68.0} & \underline{81.2} & \underline{87.1} & \underline{89.9} \\
\name \ w/ mixture (ours) & \textbf{78.5} & \textbf{86.1} & \textbf{84.5} & \underline{33.4} & \textbf{40.0} & \textbf{41.8} & \textbf{68.8} & \underline{70.7} & \textbf{70.3} & \textbf{82.9} & \textbf{94.0}& \textbf{93.1} \\
\bottomrule
\end{tabular}
}
\end{table}

\paragraph{Complete evaluation results.} We provide the complete evaluation results, along with those reported in \done~\citep{d1} and \wdone~\citep{tang2025wd1}, in \Cref{tab:main_results_complete}. Our reproduced numbers closely match the reported results. d1-LLaDA~\citep{d1} denotes the model that conducts first SFT and then RL (using \done). All other models are trained solely with RL. 
In \done\ and d1-LLaDA, the best result for each generation length is reported separately, whereas we select a single checkpoint with the highest average accuracy across all three generation lengths, leading to slightly worse results than the reported numbers. The reported results in \wdone\ are based on evaluations on fewer checkpoints, so they are generally a bit lower than our reproduced values.

\begin{figure}[ht]
    \centering
    \includegraphics[width=\linewidth]{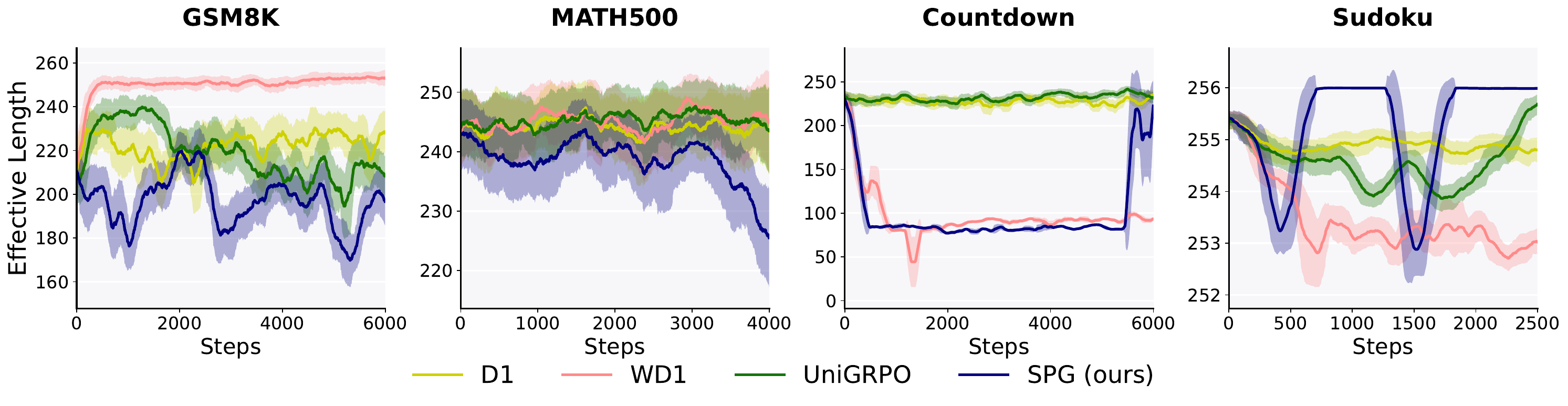}
    \vspace{-10pt}
    \caption{Dynamics of the effective generation length of \name \ during RL training, compared with \done, \wdone, and UniGRPO. \name \ leads to concise solutions with better token efficiency. We report mean and standard deviation over a rolling window of 50 steps.}
    \label{fig:completion_length}
\end{figure}

\paragraph{Dynamics of Completion Length.} We provide the dynamics of the effective sequence length of \name \ during RL training in \Cref{fig:completion_length}. We also report the effective length of the best checkpoint in \Cref{tab:length}. \name \ leads to effective usage of the total given length and good adaptation to task difficulties.

\begin{table}[!h]
\centering
\caption{Effective sequence length of each model at the best checkpoint corresponding to \Cref{tab:main_results} on four reasoning benchmarks.}
\label{tab:length}
\adjustbox{width=0.85\textwidth,center}{
\begin{tabular}{lccc ccc ccc ccc}
\toprule
& \multicolumn{3}{c}{\textbf{GSM8K (0-shot)}} & \multicolumn{3}{c}{\textbf{MATH500 (0-shot)}} & \multicolumn{3}{c}{\textbf{Countdown (0-shot)}} & \multicolumn{3}{c}{\textbf{Sudoku (3-shot)}} \\\cmidrule(lr){2-4} \cmidrule(lr){5-7} \cmidrule(lr){8-10} \cmidrule(lr){11-13}
\textbf{Model / Seq Len} & \textbf{128} & \textbf{256} & \textbf{512} & \textbf{128} & \textbf{256} & \textbf{512} & \textbf{128} & \textbf{256} & \textbf{512} & \textbf{128} & \textbf{256} & \textbf{512} \\
\midrule
LLaDA-8B-Instruct & 114 & 212 & 257 & 123 & 235 & 402 & 111 & 213 & 407 & 111 & 232 & 448 \\
LLaDA-1.5 & 115 & 214 & 265 & 123 & 237 & 407 & 114 & 215 & 411 & 112 & 232 & 419 \\
\done & 115 & 209 & 261 & 123 & 234 & 399 & 107 & 211 & 397 & 111 & 231 & 449 \\
\wdone & 115 & 225 & 312 & 123 & 231 & 378 & 83 & 84 & 90 & 105 & 227 & 473 \\
UniGRPO & 114 & 211 & 257 & 123 & 235 & 400 & 100 & 207 & 374 & 113 & 230 & 472 \\
\midrule
\name\ w/ EUBO & 110 & 196 & 227 & 120 & 228 & 382 & 68 & 70 & 78 & 89 & 137 & 249 \\
\name \ w/ mixture & 108 & 176 & 195 & 121 & 229 & 384 & 75 & 78 & 79 & 115 & 239 & 491 \\
\bottomrule
\end{tabular}
}
\end{table}

\subsection{Additional Ablation Results}
\label{app:results.ablation}
In this section, we provide the complete results for each generation length and task in supplement to \Cref{subsec:ablation}. We also include additional ablation studies on the looser upper bound, different log-likelihood estimation methods for positive advantage traces, and Pass@K performance.
\paragraph{Ablations on Algorithm Components.}  We provide the complete results for ablations on log-likelihood estimation methods in \Cref{tab:ablation_component_complete} and for ablations on masking strategies in \Cref{tab:ablation_mask_complete}.

\begin{table}[!h]
\centering
\caption{Ablations on log-likelihood estimation methods for negative advantage traces. The best results are bolded and the second best are underlined. \name \ w/ Mixture consistently outperforms other likelihood estimation methods.}
\label{tab:ablation_component_complete}
\adjustbox{width=\textwidth,center}{
\begin{tabular}{lcccc cccc cccc cccc}
\toprule
& \multicolumn{4}{c}{\textbf{GSM8K (0-shot)}} & \multicolumn{4}{c}{\textbf{MATH500 (0-shot)}} & \multicolumn{4}{c}{\textbf{Countdown (0-shot)}} & \multicolumn{4}{c}{\textbf{Sudoku (3-shot)}} \\\cmidrule(lr){2-5} \cmidrule(lr){6-9} \cmidrule(lr){10-13} \cmidrule(lr){14-17}
\textbf{Model} & \textbf{128} & \textbf{256} & \textbf{512} & \textit{Avg.} & \textbf{128} & \textbf{256} & \textbf{512} & \textit{Avg.} & \textbf{128} & \textbf{256} & \textbf{512} & \textit{Avg.} & \textbf{128} & \textbf{256} & \textbf{512} & \textit{Avg.} \\
\midrule
\name \ wo/ neg  & 72.0 & 79.0 & 81.3 & 77.4 & 28.2 & 32.2 & 37.8 & 32.7 & 43.8 & 48.1 & 44.5 & 45.5 & 55.0 & 82.9 & 68.4 & 68.8\\
\name \ w/ ELBO & 75.6 & 82.8 & 84.4 & 80.9 & 35.8 & 37.6 & 38.8 & \underline{37.4} & 66.8 & 66.0 & 68.4 & 67.1 & 73.8 & 89.4 & 84.1 & 82.4 \\
\name \ w/ EUBO & 77.1 & 83.8 & 83.9 & \underline{81.6} & 33.2 & 37.6 & 39.4 & 36.7 & 68.4 & 71.5 & 68.0 & \underline{69.3} & 81.2 & 87.1 & 89.9 & \underline{86.1} \\
\name \ w/ Mixture & 78.5 & 86.1 & 84.5 & \textbf{83.0} & 33.4 & 40.0 & 41.8 & \textbf{38.4} & 68.8 & 70.7 & 70.3 & \textbf{69.9} & 82.9 & 94.0 & 93.1 & \textbf{90.0} \\
\bottomrule
\end{tabular}
}
\end{table}

\begin{table}[h]
\centering
\caption{Ablations on the masking strategies in Monte Carlo estimation. Our block-wise masking strategy leads to consistent improvement to random masking on both benchmarks.}
\label{tab:ablation_mask_complete}
\adjustbox{width=0.75\textwidth,center}{
\begin{tabular}{llcccc cccc}
\toprule
&& \multicolumn{4}{c}{\textbf{MATH500 (0-shot)}} & \multicolumn{4}{c}{\textbf{Countdown (0-shot)}}  \\\cmidrule(lr){3-6} \cmidrule(lr){7-10} 
\textbf{Model} & \textbf{Masking}  & \textbf{128} & \textbf{256} & \textbf{512} & \textit{Avg.} & \textbf{128} & \textbf{256} & \textbf{512} & \textit{Avg.} \\
\midrule
\name \ w/ EUBO & random & 33.4 & 35.4 & 41.4 & \textbf{36.7} & 42.6 & 41.0 & 52.7 & 45.4 \\
& block-wise & 33.2 & 37.6 & 39.4 & \textbf{36.7} & 68.4 & 71.5 & 68.0 & \textbf{69.3} \\\midrule
\name \ w/ Mixture & random & 33.8 & 38.2 & 38.8 & 36.9 & 52.3 & 64.5 & 71.5 & 62.8 \\
& block-wise & 33.4 & 40.0 & 41.8 & \textbf{38.4} & 68.8 & 70.7 & 70.3 & \textbf{69.9} \\

\bottomrule
\end{tabular}
}
\end{table}

\paragraph{Ablations on Key Hyperparameters $\beta$ and $\omega$.} We provide the complete results for ablations on $\beta$ in \Cref{tab:ablation_beta_complete} and for ablations on $\omega$ in \Cref{tab:ablation_omega_complete}.

\begin{table}[!h]
\centering
\caption{Ablations on the value of $\beta$ in the upper bound.}
\label{tab:ablation_beta_complete}
\adjustbox{width=\textwidth,center}{
\begin{tabular}{llcccc cccc cccc cccc}
\toprule
& & \multicolumn{4}{c}{\textbf{GSM8K (0-shot)}} & \multicolumn{4}{c}{\textbf{MATH500 (0-shot)}} & \multicolumn{4}{c}{\textbf{Countdown (0-shot)}} & \multicolumn{4}{c}{\textbf{Sudoku (3-shot)}} \\\cmidrule(lr){3-6} \cmidrule(lr){7-10} \cmidrule(lr){11-14} \cmidrule(lr){15-18}
\textbf{Model}& $\beta$ & \textbf{128} & \textbf{256} & \textbf{512} & \textit{Avg.} & \textbf{128} & \textbf{256} & \textbf{512} & \textit{Avg.} & \textbf{128} & \textbf{256} & \textbf{512} & \textit{Avg.} & \textbf{128} & \textbf{256} & \textbf{512} & \textit{Avg.}\\
\midrule
& 0.50 & 77.7 & 83.2 & 84.5 & 81.8 & 32.8 & 36.4 & 41.2 & 36.8 & 71.1 & 68.8 & 74.6 & 71.5 & 64.7 & 53.4 & 57.4 & 58.5 \\
& 0.75 & 77.2 & 83.9 & 84.5 & 81.9 & 31.0 & 36.6 & 40.0 & 35.9 & 70.7 & 70.7 & 70.7 & 70.7 & 63.4 & 65.7 & 45.4 & 58.2 \\\cmidrule(lr){2-18}
\name\ w/ EUBO&1.00 & 76.5 & 83.9 & 83.6 & 81.3 & 31.0 & 37.4 & 38.8 & 35.7 & 66.0 & 66.8 & 66.4 & 66.4 & 81.2 & 87.1 & 89.9 & 86.1 \\
&1.50 & 77.1 & 83.8 & 83.9 & 81.6 & 33.2 & 37.6 & 39.4 & 36.7 & 69.5 & 64.5 & 66.4 & 66.8 & 32.7 & 40.5 & 39.9 & 37.7 \\
&2.00 & 76.5 & 83.9 & 83.2 & 81.2 & 32.4 & 36.8 & 38.2 & 35.8 & 68.4 & 71.5 & 68.0 & 69.3 & 28.1 & 31.9 & 28.0 & 29.3 \\\midrule
&1.00 & 78.8 & 85.6 & 84.9 & 83.1 & 34.0 & 40.2 & 39.2 & 37.8 & 69.9 & 69.5 & 70.3 & 69.9 & 82.9 & 94.0 & 93.1 & 90.0 \\
\name\ w/ Mixture&1.50 & 78.5 & 86.1 & 84.5 & 83.1 & 33.4 & 40.0 & 41.8 & 38.4 & 68.8 & 70.7 & 70.3 & 69.9 & 83.2 & 86.0 & 84.6 & 84.6 \\
&2.00 & 78.8 & 85.7 & 84.7 & 83.1 & 32.4 & 38.8 & 39.8 & 37.0 & 70.3 & 69.1 & 69.5 & 69.6 & 44.3 & 60.5 & 60.7 & 55.2\\
\bottomrule
\end{tabular}
}
\end{table}

\begin{table}[!h]
\centering
\caption{Ablations on the mixture coefficient $\omega$ on MATH500 and Countdown.}
\label{tab:ablation_omega_complete}
\adjustbox{width=0.65\textwidth,center}{
\begin{tabular}{lcccc cccc}
\toprule
\name \ w/ Mixture& \multicolumn{4}{c}{\textbf{MATH500 (0-shot)}} & \multicolumn{4}{c}{\textbf{Countdown (0-shot)}}  \\\cmidrule(lr){2-5} \cmidrule(lr){6-9} 
$\omega$ & \textbf{128} & \textbf{256} & \textbf{512} & \textit{Avg.} & \textbf{128} & \textbf{256} & \textbf{512} & \textit{Avg.}\\
\midrule
0.00 & 35.8 & 37.6 & 38.8 & 37.4 & 66.8 & 66.0 & 68.4 & 67.1\\
0.25 & 34.6 & 37.6 & 42.2 & 38.1 & 71.5 & 68.0 & 67.2 & 68.9 \\
0.50 & 33.4 & 40.0 & 41.8 & 38.4 & 68.8 & 70.7 & 70.3 & 69.9 \\
0.75 & 34.2 & 38.6 & 41.2 & 38.0 & 69.5 & 69.1 & 74.2 & 70.9 \\
1.00 & 33.2 & 37.6 & 39.4 & 36.7 & 69.5 & 64.5 & 66.4 & 66.8\\
\bottomrule
\end{tabular}
}
\end{table}

\paragraph{Ablations on Inference Strategies.} We provide complete results for ablations on different inference strategies in \Cref{tab:ablation_inference}. Note that the reported numbers of each method for ``Semi-AR, Confidence, Block=32'' is in general slightly higher than the results in \Cref{tab:main_results} under the same inference setting. This is because in \Cref{tab:ablation_inference}, we select the checkpoint with the highest accuracy specifically for generation length 256 to maintain consistency with other inference settings, while in \Cref{tab:main_results}, we choose the checkpoint with the highest average accuracy across generation lengths 128, 256, and 512.

\paragraph{Ablations on the Looser Upper Bound.} As mentioned in \Cref{subsec:method.eubo} and \Cref{app:theory}, a looser but unbiased bound can be derived using inequalities like $\log(x) \leq x-1$, i.e., $\tilde{\mathcal{L}}_{\text{Loose}}$ (\Cref{eq:loose}). However, as shown in \Cref{tab:ablation_loose}, this looser bound performs worse empirically than the tighter upper bound $\tilde{\mathcal{L}}_{\text{EUBO}}$  we used, possibly due to a larger discrepancy from the true log-likelihood.

\begin{table}[!h]
\centering
\caption{Ablations on the looser upper bound. The loose bound performs worse than the tighter upper bound we used, indicating inferior performance due to a larger discrepancy from the true log-likelihood.}
\label{tab:ablation_loose}
\adjustbox{width=0.75\textwidth,center}{
\begin{tabular}{llcccc cccc}
\toprule
\name \ w/ EUBO && \multicolumn{4}{c}{\textbf{MATH500 (0-shot)}} & \multicolumn{4}{c}{\textbf{Countdown (0-shot)}}  \\\cmidrule(lr){3-6} \cmidrule(lr){7-10} 
$\beta$ & \textbf{Upper Bound} & \textbf{128} & \textbf{256} & \textbf{512} & \textit{Avg.} & \textbf{128} & \textbf{256} & \textbf{512} & \textit{Avg.} \\
\midrule
1.0 & $\tilde{\mathcal{L}}_{\text{Loose}}$  & 29.4& 35.4 & 39.4 & 34.7 & 43.8 & 65.2 & 64.8 & 57.9 \\
& $\tilde{\mathcal{L}}_{\text{EUBO}}$  & 31.0 & 37.4 & 38.8 & \textbf{35.7} & 66.0 & 66.8 & 66.4 & \textbf{66.4} \\\midrule
1.5 & $\tilde{\mathcal{L}}_{\text{Loose}}$  & 29.8 & 31.8 & 38.8 & 33.5 & 46.9 & 54.7 & 57.0 & 52.9 \\
& $\tilde{\mathcal{L}}_{\text{EUBO}}$  & 33.2 & 37.6 & 39.4 & \textbf{36.7} & 69.5 & 64.5 & 66.4 & \textbf{66.8} \\

\bottomrule
\end{tabular}
}
\end{table}

\paragraph{Ablations on Log-Likelihood Estimations for Positive Advantage Traces.} Instead of always using $\mathcal{L}_{\text{ELBO}}$ for positive advantage traces, we experiment on MATH500 and Countdown benchmarks using both $\tilde{\mathcal{L}}_{\text{EUBO}}$ and $\tilde{\mathcal{L}}_{\text{Mix}}$ for positive advantage traces.
Correspondingly, we use $\omega=0.5$ and the best performed $\beta$ as previously discussed for negative advantage traces.
For the positive advantage traces, we always use the tightest $\beta=1.0$ for both $\tilde{\mathcal{L}}_{\text{EUBO}}$ and $\tilde{\mathcal{L}}_{\text{Mix}}$. The results are shown in \Cref{tab:ablation_pos}, indicating that using the upper bound for likelihood estimation of positive advantage traces performs worse than using $\mathcal{L}_{\text{ELBO}}$. This aligns well with our theoretical insights that the lower bound is a better objective for log-likelihood maximization.
\begin{table}[t]
\centering
\caption{Ablations on log-likelihood estimation for positive advantage traces. Using the upper bound for log-likelihood estimation of positive advantage traces perform worse than using the lower bound.}
\label{tab:ablation_pos}
\adjustbox{width=0.8\textwidth,center}{
\begin{tabular}{llcccc cccc}
\toprule
&Positive traces  & \multicolumn{4}{c}{\textbf{MATH500 (0-shot)}} & \multicolumn{4}{c}{\textbf{Countdown (0-shot)}}  \\\cmidrule(lr){3-6} \cmidrule(lr){7-10} 
\textbf{Model} & likelihood estimation & \textbf{128} & \textbf{256} & \textbf{512} & \textit{Avg.} & \textbf{128} & \textbf{256} & \textbf{512} & \textit{Avg.} \\
\midrule
\name \ w/ EUBO& $\tilde{\mathcal{L}}_{\text{EUBO}}$ ($\beta=1.0$) & 34.4 & 36.2 & 39.2 & 36.6 & 48.1 & 46.7 & 50.8 & 48.5 \\
 & $\mathcal{L}_{\text{ELBO}}$& 33.2 & 37.6 & 39.4 & \textbf{36.7} & 68.4 & 71.5 & 68.0 & \textbf{69.3}\\
\midrule
\name \ w/ Mixture& $\tilde{\mathcal{L}}_{\text{Mix}}$ ($\beta=1.0,\omega=0.5$) & 35.4 & 38.4 & 39.0 & 37.6 & 69.1 & 68.4 & 70.3 & 69.3 \\
 & $\mathcal{L}_{\text{ELBO}}$ &33.4 & 40.0 & 41.8 & \textbf{38.4} & 68.8 & 70.7 & 70.3 & \textbf{69.9}\\

\bottomrule
\end{tabular}
}
\end{table}

\paragraph{Ablations on Pass@K Performance.} In all previous experiments, we apply greedy sampling by setting temperature as 0.0 following \done\ and LLaDA. However, beyond accuracy, it is essential for models to generate a diverse set of outputs that can cover the correct solution and allow for explorations. In this section, we investigate the models’ ability to generate diverse outputs using a higher temperature, and evaluate their Pass@K performance on MATH500 and Countdown, as shown in \Cref{tab:ablation_passatk}. Specifically, we set temperature to 0.9 and generation length to 256, conduct evaluations every 100 steps, and report results from the checkpoint with the highest accuracy. For comparison, we also include results from greedy sampling, denoted as Pass@1{\scriptsize{Greedy}}. As expected, increasing the temperature leads to a decrease in Pass@1 performance across all models, aligning with observations from previous work. For K>1, the Pass@K scores improve for all models as K increases from 1 to 4. \name \ achieves the best performance across all settings, with \name \ w/ Mixture reaching 55.6\% Pass@4 accuracy on MATH500 and 76.6\% on Countdown, demonstrating the ability of \name \ to generate diverse outputs that can recover the correct solution.

\begin{table}[!h]
\centering
\caption{Pass@K performance of each model on MATH500 and Countdown. We set temperature as 0.9 and report results of the best checkpoint of each case at a generation length of 256. For comparison, we also include the greedy sampling performance, i.e., Pass@1{\scriptsize{Greedy}}. The best results are bolded and the second best are underlined.}
\label{tab:ablation_passatk}
\adjustbox{width=\textwidth,center}{
\begin{tabular}{lccccc ccccc}
\toprule
& \multicolumn{5}{c}{\textbf{MATH500 (0-shot)}} & \multicolumn{5}{c}{\textbf{Countdown (0-shot)}} \\\cmidrule(lr){2-6} \cmidrule(lr){7-11}
\textbf{Model} & \textbf{Pass@1}\scriptsize{\textbf{Greedy}} & \textbf{Pass@1} & \textbf{Pass@2} & \textbf{Pass@3} & \textbf{Pass@4} & \textbf{Pass@1}\scriptsize{\textbf{Greedy}} & \textbf{Pass@1} & \textbf{Pass@2} & \textbf{Pass@3} & \textbf{Pass@4}\\
\midrule
LLaDA-8B-Instruct & 32.4 & 31.5 & 40.9 & 45.7 & 48.8 & 16.8 & 15.8 & 28.1 & 37.7 & 45.3 \\
LLaDA-1.5 & 32.2 & 32.6 & 42.2 & 47.4 & 50.4 & 21.1 & 18.2 & 32.1 & 42.5 & 50.0 \\
\done & 37.8 & 34.3 & 43.1 & 48.0 & 52.0 & 32.4 & 24.5 & 40.4 & 51.4 & 60.6\\
\wdone &\underline{38.6} & \underline{36.0} & \underline{44.9} & \underline{49.9} & 53.6 & 54.7 & 44.3 & 60.6 & 68.0 & 73.1 \\
UniGRPO & 38.4 & 34.7 & 43.9 & 49.5 & 53.2 & 44.9 & 36.8 & 55.2 & 65.0 & 72.3\\
\midrule
\name\ w/ EUBO & 38.0 & 34.4 & 44.3 & \underline{49.9} & \underline{54.0} & \textbf{71.5} & \textbf{68.2} & \underline{71.9} & \underline{73.9} & \textbf{76.6} \\
\name \ w/ mixture & \textbf{40.0} & \textbf{36.5} & \textbf{46.0} & \textbf{51.2} & \textbf{55.6} & \underline{71.1} & \underline{67.5} & \textbf{72.5} & \textbf{75.1} & \textbf{76.6} \\
\bottomrule
\end{tabular}
}
\end{table}

\begin{table}[h]
\centering
\caption{Ablations on the inference strategy. \name \ leads to consistently superior performance to baselines with different inference strategies. The best results are bolded and the second best are underlined for each setting. We report results for generation length 256.}
\label{tab:ablation_inference}
\adjustbox{width=\textwidth,center}{
\begin{tabular}{llc c c c c}
\toprule
\textbf{Inference Strategy} & \textbf{Model} & \multicolumn{1}{c}{\textbf{GSM8K}} & \multicolumn{1}{c}{\textbf{MATH500}} & \multicolumn{1}{c}{\textbf{Countdown}} & \multicolumn{1}{c}{\textbf{Sudoku}} & \textit{Avg.}\\ 
\midrule
& LLaDA-8B-Instruct & 78.7 & 31.4 & 13.7& 26.2 & 37.5\\
& LLaDA-1.5 & 78.8 & 33.4 & 16.0 & 23.0 & 37.8\\
& \done & 79.7 & 37.2 & 27.0 & 31.4 & 43.8 \\
Semi-AR, Confidence, Block=16 & \wdone & 82.3 & \underline{37.4} & 53.9 & 36.8 & 52.6 \\
& UniGRPO & 82.5 & 36.8 & 46.5 & 63.4 & 57.3\\
& \name \ w/ EUBO & \underline{84.7} & \underline{37.4}& \underline{70.3} & \underline{82.2} & \underline{68.7} \\
& \name \ w/ Mixture & \textbf{86.4} & \textbf{40.8} & \textbf{70.7} & \textbf{96.2} & \textbf{73.5} \\

\midrule
& LLaDA-8B-Instruct & 77.2 & 32.4 & 16.8 & 27.7 & 38.5 \\
& LLaDA-1.5 & 80.5 & 32.2 & 21.1 & 26.9 & 40.2\\
& \done & 80.6 & 37.8 & 32.4 & 32.8 & 45.9 \\
Semi-AR, Confidence, Block=32 & \wdone & 81.7 & \underline{38.6} & 54.7 & 35.7 & 58.1\\
& UniGRPO & 82.6 & 38.4 & 44.9 & 67.0 & 58.2\\
& \name \ w/ EUBO & \underline{84.8} & 38.0 & \textbf{71.5} & \underline{88.5} & \underline{70.7} \\
& \name \ w/ Mixture & \textbf{86.2} & \textbf{40.0} & \underline{71.1} & \textbf{95.6} & \textbf{73.2} \\

\midrule
& LLaDA-8B-Instruct & 78.6 & 33.2 & 27.3 & 32.6 & 42.9 \\
& LLaDA-1.5 & 81.0 & 35.4 & 20.3 & 36.4 & 43.3 \\
& \done & 80.9 & \underline{37.6} & 38.3 & 39.8 & 49.2\\
Semi-AR, Confidence, Block=64 & \wdone & 82.5 & 37.4 & 52.3 & 41.8 & 53.5 \\
& UniGRPO & 82.3 & 37.4 & 53.5 & 82.9 & 64.0 \\
& \name \ w/ EUBO & \underline{84.3} & 37.4 & \underline{69.5} & \underline{88.8} & \underline{70.0} \\
& \name \ w/ Mixture & \textbf{85.5} & \textbf{41.4} & \textbf{69.9} & \textbf{93.8} & \textbf{72.7} \\

\midrule
& LLaDA-8B-Instruct & 63.5 & 21.0 & 6.3 & 24.4 & 28.8 \\
& LLaDA-1.5 & 67.1 & 24.8 & 10.9 & 27.5 & 32.6\\
& \done & 69.7 & 27.4 & 18.4 & 29.9 & 36.4\\
Semi-AR, Random, Block=32 & \wdone & \underline{74.1} & 30.8 & 37.5 & 29.9 & 43.1 \\
& UniGRPO & 72.8 & 29.8 & 41.4 & 60.1 & 51.0\\
& \name \ w/ EUBO & \underline{74.1} & \textbf{31.4} & \underline{42.6} & \underline{72.6} & \underline{55.2}\\
& \name \ w/ Mixture & \textbf{78.4} & \underline{31.0} & \textbf{66.0} & \textbf{86.9} & \textbf{65.6}\\

\midrule
& LLaDA-8B-Instruct & 52.7 & 17.8 & 13.3 & 27.1 & 27.7\\
& LLaDA-1.5 & 53.5 & 19.8 & 13.7 & 28.0 & 28.8 \\
& \done & 53.0 & 22.6 & 28.5 & 31.8 & 34.0 \\
Full Sequence, Random & \wdone & 54.6 & 23.4 & 43.4 & 30.0 & 37.9\\
& UniGRPO & 57.6 & 22.6 & 47.7 & 57.3 & 46.3\\
& \name \ w/ EUBO & \underline{58.5} & \textbf{25.2} & \underline{56.6} & \underline{71.3} & \underline{52.9}\\
& \name \ w/ Mixture & \textbf{61.2} & \textbf{25.2} & \textbf{64.1} & \textbf{72.2} & \textbf{55.7} \\

\midrule
& LLaDA-8B-Instruct & 23.9 & 17.8 & 0.0 & 68.3 & 27.5 \\
& LLaDA-1.5 & 41.4 & 20.4 & 0.0 & 67.9 & 32.4 \\
& \done & \underline{57.5} & 22.6 & 0.0 & 72.3 & 38.1\\
Full Sequence, Confidence & \wdone & 56.7 & \textbf{25.0} & 10.2 & 68.9 & 40.2\\
& UniGRPO & 50.0 & \underline{24.2} & 8.2 & \underline{95.6} & 44.5\\
& \name \ w/ EUBO & 54.3 & 23.4 & \underline{63.3} & 88.9& \underline{57.5} \\
& \name \ w/ Mixture & \textbf{58.8} & 24.0 & \textbf{65.6} & \textbf{96.5} & \textbf{61.2} \\
\bottomrule
\end{tabular}
}
\end{table}

\paragraph{\rebuttal{Ablations on Full Fine-Tuning.}}

\rebuttal{In our main experiments, we follow the setting of \done{} and \wdone, using LoRA for \name{} and all the baselines. This is a typical setting for RL-based fine-tuning of language models. As noted by \citet{schulman2025lora}, LoRA performs equivalently to full fine-tuning for reinforcement learning even with small ranks since RL requires very low capacity.}

\rebuttal{To further validate this in our context, we provide additional results for \name{} w/ Mixture using full fine-tuning, alongside those obtained with LoRA, in \Cref{tab:fullft}. Full fine-tuning yields comparable or even better performance relative to LoRA, demonstrating the robustness and consistent improvements of \name. Note that full fine-tuning generally requires a smaller learning rate than LoRA. Accordingly, we use a learning rate of $2\times10^{-7}$ for MATH500 and $3\times10^{-7}$ for Countdown.}

\begin{table}[!h]
\centering
\caption{\rebuttal{Ablations on full fine-tuning on MATH500 and Countdown.}}
\label{tab:fullft}
\adjustbox{width=0.65\textwidth,center}{
\begin{tabular}{lcccc cccc}
\toprule
\name{} w/ Mixture & \multicolumn{4}{c}{\textbf{MATH500 (0-shot)}} & \multicolumn{4}{c}{\textbf{Countdown (0-shot)}}  \\\cmidrule(lr){2-5} \cmidrule(lr){6-9} 
\textbf{FT Setting / Seq Len} & \textbf{128} & \textbf{256} & \textbf{512} & \textit{Avg.} & \textbf{128} & \textbf{256} & \textbf{512} & \textit{Avg.}\\
\midrule
LoRA & 33.4 & 40.0 & 41.8 & \textbf{38.4} & 68.8 & 70.7 & 70.3 & 69.9\\
Full-FT & 34.0 & 39.8 & 40.6 & 38.1 & 78.5 & 76.6 & 76.2 & \textbf{77.1} \\
\bottomrule
\end{tabular}
}
\end{table}

\paragraph{\rebuttal{Ablations on Prompt Masking.}}
\rebuttal{We adopt the same prompt masking technique as \done, randomly masking some prompt tokens in the perturbed sample $\bm{z}_t$ with a small probability during log-likelihood estimation. As discussed in \citet{d1}, this stochastic masking creates perturbed versions of the same (prompt, completion) pairs, which acts as a regularizer for policy optimization. It can also be interpreted as data augmentation, enabling the extraction of more supervision signals from the same dataset.}

\rebuttal{Additionally, we report results for \name{} w/ Mixture without applying the prompt masking trick in \Cref{tab:prompt_masking}. In practice, prompt masking has minimal impact on \name, yielding nearly identical average accuracy on both MATH and Countdown. This highlights the robustness of \name's log-likelihood estimation method, in contrast to the one-step estimation used in \done, which already provides strong supervision signals even without prompt masking.}

\begin{table}[!h]
\centering
\caption{\rebuttal{Ablations on prompt masking on MATH500 and Countdown.}}
\label{tab:prompt_masking}
\adjustbox{width=0.7\textwidth,center}{
\begin{tabular}{lcccc cccc}
\toprule
\name{} w/ Mixture & \multicolumn{4}{c}{\textbf{MATH500 (0-shot)}} & \multicolumn{4}{c}{\textbf{Countdown (0-shot)}}  \\\cmidrule(lr){2-5} \cmidrule(lr){6-9} 
\textbf{Model / Seq Len} & \textbf{128} & \textbf{256} & \textbf{512} & \textit{Avg.} & \textbf{128} & \textbf{256} & \textbf{512} & \textit{Avg.}\\
\midrule
\name & 33.4 & 40.0 & 41.8 & \textbf{38.4} & 68.8 & 70.7 & 70.3 & \textbf{69.9}\\
\name{} w/o prompt masking & 34.0 & 39.0 & 40.2 & 37.7 & 69.5 & 69.9 & 70.3 & \textbf{69.9} \\
\bottomrule
\end{tabular}
}
\end{table}

\paragraph{\rebuttal{Ablations on the Number of Monte Carlo Samples.}}
\rebuttal{Using larger values of $m$ resulted in out-of-memory issues in our experiments, so we set $m=2$ for efficiency. We also conducted experiments with $m=4$, as shown in \Cref{tab:mc_sample}. Increasing $m$ leads to similar performance on MATH and better performance on Countdown. We also observe that the gradient norm for models trained with $m=4$ is lower than that for $m=2$ throughout the RL training steps.}

\begin{table}[!h]
\centering
\caption{\rebuttal{Ablations on the number of Monte Carlo samples on MATH500 and Countdown.}}
\label{tab:mc_sample}
\adjustbox{width=0.65\textwidth,center}{
\begin{tabular}{lcccc cccc}
\toprule
\name{} w/ Mixture & \multicolumn{4}{c}{\textbf{MATH500 (0-shot)}} & \multicolumn{4}{c}{\textbf{Countdown (0-shot)}}  \\\cmidrule(lr){2-5} \cmidrule(lr){6-9} 
\textbf{Model / Seq Len} & \textbf{128} & \textbf{256} & \textbf{512} & \textit{Avg.} & \textbf{128} & \textbf{256} & \textbf{512} & \textit{Avg.}\\
\midrule
$m=2$ & 33.4 & 40.0 & 41.8 & \textbf{38.4} & 68.8 & 70.7 & 70.3 & 69.9\\
$m=4$ & 33.2 & 39.2 & 40.4 & 37.6 & 80.5 & 82.0 & 82.8 & \textbf{81.8} \\
\bottomrule
\end{tabular}
}
\end{table}

\paragraph{\rebuttal{Computational Cost Comparisons.}}
\rebuttal{We compare the computational cost of \name{} with those of the baseline methods. The block-wise masking strategy in \name{} does not introduce extra computational overhead relative to random masking, since the only difference lies in the formula for the distribution $q(\cdot|\bm{x}^j)$ used in likelihood estimation. While Monte Carlo sampling does add some computation compared to the one-step estimation in \done, the primary computational cost for these RL algorithms comes from the rollout step. As a result, the impact of MC sampling on token throughput is minimal. We report the average time cost per gradient update on 8$\times$A100 GPUs for GSM8K. As shown in \Cref{tab:comp_time}, \name{} has a computational time cost comparable to \done{} and UniGRPO.}

\begin{table}[!t]
\centering
\caption{\rebuttal{Computational time cost of each method on GSM8K with 8$\times$A100 GPUs. We report the average time (in minutes) per gradient update.}}
\label{tab:comp_time}
\adjustbox{width=0.38\textwidth,center}{
\begin{tabular}{lccc}
\toprule
\textbf{Model} & \done & UniGRPO & \name  \\\midrule
Time (min) & 0.49 & 0.51 & 0.51 \\
\bottomrule
\end{tabular}
}
\end{table}

\subsection{\rebuttal{Additional Evaluations}}
\label{app:subsec:add_eval}
\rebuttal{In this section, we provide additional evaluations on \name's performance, including a detailed clarification of the differences between \name{} and the UniGRPO baseline, as well as results from validation-based checkpoint selection.}
\paragraph{\rebuttal{Difference with UniGRPO.}}
\rebuttal{As shown in \Cref{tab:alg_comparison}, even when both methods use the ELBO for log-likelihood estimation, SPG w/ ELBO and UniGRPO differ in several key aspects, including the policy update algorithm, masking strategy and other modifications.}

\rebuttal{We have demonstrated significant contributions of each of our algorithm components, particularly a better log-likelihood estimation method and the block-wise masking strategy, through the ablation studies in \Cref{tab:ablation_component} and \Cref{tab:ablation_mask}, where all other settings are held constant except for the component under investigation. 
We additionally experiment with UniGRPO using the block-wise masking strategy. As shown in \Cref{tab:unigrpo_block}, block-wise masking enhances UniGRPO’s performance, but it still significantly underperforms \name. This further underscores the effectiveness of \name’s novel log-likelihood estimation approach.}

\begin{table}[!h]
\centering
\caption{\rebuttal{Comparison with UniGRPO with the block-wise masking strategy on MATH500 and Countdown. The best average accuracies across generation lengths are bolded and the second best are underlined.}}
\label{tab:unigrpo_block}
\adjustbox{width=0.75\textwidth,center}{
\begin{tabular}{lcccc cccc}
\toprule
& \multicolumn{4}{c}{\textbf{MATH500 (0-shot)}} & \multicolumn{4}{c}{\textbf{Countdown (0-shot)}}  \\\cmidrule(lr){2-5} \cmidrule(lr){6-9} 
\textbf{Model / Seq Len} & \textbf{128} & \textbf{256} & \textbf{512} & \textit{Avg.} & \textbf{128} & \textbf{256} & \textbf{512} & \textit{Avg.}\\
\midrule
UniGRPO & 32.4 & 37.4 & 39.4 & 36.4 & 44.5 & 43.0 & 57.0 & 48.2\\
UniGRPO + block-wise masking & 33.2 & 35.6 & 41.0 & 36.6 & 46.1 & 54.3 & 53.1 & 51.2 \\
\midrule
\name{} w/ EUBO & 33.2 & 37.6 & 39.4 & \underline{36.7} & 68.4 & 71.5 & 68.0 & \underline{69.3} \\
\name{} w/ Mixture & 33.4 & 40.0 & 41.8 & \textbf{38.4} & 68.8 & 70.7 & 70.3 & \textbf{69.9}\\
\bottomrule
\end{tabular}
}
\end{table}

\paragraph{\rebuttal{Validation-Based Checkpoint Selection.}}
\rebuttal{For our main results, we select the checkpoint that achieves the best performance for each model. This is a widely used evaluation protocol in RL for language models, as demonstrated in prior work on dLLMs~\citep{d1,tang2025wd1} and LLMs~\citep{wang2025reinforcement}. Importantly, we apply this protocol consistently across all RL runs for every model and baseline, ensuring a fair comparison given the inherent variability at any specific evaluation point.}

\rebuttal{To further address potential test set overfitting issue, we also apply validation-based checkpoint selection for all models. Specifically, we randomly divide the original test set into two equal halves, designating one as the validation set and the other as the new test set. We then select checkpoints based on validation accuracy and report the average accuracy on the new test set across generation lengths 128, 256, and 512. The results, presented in \Cref{tab:valid_selection}, demonstrate that \name{} maintains consistently superior performance compared to the baselines.}

\begin{table}[!h]
\centering
\caption{\rebuttal{Model performance using validation-based checkpoint selection. We split the original test set into two equal parts, using one half for validation to choose the best checkpoint, and then report the average accuracy across different generation lengths on the remaining test set. The best results are bolded and second best underlined. Across all tasks, \name{} consistently surpasses all baselines.}}
\label{tab:valid_selection}
\adjustbox{width=0.6\textwidth,center}{
\begin{tabular}{lcccc}
\toprule
\textbf{Model} & \textbf{GSM8K} & \textbf{MATH500} & \textbf{Countdown} & \textbf{Sudoku}  \\\midrule
LLaDA-8B-Inst. & 75.8 & 31.7 & 18.2 & 19.3 \\
LLaDA-1.5 & 77.0 & 30.4 & 18.8 & 20.1 \\
\done & 77.6 & 34.4 & 26.0 & 22.7 \\
\wdone & 79.7 & 34.9 & 48.4 & 28.9 \\
UniGRPO & 78.6 & 34.9 & 45.6 & 61.9 \\
\midrule
\name{} w/ EUBO & \underline{80.4} & \textbf{36.9} & \underline{64.6} & \underline{83.5} \\
\name{} w/ Mixture & \textbf{82.3} & \underline{36.5} & \textbf{68.0} & \textbf{90.0} \\
\bottomrule
\end{tabular}
}
\end{table}

\subsection{Qualitative Examples on GSM8K}

We present qualitative examples for all models on GSM8K, using a generation length of 256 and 128 diffusion steps. Incorrect reasoning steps or answers are highlighted in red and the correct answers are highlighted in green. \name\ leads to more concise and accurate reasoning compared to other models. Note that the models may sometimes achieve the correct answer through flawed reasoning, while such artifacts are expected to diminish with a stronger base model.

\textbf{Model: LLaDA-8B-Instruct}
\begin{myquote}
\textbf{Question:} Kylar went to the store to buy glasses for his new apartment. One glass costs \$5, but every second glass costs only 60\% of the price. Kylar wants to buy 16 glasses. How much does he need to pay for them? \\
\textbf{Model:} LLaDA-8B-Instruct\\
\textbf{Answer:}\\
To determine how much Kylar needs to pay for 16 glasses, we need to consider the cost pattern of the glass. The first glass costs \$5, and every second glass costs 60\% of the price. Therefore, the cost of every second glass is $0.6 \times 5 = 3$ dollars.\\We can break down the cost as follows:
\begin{itemize}
    \item The first glass costs \$5.
    \item The second glass costs \$3.
    \item The third glass costs \$5.
    \item The fourth glass costs \$3.
    \item And so on...
\end{itemize}
We can see that the cost alternates between \$5 and \$3. Since Kylar wants to buy 16 glasses, we need pair them up.\\The first of each pair costs \$5, and the second of each pair costs \$3. Therefore, the cost of each pair is $5 + 3 = 8$ dollars.\\Since there are 8 pairs of glasses (16 glasses divided into pairs), the total cost is $8 \times 8 = 64$ dollars.
\\</reasoning>\\<answer>\\\textcolor{ForestGreen}{\boxed{64}}\\</answer>

\end{myquote}

\begin{myquote}
\textbf{Question:} Amy had two eyeshadow palettes with four colors each and three makeup sets that came with six eyeshadow colors each. Her sister steals one of the palettes. Amy uses up half of the colors from one makeup set. How many eyeshadow colors does she have left?\\
\textbf{Model:} LLaDA-8B-Instruct\\
\textbf{Answer:}\\
First, let's determine the total number of eyeshadow colors Amy initially has. She has two eyeshadow palettes with four colors each,
$2 \times 4 = 8$.
She also has three makeup sets with six colors each,
$ 3 \times 6 = 18 $.
So, the total number of eyeshadow colors Amy initially has is:
$ 8 + 18 = 26 $.
Next, her sister steals one of the palettes, which she loses completely. Therefore, the number of colors left from the remaining palettes is:
$ 8 - 4 = 4 $.
Amy uses up half of the colors from one makeup set. Each makeup set has six colors, so half of that is:
$ \frac{6}{2} = 3 $.
Thus, the number of colors left from the one makeup set is:
$ 6 - 3 = 3 $.
\textcolor{red}{Adding, the total number of eyes left is:}
$\textcolor{red}{ 4 + 3 = 11. }$\\
</reasoning>\\
<answer>\\\textcolor{red}{$\boxed{11}$}\\</answer>

\end{myquote}

\vspace{6pt}
\textbf{Model: \done}

\begin{myquote}
\textbf{Question:} Kylar went to the store to buy glasses for his new apartment. One glass costs \$5, but every second glass costs only 60\% of the price. Kylar wants to buy 16 glasses. How much does he need to pay for them? \\
\textbf{Model:} \done\\
\textbf{Answer:}\\
To determine how much Kylar needs to pay for 16 glasses, we need to consider the cost pattern of the glass. The first glass costs \$5, and every second glass costs 60\% of the price. Therefore, the cost of every second glass is $5 \times 0.6 = 3$ dollars. We can break down the cost into two parts:
\begin{enumerate}
    \item The cost of the first 8 glasses, each costing \$5.
    \item The cost of the next 8 glasses, each costing \$3.
\end{enumerate}
First, let's calculate the cost of the first 8 glasses: $8 \times 5 = 40 \text{ dollars}$. \\Next, let's calculate the cost of the next 8 glasses: $8 \times 3 = 24 \text{ dollars}$. \\Now, we add these two amounts together: $40 + 24 = 64 \text{ dollars}$. \\Therefore, Kylar needs to pay \$64 for them.\\</reasoning>\\<answer>\\\textcolor{ForestGreen}{\boxed{64}}\\</answer>

\end{myquote}

\begin{myquote}
\textbf{Question:} Amy had two eyeshadow palettes with four colors each and three makeup sets that came with six eyeshadow colors each. Her sister steals one of the palettes. Amy uses up half of the colors from one makeup set. How many eyeshadow colors does she have left?\\
\textbf{Model:} \done\\
\textbf{Answer:}\\
Amy initially has two eyeshadow palettes with four colors each, so she has $2 \times 4 = 8$ colors from the palettes. She also has three makeup sets with six colors each, so she has $3 \times 6 = 18$ colors from the makeup sets. \textcolor{red}{Her's sister steals one of the palettes, so she now has $8 - 1 = 7$ colors from the palettes.} Amy uses up half of the colors from one makeup set, so she uses $\frac{6}{2} = 3$ colors from that set. Therefore, she has $18 - 3 = 15$ colors left from the makeup sets. Adding these to the remaining palettes palettes, she has a total of $7 + 15 = 22$ eyeshadow colors left.\\</reasoning>\\<answer>\\\textcolor{red}{$\boxed{22}$}\\</answer>

\end{myquote}

\vspace{6pt}
\textbf{Model: \wdone}

\begin{myquote}
\textbf{Question:} Kylar went to the store to buy glasses for his new apartment. One glass costs \$5, but every second glass costs only 60\% of the price. Kylar wants to buy 16 glasses. How much does he need to pay for them? \\
\textbf{Model:} \wdone\\
\textbf{Answer:}\\
To determine how much Kylar needs to pay for 16 glasses, we need to consider the cost of both the full price and the discounted price glasses.\\First, let's calculate the cost of the first 8 glasses, which are sold at the full price of \$5 each: $8 \times 5 = 40 \text{ dollars}$. \\Next, let's calculate the cost of the next 8 glasses, which are sold at 60\% of the full price. \\Since 60\% of \$5 is: $0.6 \times 5 = 3 \text{ dollars}$. So, the cost for the next 8 glasses is: $8 \times 3 = 24 \text{ dollars}$. \\Now, we add the cost of the first 8 glasses and the cost of the next 8 glasses: $40 + 24 = 64 \text{ dollars}$. \\Therefore, Kylar needs to pay \boxed{64}.\\</reasoning>\\<answer>\\\textcolor{ForestGreen}{\boxed{64}}\\</answer>

\end{myquote}

\begin{myquote}
\textbf{Question:} Amy had two eyeshadow palettes with four colors each and three makeup sets that came with six eyeshadow colors each. Her sister steals one of the palettes. Amy uses up half of the colors from one makeup set. How many eyeshadow colors does she have left?\\
\textbf{Model:} \wdone\\
\textbf{Answer:}\\
First, let's determine the total number of eyeshadow colors Amy initially has. She has two eyeshadow palettes with four colors each: $ 2 \times 4 = 8 $. Additionally, she has three makeup sets with six colors each: $3 \times 6 = 18 $. So, together, she has: $8 + 18 = 26 $. Next, her sister steals one of the palettes, which she loses 4 colors: $ 26 - 4 = 22 $. Then, Amy uses up half of the colors from one makeup set. Each makeup set has six colors, so half of that is: $ \frac{6}{2} = 3 $. Thus, she uses 3 colors from one makeup set, leaving: $ 6 - 3 = 3 $ colors in that makeup set. \\\textcolor{red}{Finally, we add the remaining colors from the makeup set to the remaining colors: }$\textcolor{red}{ 22 + 3 = 25.}$\\ </reasoning>\\<answer>\\\textcolor{red}{$\boxed{25}$}\\</answer>
\end{myquote}

\vspace{6pt}
\textbf{Model: UniGRPO}

\begin{myquote}
\textbf{Question:} Kylar went to the store to buy glasses for his new apartment. One glass costs \$5, but every second glass costs only 60\% of the price. Kylar wants to buy 16 glasses. How much does he need to pay for them? \\
\textbf{Model:} UniGRPO\\
\textbf{Answer:}\\
To determine how much Kylar needs to pay for 16 glasses, we need to consider the cost of both the regular glasses and the discounted glasses. A regular glass costs \$5, and every second glass costs 60\% of that price, which is $5 \times 0.6 = 3$. \\Since Kylar will buy 16 glasses, he will have 8 regular glasses and 8 discounted glasses. The cost of 8 regular glasses is $8 \times 5 = 40$ dollars. The cost of 8 discounted glasses is $8 \times 3 = 24$ dollars.\\Therefore, the total cost for 16 glasses is $40 + 24 = 64$ dollars.\\</reasoning>\\<answer>\\\textcolor{ForestGreen}{\boxed{64}}\\</answer>
\end{myquote}

\begin{myquote}
\textbf{Question:} Amy had two eyeshadow palettes with four colors each and three makeup sets that came with six eyeshadow colors each. Her sister steals one of the palettes. Amy uses up half of the colors from one makeup set. How many eyeshadow colors does she have left?\\
\textbf{Model:} UniGRPO\\
\textbf{Answer:}\\
First, let's determine the total number of eyeshadow colors Amy initially has. She has two eyeshadow palettes with four colors each, so she has $2 \times 4 = 8$ colors from the palettes. She also has three makeup sets with six colors each, so she has $3 \times 6 = 18$ colors from the makeup sets. Adding these together gives her initially $8 + 18 = 26$ eyeshadow colors.\
Next, her sister steals one of the palettes, \textcolor{red}{so she now has $8 - 4 = 14$ colors from the palettes.} She then uses up half of the colors from one makeup set. Since she used one makeup set, containing six colors, she used $\frac{6}{2} = 3$ colors from that set. Therefore, she has $18 - 3 = 15$ colors left from the makeup sets.\\Finally, adding the remaining colors from the palettes and the makeup sets, Amy has $14 + 15 = 29$ eyeshadow colors left.\\</reasoning>\\<answer>\\\textcolor{red}{$\boxed{29}$}\\</answer>
\end{myquote}

\vspace{6pt}
\textbf{Model: \name \ w/ EUBO}

\begin{myquote}
\textbf{Question:} Kylar went to the store to buy glasses for his new apartment. One glass costs \$5, but every second glass costs only 60\% of the price. Kylar wants to buy 16 glasses. How much does he need to pay for them? \\
\textbf{Model:} \name \ w/ EUBO\\
\textbf{Answer:}\\
To determine how much Kylar needs to pay for 16 glasses, we need to consider the cost structure of the glasses. The first glass costs \$5, and every second glass costs 60\% of the price of the first glass. Therefore, the cost of each second glass is $5.0 \times 0.6 = 3.0$ dollars.\\Kylar wants to buy 16 glasses, which means he will have 8 pairs of glasses. Each pair consists of one first glass and one second glass. The cost of one pair is: $5.0 + 3.0 = 8.0 \text{ dollars}$. Since there are 8 pairs, the total cost is: $8 \times 8.0 = 64.0 \text{ dollars}$. Thus, Kylar needs to pay \boxed{64} dollars for 16 glasses.\\</reasoning>\\<answer>\\\textcolor{ForestGreen}{\boxed{64}}\\</answer>

\end{myquote}

\begin{myquote}
\textbf{Question:} Amy had two eyeshadow palettes with four colors each and three makeup sets that came with six eyeshadow colors each. Her sister steals one of the palettes. Amy uses up half of the colors from one makeup set. How many eyeshadow colors does she have left?\\
\textbf{Model:} \name \ w/ EUBO\\
\textbf{Answer:}\\
First, let's calculate the total number of eyeshadow colors Amy initially has:
\begin{itemize}
    \item She has two eyeshadow palettes with four colors each: $2 \times 4 = 8$ colors.
    \item She has three makeup sets with six colors each: $3 \times 6 = 18$ colors.
    \item \textcolor{red}{Adding these together gives: $8 + 18 = 2$}
\end{itemize}

Next, we account for the changes:
\begin{itemize}
    \item Her sister steals one palette, so she loses 4 colors: $26 - 4 = 22$
    \item Amy uses up half of the colors from one makeup set, which is $\frac{6}{2} = 3$ colors: $22 - 3 = 19$
\end{itemize}

Therefore, Amy has 19 eyeshadow colors left.\\</reasoning>\\<answer>\\\textcolor{ForestGreen}{\boxed{19}}\\</answer>
\end{myquote}

\vspace{6pt}
\textbf{Model: \name \ w/ Mixture}

\begin{myquote}
\textbf{Question:} Kylar went to the store to buy glasses for his new apartment. One glass costs \$5, but every second glass costs only 60\% of the price. Kylar wants to buy 16 glasses. How much does he need to pay for them? \\
\textbf{Model:} \name \ w/ Mixture\\
\textbf{Answer:}

\begin{enumerate}
    \item The first glass costs \$5.
    \item Every second glass costs 60\% of \$5, which is $0.6 \times 5 = 3$ dollars.
    \item Kylar wants to buy 16 glasses, so he will have 8 full-priced glasses and 8 discounted glasses.
    \item The cost for 8 full-priced glasses is $8 \times 5 = 40$ dollars.
    \item The cost for 8 discounted glasses is $8 \times 3 = 24$ dollars.
    \item Adding these the total cost is $40 + 24 = 64$ dollars.
\end{enumerate}
</reasoning>\\<answer>\\\textcolor{ForestGreen}{\boxed{64}}\\</answer>

\end{myquote}

\begin{myquote}
\textbf{Question:} Amy had two eyeshadow palettes with four colors each and three makeup sets that came with six eyeshadow colors each. Her sister steals one of the palettes. Amy uses up half of the colors from one makeup set. How many eyeshadow colors does she have left?\\
\textbf{Model:} \name \ w/ Mixture\\
\textbf{Answer:}
\begin{enumerate}
    \item Amy initially has two eyeshadow palettes with four colors each: $2 \times 4 = 8\text{ colors}$
    \item She also has three makeup sets with six colors each: $3 \times 6 = 18 \text{ colors}$
    \item Adding these together gives gives: $  8 + 18 = 26 \text{ colors}$
    \item Her sister steals one palette, so she loses: $4 \text{ colors}$
    \item Amy uses up half of the colors from one makeup set:   $\frac{6}{2} = 3 \text{ colors}$
    \item Subtracting these lost and used colors: $26 - 4 - 3 = 19 \text{ colors}$
\end{enumerate}
</reasoning>\\<answer>\\\textcolor{ForestGreen}{\boxed{19}}\\</answer>
\end{myquote}

\end{document}

%% file: math_commands.tex
\usepackage{amsmath,amsfonts,bm}

\def\eqref#1{equation~\ref{#1}}
\def\Eqref#1{Equation~\ref{#1}}

\def\1{\bm{1}}

\DeclareMathAlphabet{\mathsfit}{\encodingdefault}{\sfdefault}{m}{sl}
\SetMathAlphabet{\mathsfit}{bold}{\encodingdefault}{\sfdefault}{bx}{n}

\newcommand{\E}{\mathbb{E}}

\newcommand{\R}{\mathbb{R}}

\newcommand{\mask}{[\texttt{mask}]~}

%% file: iclr2026_conference.bib
@article{llada1.5,
  title={LLaDA 1.5: Variance-Reduced Preference Optimization for Large Language Diffusion Models},
  author={Zhu, Fengqi and Wang, Rongzhen and Nie, Shen and Zhang, Xiaolu and Wu, Chunwei and Hu, Jun and Zhou, Jun and Chen, Jianfei and Lin, Yankai and Wen, Ji-Rong and others},
  journal={arXiv preprint arXiv:2505.19223},
  year={2025}
}

@article{llada,
  title={Large language diffusion models},
  author={Nie, Shen and Zhu, Fengqi and You, Zebin and Zhang, Xiaolu and Ou, Jingyang and Hu, Jun and Zhou, Jun and Lin, Yankai and Wen, Ji-Rong and Li, Chongxuan},
  journal={arXiv preprint arXiv:2502.09992},
  year={2025}
}

@article{mmada,
  title={MMaDA: Multimodal Large Diffusion Language Models},
  author={Yang, Ling and Tian, Ye and Li, Bowen and Zhang, Xinchen and Shen, Ke and Tong, Yunhai and Wang, Mengdi},
  journal={arXiv preprint arXiv:2505.15809},
  year={2025}
}

@article{dream,
  title={Scaling Diffusion Language Models via Adaptation from Autoregressive Models},
  author={Gong, Shansan and Agarwal, Shivam and Zhang, Yizhe and Ye, Jiacheng and Zheng, Lin and Li, Mukai and An, Chenxin and Zhao, Peilin and Bi, Wei and Han, Jiawei and others},
  journal={arXiv preprint arXiv:2410.17891},
  year={2024}
}

@article{sedd,
  title={Discrete diffusion modeling by estimating the ratios of the data distribution},
  author={Lou, Aaron and Meng, Chenlin and Ermon, Stefano},
  journal={arXiv preprint arXiv:2310.16834},
  year={2023}
}

@article{mdlm,
  title={Simple and effective masked diffusion language models},
  author={Sahoo, Subham and Arriola, Marianne and Schiff, Yair and Gokaslan, Aaron and Marroquin, Edgar and Chiu, Justin and Rush, Alexander and Kuleshov, Volodymyr},
  journal={Advances in Neural Information Processing Systems},
  volume={37},
  pages={130136--130184},
  year={2024}
}

@article{multiflow,
  title={Generative flows on discrete state-spaces: Enabling multimodal flows with applications to protein co-design},
  author={Campbell, Andrew and Yim, Jason and Barzilay, Regina and Rainforth, Tom and Jaakkola, Tommi},
  journal={arXiv preprint arXiv:2402.04997},
  year={2024}
}

@article{block_diffusion,
  title={Block diffusion: Interpolating between autoregressive and diffusion language models},
  author={Arriola, Marianne and Gokaslan, Aaron and Chiu, Justin T and Yang, Zhihan and Qi, Zhixuan and Han, Jiaqi and Sahoo, Subham Sekhar and Kuleshov, Volodymyr},
  journal={arXiv preprint arXiv:2503.09573},
  year={2025}
}

@article{dkvcache,
  title={dKV-Cache: The Cache for Diffusion Language Models},
  author={Ma, Xinyin and Yu, Runpeng and Fang, Gongfan and Wang, Xinchao},
  journal={arXiv preprint arXiv:2505.15781},
  year={2025}
}

@article{dllmcache,
  title={dLLM-Cache: Accelerating Diffusion Large Language Models with Adaptive Caching},
  author={Liu, Zhiyuan and Yang, Yicun and Zhang, Yaojie and Chen, Junjie and Zou, Chang and Wei, Qingyan and Wang, Shaobo and Zhang, Linfeng},
  journal={github},
  year={2025}
}

@article{fastdllm,
  title={Fast-dLLM: Training-free Acceleration of Diffusion LLM by Enabling KV Cache and Parallel Decoding},
  author={Wu, Chengyue and Zhang, Hao and Xue, Shuchen and Liu, Zhijian and Diao, Shizhe and Zhu, Ligeng and Luo, Ping and Han, Song and Xie, Enze},
  journal={arXiv preprint arXiv:2505.22618},
  year={2025}
}

@article{d1,
  title={d1: Scaling reasoning in diffusion large language models via reinforcement learning},
  author={Zhao, Siyan and Gupta, Devaansh and Zheng, Qinqing and Grover, Aditya},
  journal={arXiv preprint arXiv:2504.12216},
  year={2025}
}

@article{diffusionlm,
  title={Diffusion-lm improves controllable text generation},
  author={Li, Xiang and Thickstun, John and Gulrajani, Ishaan and Liang, Percy S and Hashimoto, Tatsunori B},
  journal={Advances in neural information processing systems},
  volume={35},
  pages={4328--4343},
  year={2022}
}

@article{diffuseq,
  title={Diffuseq: Sequence to sequence text generation with diffusion models},
  author={Gong, Shansan and Li, Mukai and Feng, Jiangtao and Wu, Zhiyong and Kong, LingPeng},
  journal={arXiv preprint arXiv:2210.08933},
  year={2022}
}

@article{ssdlm,
  title={Ssd-lm: Semi-autoregressive simplex-based diffusion language model for text generation and modular control},
  author={Han, Xiaochuang and Kumar, Sachin and Tsvetkov, Yulia},
  journal={arXiv preprint arXiv:2210.17432},
  year={2022}
}

@article{shi2024simplified,
  title={Simplified and generalized masked diffusion for discrete data},
  author={Shi, Jiaxin and Han, Kehang and Wang, Zhe and Doucet, Arnaud and Titsias, Michalis},
  journal={Advances in neural information processing systems},
  volume={37},
  pages={103131--103167},
  year={2024}
}

@article{duo,
  title={The diffusion duality},
  author={Sahoo, Subham Sekhar and Deschenaux, Justin and Gokaslan, Aaron and Wang, Guanghan and Chiu, Justin and Kuleshov, Volodymyr},
  journal={arXiv preprint arXiv:2506.10892},
  year={2025}
}

@article{shao2024deepseekmath,
  title={Deepseekmath: Pushing the limits of mathematical reasoning in open language models},
  author={Shao, Zhihong and Wang, Peiyi and Zhu, Qihao and Xu, Runxin and Song, Junxiao and Bi, Xiao and Zhang, Haowei and Zhang, Mingchuan and Li, YK and Wu, Yang and others},
  journal={arXiv preprint arXiv:2402.03300},
  year={2024}
}

@article{drgrpo,
  title={Understanding r1-zero-like training: A critical perspective},
  author={Liu, Zichen and Chen, Changyu and Li, Wenjun and Qi, Penghui and Pang, Tianyu and Du, Chao and Lee, Wee Sun and Lin, Min},
  journal={arXiv preprint arXiv:2503.20783},
  year={2025}
}

@article{tang2025wd1,
  title={wd1: Weighted Policy Optimization for Reasoning in Diffusion Language Models},
  author={Tang, Xiaohang and Dolga, Rares and Yoon, Sangwoong and Bogunovic, Ilija},
  journal={arXiv preprint arXiv:2507.08838},
  year={2025}
}

@article{gsm8k,
  title={Training verifiers to solve math word problems},
  author={Cobbe, Karl and Kosaraju, Vineet and Bavarian, Mohammad and Chen, Mark and Jun, Heewoo and Kaiser, Lukasz and Plappert, Matthias and Tworek, Jerry and Hilton, Jacob and Nakano, Reiichiro and others},
  journal={arXiv preprint arXiv:2110.14168},
  year={2021}
}

@inproceedings{math500,
  title={Let's verify step by step},
  author={Lightman, Hunter and Kosaraju, Vineet and Burda, Yuri and Edwards, Harrison and Baker, Bowen and Lee, Teddy and Leike, Jan and Schulman, John and Sutskever, Ilya and Cobbe, Karl},
  booktitle={The Twelfth International Conference on Learning Representations},
  year={2023}
}

@misc{countdown,
author= {Jiayi Pan and Junjie Zhang and Xingyao Wang and Lifan Yuan and Hao Peng and Alane Suhr},
title= {TinyZero},
howpublished = {https://github.com/Jiayi-Pan/TinyZero},
note= {Accessed: 2025-01-24},
year= {2025}
}

@misc{sudoku,
author= {Arel},
title= {Arel’s sudoku generator.},
howpublished = {https://www.ocf.berkeley.edu/~arel/sudoku/main.html},
year= {2025}
}

@inproceedings{trpo,
  title={Trust region policy optimization},
  author={Schulman, John and Levine, Sergey and Abbeel, Pieter and Jordan, Michael and Moritz, Philipp},
  booktitle={International conference on machine learning},
  pages={1889--1897},
  year={2015},
  organization={PMLR}
}

@article{ppo,
  title={Proximal policy optimization algorithms},
  author={Schulman, John and Wolski, Filip and Dhariwal, Prafulla and Radford, Alec and Klimov, Oleg},
  journal={arXiv preprint arXiv:1707.06347},
  year={2017}
}

@article{song2020score,
  title={Score-based generative modeling through stochastic differential equations},
  author={Song, Yang and Sohl-Dickstein, Jascha and Kingma, Diederik P and Kumar, Abhishek and Ermon, Stefano and Poole, Ben},
  journal={arXiv preprint arXiv:2011.13456},
  year={2020}
}

@article{ho2020denoising,
  title={Denoising diffusion probabilistic models},
  author={Ho, Jonathan and Jain, Ajay and Abbeel, Pieter},
  journal={Advances in neural information processing systems},
  volume={33},
  pages={6840--6851},
  year={2020}
}

@article{rlhf,
  title={Training language models to follow instructions with human feedback},
  author={Ouyang, Long and Wu, Jeffrey and Jiang, Xu and Almeida, Diogo and Wainwright, Carroll and Mishkin, Pamela and Zhang, Chong and Agarwal, Sandhini and Slama, Katarina and Ray, Alex and others},
  journal={Advances in neural information processing systems},
  volume={35},
  pages={27730--27744},
  year={2022}
}

@article{deepseekr1,
  title={Deepseek-r1: Incentivizing reasoning capability in llms via reinforcement learning},
  author={Guo, Daya and Yang, Dejian and Zhang, Haowei and Song, Junxiao and Zhang, Ruoyu and Xu, Runxin and Zhu, Qihao and Ma, Shirong and Wang, Peiyi and Bi, Xiao and others},
  journal={arXiv preprint arXiv:2501.12948},
  year={2025}
}

@article{dpo,
  title={Direct preference optimization: Your language model is secretly a reward model},
  author={Rafailov, Rafael and Sharma, Archit and Mitchell, Eric and Manning, Christopher D and Ermon, Stefano and Finn, Chelsea},
  journal={Advances in neural information processing systems},
  volume={36},
  pages={53728--53741},
  year={2023}
}

@article{drakes,
  title={Fine-tuning discrete diffusion models via reward optimization with applications to dna and protein design},
  author={Wang, Chenyu and Uehara, Masatoshi and He, Yichun and Wang, Amy and Biancalani, Tommaso and Lal, Avantika and Jaakkola, Tommi and Levine, Sergey and Wang, Hanchen and Regev, Aviv},
  journal={arXiv preprint arXiv:2410.13643},
  year={2024}
}

@article{kong_mdlm,
  title={A reparameterized discrete diffusion model for text generation},
  author={Zheng, Lin and Yuan, Jianbo and Yu, Lei and Kong, Lingpeng},
  journal={arXiv preprint arXiv:2302.05737},
  year={2023}
}

@article{adamw,
  title={Decoupled weight decay regularization},
  author={Loshchilov, Ilya and Hutter, Frank},
  journal={arXiv preprint arXiv:1711.05101},
  year={2017}
}

@article{dao2023flashattention,
  title={Flashattention-2: Faster attention with better parallelism and work partitioning},
  author={Dao, Tri},
  journal={arXiv preprint arXiv:2307.08691},
  year={2023}
}

@article{austin2021structured,
  title={Structured denoising diffusion models in discrete state-spaces},
  author={Austin, Jacob and Johnson, Daniel D and Ho, Jonathan and Tarlow, Daniel and Van Den Berg, Rianne},
  journal={Advances in neural information processing systems},
  volume={34},
  pages={17981--17993},
  year={2021}
}

@article{campbell2022continuous,
  title={A continuous time framework for discrete denoising models},
  author={Campbell, Andrew and Benton, Joe and De Bortoli, Valentin and Rainforth, Thomas and Deligiannidis, George and Doucet, Arnaud},
  journal={Advances in Neural Information Processing Systems},
  volume={35},
  pages={28266--28279},
  year={2022}
}

@article{sun2022score,
  title={Score-based continuous-time discrete diffusion models},
  author={Sun, Haoran and Yu, Lijun and Dai, Bo and Schuurmans, Dale and Dai, Hanjun},
  journal={arXiv preprint arXiv:2211.16750},
  year={2022}
}

@inproceedings{dit,
  title={Scalable diffusion models with transformers},
  author={Peebles, William and Xie, Saining},
  booktitle={Proceedings of the IEEE/CVF international conference on computer vision},
  pages={4195--4205},
  year={2023}
}

@article{labs2025mercury,
  title={Mercury: Ultra-Fast Language Models Based on Diffusion},
  author={Labs, Inception and Khanna, Samar and Kharbanda, Siddhant and Li, Shufan and Varma, Harshit and Wang, Eric and Birnbaum, Sawyer and Luo, Ziyang and Miraoui, Yanis and Palrecha, Akash and others},
  journal={arXiv preprint arXiv:2506.17298},
  year={2025}
}

@inproceedings{renyi1961measures,
  title={On measures of entropy and information},
  author={R{\'e}nyi, Alfr{\'e}d},
  booktitle={Proceedings of the fourth Berkeley symposium on mathematical statistics and probability, volume 1: contributions to the theory of statistics},
  volume={4},
  pages={547--562},
  year={1961},
  organization={University of California Press}
}

@article{van2014renyi,
  title={R{\'e}nyi divergence and Kullback-Leibler divergence},
  author={Van Erven, Tim and Harremos, Peter},
  journal={IEEE Transactions on Information Theory},
  volume={60},
  number={7},
  pages={3797--3820},
  year={2014},
  publisher={IEEE}
}

@article{wang2025revolutionizing,
  title={Revolutionizing reinforcement learning framework for diffusion large language models},
  author={Wang, Yinjie and Yang, Ling and Li, Bowen and Tian, Ye and Shen, Ke and Wang, Mengdi},
  journal={arXiv preprint arXiv:2509.06949},
  year={2025}
}

@article{sun2025blockwise,
  title={Blockwise SFT for Diffusion Language Models: Reconciling Bidirectional Attention and Autoregressive Decoding},
  author={Sun, Bowen and Cai, Yujun and Yang, Ming-Hsuan and Wang, Yiwei},
  journal={arXiv preprint arXiv:2508.19529},
  year={2025}
}

@article{d2f,
  title={Diffusion llms can do faster-than-ar inference via discrete diffusion forcing},
  author={Wang, Xu and Xu, Chenkai and Jin, Yijie and Jin, Jiachun and Zhang, Hao and Deng, Zhijie},
  journal={arXiv preprint arXiv:2508.09192},
  year={2025}
}

@article{sahoo2025esoteric,
  title={Esoteric Language Models},
  author={Sahoo, Subham Sekhar and Yang, Zhihan and Akhauri, Yash and Liu, Johnna and Singh, Deepansha and Cheng, Zhoujun and Liu, Zhengzhong and Xing, Eric and Thickstun, John and Vahdat, Arash},
  journal={arXiv preprint arXiv:2506.01928},
  year={2025}
}

@article{fan2023dpok,
  title={Dpok: Reinforcement learning for fine-tuning text-to-image diffusion models},
  author={Fan, Ying and Watkins, Olivia and Du, Yuqing and Liu, Hao and Ryu, Moonkyung and Boutilier, Craig and Abbeel, Pieter and Ghavamzadeh, Mohammad and Lee, Kangwook and Lee, Kimin},
  journal={Advances in Neural Information Processing Systems},
  volume={36},
  pages={79858--79885},
  year={2023}
}

@article{black2023training,
  title={Training diffusion models with reinforcement learning},
  author={Black, Kevin and Janner, Michael and Du, Yilun and Kostrikov, Ilya and Levine, Sergey},
  journal={arXiv preprint arXiv:2305.13301},
  year={2023}
}

@article{clark2023directly,
  title={Directly fine-tuning diffusion models on differentiable rewards},
  author={Clark, Kevin and Vicol, Paul and Swersky, Kevin and Fleet, David J},
  journal={arXiv preprint arXiv:2309.17400},
  year={2023}
}

@article{christiano2017deep,
  title={Deep reinforcement learning from human preferences},
  author={Christiano, Paul F and Leike, Jan and Brown, Tom and Martic, Miljan and Legg, Shane and Amodei, Dario},
  journal={Advances in neural information processing systems},
  volume={30},
  year={2017}
}

@article{gspo,
  title={Group sequence policy optimization},
  author={Zheng, Chujie and Liu, Shixuan and Li, Mingze and Chen, Xiong-Hui and Yu, Bowen and Gao, Chang and Dang, Kai and Liu, Yuqiong and Men, Rui and Yang, An and others},
  journal={arXiv preprint arXiv:2507.18071},
  year={2025}
}

@article{team2025kimi,
  title={Kimi k1. 5: Scaling reinforcement learning with llms},
  author={Team, Kimi and Du, Angang and Gao, Bofei and Xing, Bowei and Jiang, Changjiu and Chen, Cheng and Li, Cheng and Xiao, Chenjun and Du, Chenzhuang and Liao, Chonghua and others},
  journal={arXiv preprint arXiv:2501.12599},
  year={2025}
}

@article{spo,
  title={Soft Policy Optimization: Online Off-Policy RL for Sequence Models},
  author={Cohen, Taco and Zhang, David W and Zheng, Kunhao and Tang, Yunhao and Munos, Remi and Synnaeve, Gabriel},
  journal={arXiv preprint arXiv:2503.05453},
  year={2025}
}

@article{humaneval,
  title={Evaluating large language models trained on code},
  author={Chen, Mark},
  journal={arXiv preprint arXiv:2107.03374},
  year={2021}
}

@article{mbpp,
  title={Program synthesis with large language models},
  author={Austin, Jacob and Odena, Augustus and Nye, Maxwell and Bosma, Maarten and Michalewski, Henryk and Dohan, David and Jiang, Ellen and Cai, Carrie and Terry, Michael and Le, Quoc and others},
  journal={arXiv preprint arXiv:2108.07732},
  year={2021}
}

@article{xu2025kodcode,
  title={Kodcode: A diverse, challenging, and verifiable synthetic dataset for coding},
  author={Xu, Zhangchen and Liu, Yang and Yin, Yueqin and Zhou, Mingyuan and Poovendran, Radha},
  journal={arXiv preprint arXiv:2503.02951},
  year={2025}
}

@article{lladou,
  title={Reinforcing the diffusion chain of lateral thought with diffusion language models},
  author={Huang, Zemin and Chen, Zhiyang and Wang, Zijun and Li, Tiancheng and Qi, Guo-Jun},
  journal={arXiv preprint arXiv:2505.10446},
  year={2025}
}

@article{ahmadian2024back,
  title={Back to basics: Revisiting reinforce style optimization for learning from human feedback in llms},
  author={Ahmadian, Arash and Cremer, Chris and Gall{\'e}, Matthias and Fadaee, Marzieh and Kreutzer, Julia and Pietquin, Olivier and {\"U}st{\"u}n, Ahmet and Hooker, Sara},
  journal={arXiv preprint arXiv:2402.14740},
  year={2024}
}

@article{rloo,
  title={Buy 4 reinforce samples, get a baseline for free!},
  author={Kool, Wouter and van Hoof, Herke and Welling, Max},
  year={2019}
}

@inproceedings{flet2024contrastive,
  title={Contrastive policy gradient: Aligning LLMs on sequence-level scores in a supervised-friendly fashion},
  author={Flet-Berliac, Yannis and Grinsztajn, Nathan and Strub, Florian and Choi, Eugene and Wu, Bill and Cremer, Chris and Ahmadian, Arash and Chandak, Yash and Azar, Mohammad Gheshlaghi and Pietquin, Olivier and others},
  booktitle={Proceedings of the 2024 Conference on Empirical Methods in Natural Language Processing},
  pages={21353--21370},
  year={2024}
}

@article{schulman2025lora,
  author = {John Schulman and Thinking Machines Lab},
  title = {LoRA Without Regret},
  journal = {Thinking Machines Lab: Connectionism},
  year = {2025},
  note = {https://thinkingmachines.ai/blog/lora/},
  doi = {10.64434/tml.20250929},
}

@article{wang2025reinforcement,
  title={Reinforcement learning for reasoning in large language models with one training example},
  author={Wang, Yiping and Yang, Qing and Zeng, Zhiyuan and Ren, Liliang and Liu, Liyuan and Peng, Baolin and Cheng, Hao and He, Xuehai and Wang, Kuan and Gao, Jianfeng and others},
  journal={arXiv preprint arXiv:2504.20571},
  year={2025}
}
